\documentclass[alpha-refs]{wiley-article}

\usepackage{comment}
\usepackage{siunitx}
\usepackage{amssymb}
\usepackage{graphicx}
\usepackage{booktabs} 
\newcommand{\tabitem}{~~\llap{\textendash}~~}
\usepackage{multirow, bigdelim}
\usepackage{subcaption}
\usepackage[ruled]{algorithm2e}
\usepackage[toc,page]{appendix}
\PassOptionsToPackage{hyphens}{url}\usepackage{hyperref}
\usepackage{threeparttable}
\usepackage{array}
\usepackage{float}
\newcolumntype{M}[1]{>{\centering\arraybackslash}m{#1}}
\usepackage[normalem]{ulem}
\usepackage{tabu}
\usepackage{caption}
\usepackage[table]{xcolor}

\newcommand{\fooo}{\color{black}\makebox[-10pt]{{\large $\textbullet$}}\hskip-0.5pt\vrule width 1pt\hspace{\labelsep}}

\newcommand{\foooo}{\color{black}\makebox[0pt]{{\large \textbullet}}\hskip-0.5pt\vrule width 1pt\hspace{\labelsep}}

\newcolumntype{L}[1]{>{\raggedright\arraybackslash$}p{#1}<{$}}

\newcolumntype{P}[1]{>{\centering\arraybackslash}p{#1}}

\SetExpansion[stretch = 70, shrink = 70,] { encoding = {T2A} } { }
\DeclareMicrotypeSet{t2atext}{encoding=T2A}
\UseMicrotypeSet{t2atext}

\usetikzlibrary{arrows,shapes,positioning,shadows,trees}

\tikzset{
	basic/.style  = {draw, text width=2cm, drop shadow, font=\sffamily, rectangle},
	root/.style   = {basic, rounded corners=2pt, thin, align=center, fill=white},
	level 2/.style = {basic, rounded corners=6pt, thin,align=center, fill=white, text width=6em},
	level 3/.style = {basic, thin, align=left, fill=white, text width=6em}
}

\usepackage{todonotes}

\papertype{Review}

\title{A Review of Keyphrase Extraction}

\abbrevs{-}

\author[1]{Eirini Papagiannopoulou}
\author[1]{Grigorios Tsoumakas}

\affil[1]{School of Informatics, Aristotle University of Thessaloniki, Thessaloniki, 54124, Greece}

\corraddress{Eirini Papagiannopoulou, School of Informatics, Aristotle University of Thessaloniki, Thessaloniki, 54124, Greece}
\corremail{epapagia@csd.auth.gr}

\fundinginfo{This work was partially funded by Atypon Systems, LLC (https://www.atypon.com/), Aristotle University of Thessaloniki, GrantID: 94349.}

\runningauthor{Eirini Papagiannopoulou and Grigorios Tsoumakas}

\begin{document}
	
	\renewcommand{\refname}{References}
	
	\taburulecolor{black}
	
	\maketitle
	
	\begin{abstract}
		Keyphrase extraction is a textual information processing task concerned with the automatic extraction of representative and characteristic phrases from a document that express all the key aspects of its content. Keyphrases constitute a succinct conceptual summary of a document, which is very useful in digital information management systems for semantic indexing, faceted search, document clustering and classification. This article introduces keyphrase extraction, provides a well-structured review of the existing work, offers interesting insights on the different evaluation approaches, highlights open issues and presents a comparative experimental study of popular unsupervised techniques on five datasets. 
	\end{abstract}
	
	
	\section{Introduction}
	\label{intro}
	
	Keyphrase extraction is concerned with automatically extracting a set of representative phrases from a document that concisely summarize its content \citep{hasan+ng2014}. There exist both supervised and unsupervised keyphrase extraction methods. Unsupervised methods are popular because they are \textit{domain independent} and do not need \textit{labeled training data}, i.e. manual extraction of the keyphrases, which comes with subjectivity issues as well as significant investment in time and money. Supervised methods on the other hand, have more powerful modeling capabilities and typically achieve higher accuracy than the unsupervised ones according to previous studies \citep{DBLP:journals/lre/KimMKB13, caragea2014citation, meng2017deep}.
	
	The versatility of keyphrases renders keyphrase extraction a very important document processing task. Keyphrases can be used for semantically indexing a collection of documents either in place of their full-text or in addition to it, enabling semantic and faceted search \citep{gutwin1999improving}. In addition, they can be used for query expansion in the context of pseudo-relevance feedback \citep{DBLP:conf/jcdl/SongSAO06}. They can also serve as features for document clustering and classification \citep{DBLP:conf/acl/HulthM06}. Furthermore, the set of extracted keyphrases can be viewed as an extreme summary of the corresponding document for human inspection, while the individual keyphrases can guide the extraction of sentences in automatic document summarization systems \citep{ZhangZM04}. Keyphrase extraction is particularly important in the (academic) publishing industry for carrying out a number of important tasks, such as the recommendation of new articles or books to customers, highlighting missing citations to authors, identifying potential reviewers for submissions and the analysis of content trends \citep{augenstein2017semeval}.

	
	There exists a number of noteworthy keyphrase extraction surveys. \cite{hasan+ng2014} focus on the errors that are made by state-of-the-art keyphrase extractors: (a) \textit{evaluation errors} (when a returned keyphrase is semantically equivalent to a gold one but it is evaluated as erroneous), (b) \textit{redundancy errors} (when a method returns correct but semantically equivalent keyphrases), (c) \textit{infrequency errors} (when a keyphrase appears one or two times in a text and the method fails to detect it), and (d) \textit{overgeneration errors} (when a system correctly returns a phrase as a keyphrase because it contains a word that appears frequently in the document, but erroneously outputs additional phrases as keyphrases that contain this frequent word). Despite that their analysis is not based on a large number of documents, it is quite interesting and well-presented. An earlier survey by the same authors presents the results of an experimental study of state-of-the-art unsupervised keyphrase extraction methods, conducted with the aim of gaining deeper insights into these methods \citep{hasan+ng2010}. The main conclusions are the following: (a) methods should be evaluated on multiple datasets, (b) post-processing steps (e.g., phrase formation) have a large impact on the performance of methods, and, (c) TfIdf is a strong baseline. \cite{DBLP:conf/aclnut/BoudinMC16} study the effect of document pre-processing pipelines to the keyphrase extraction process, while \cite{DBLP:conf/ecir/FlorescuC17} examine how keyphrase extraction is affected by phrase ranking schemes.
	
	Our article constitutes a contemporary review of the keyprase extraction task, containing the following main contributions:
	\begin{itemize}
	    \item A systematic presentation of both unsupervised (Section \ref{unsupervised}) and supervised (Section \ref{supervised}) keyphrase extraction methods via comprehensive categorization schemes based on the main properties of these methods. Our article reviews 37 additional methods compared to \cite{hasan+ng2014}. In addition, we contribute a time line of unsupervised and supervised methods to shed light on their evolution, as well as a presentation of the main types of features employed in supervised methods, along with a discussion of the issue of class imbalance.    
	    \item We present the different approaches that can be followed for evaluating keyphrase extraction methods, as well as the different evaluation measures that exist, along with their popularity in the literature (Section \ref{evaluation}). 
	    \item We provide a list of popular keyphrase extraction datasets, including their sources and properties, as well as a comprehensive catalogue of commercial APIs and free software (Section \ref{data-comp-software}) related to keyphrase extraction.  
	    \item We present a thorough empirical study, both quantitative and qualitative, among commercial APIs and state-of-the-art unsupervised methods, which allows to gain a deeper understanding of how the results are affected by different evaluation approaches, evaluation measures and ground truth standards (Section \ref{comparative-eval}). 
	\end{itemize}
	
	The article search strategy that we followed, involved searching for ``keyphrase extraction'' in the following databases of scientific literature: Google Scholar, Springer Link, IEEE Xplore, ACM Digital Library and DBLP. We focused mainly on articles appearing at the high quality journals and conference proceedings that are given in Appendix \ref{sources}.
	

	
	

	
	

	\section{Unsupervised Methods}
	\label{unsupervised}
	The basic steps of an unsupervised keyphrase extraction system are the following \citep{hasan+ng2010,hasan+ng2014}:

	\begin{enumerate}
		\item Selection of the candidate lexical units based on some heuristics. Examples of such heuristics are the exclusion of stopwords and the selection of words that belong to a specific part-of-speech (POS).
		\item Ranking of the candidate lexical units. 
		\item Formation of the keyphrases by selecting words from the top-ranked ones or by selecting a phrase with a high rank score or whose parts have a high score.
	\end{enumerate}

	\begin{figure}[H]
		\centering
		\includegraphics[width=1.0\linewidth]{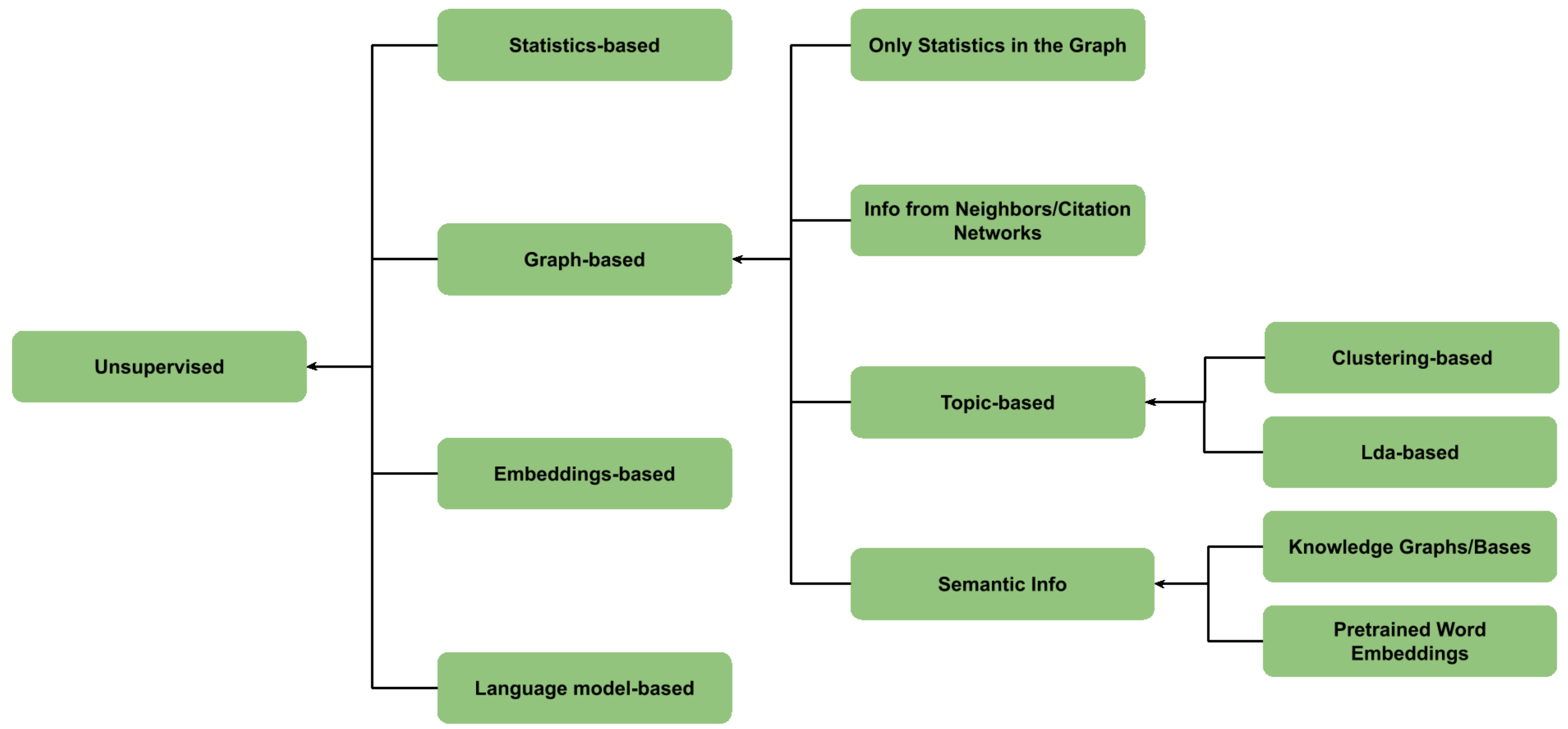}
		\caption{Presentation structure of the unsupervised keyphrase extraction methods.}
		\label{fig:general_taxonomy_unsupervised}
	\end{figure}
	
 Table \ref{tbl:unsupervisedCategorization1} presents the time line of the main research approaches related to unsupervised keyphrase extraction that are cited in this survey. We present the methods discussed below, along with their key-characteristics that support our adopted presentation structure in the following sections. Each method can be characterized as (i) statistics-based (Stat.), (ii) graph-based that incorporates statistics (Stats into Graph), (iii) topic-based (labels in gray color) that uses clustering (Clust.), LDA or knowledge graphs (KG) to find the document's topics, (iv) methods that use citation networks or neighbors' information (C/N Info) as well as (v) semantics (Sem.), and (vi) language model-based methods (Lang. Mod.). According to Table \ref{tbl:unsupervisedCategorization1}, graph-based methods are the most popular ones. However, the statistics-based methods still hold the attention of the research community. Additionally, the incorporation of semantics seems to be helpful for the task as more and more methods use them.	Figure \ref{fig:general_taxonomy_unsupervised} shows the presentation structure of the section. First, we present the statistics-based (Section \ref{sec:statistical_appr}) and the graph-based ranking methods (Section \ref{sec:graph_appr}). Then, we discuss the methods that are based on embeddings (Section \ref{embeddings_appr}) as well as the category of language model-based methods (Section \ref{sec:language_appr}).

	\begin{table}[H]
		\scalebox{0.95}{
			\renewcommand\arraystretch{1.4}\arrayrulecolor{black}
			\centering
			
			\begin{tabular}{p{0.7cm}<{\hskip 1pt} !{\foooo} >{\raggedright\arraybackslash}p{3.5cm} <{\hskip 1pt} !{\foooo} >{\raggedright\arraybackslash}p{0.7cm} p{0.7cm} p{0.7cm} p{0.7cm} p{0.7cm} p{0.7cm} p{0.7cm} p{0.7cm}}
				\hiderowcolors
				\toprule
				\addlinespace[1.5ex]
				\textbf{Year} & \textbf{Methods} & Stat. & Stats into  Graph & \cellcolor{gray!10} Clust. & \cellcolor{gray!10}LDA & \cellcolor{gray!10}KG & C/N Info & Sem. & Lang. Mod. \\ \hline
				2003 & {\small \textcolor{red}{\cite{tomokiyo2003language}}} & & & & & & & & \checkmark \\
				2004 &  {\small \textcolor{red}{\cite{mihalcea+tatau2004}}} & & \checkmark & & & &  & &   \\
				2008  & {\small \textcolor{red}{\cite{wan+xiao2008}  - SingleRank}}  \newline
				{\small \textcolor{red}{\cite{wan+xiao2008} - ExpandRank}} & & \checkmark \newline \newline \checkmark& & & &  \vspace{0.48cm} \checkmark & &    \\
				2009 & {\small \textcolor{red}{\cite{liu2009clustering}}} \newline {\small \textcolor{red}{\cite{DBLP:journals/is/El-BeltagyR09}}}  & \checkmark \newline \checkmark & & \checkmark & & & & &    \\
				2010 &  {\small \textcolor{red}{\cite{liu2010automatic}}} \newline  {\small \textcolor{red}{\cite{rose2010automatic}}} & & \checkmark \newline \checkmark & & \checkmark & & & &    \\
				2013 & {\small \textcolor{red}{\cite{bougouin2013topicrank}}} & & \checkmark & \checkmark & & & & &    \\
				2014 &  
				{\small \textcolor{red}{\cite{gollapalli2014extracting}}}  \newline  {\small \textcolor{red}{\cite{Wang2014}}} & & \checkmark \newline \checkmark & & & & \checkmark &  \vspace{0.02cm} \checkmark &    \\
				2015 & {\small \textcolor{red}{\cite{DBLP:conf/www/SterckxDDD15}}} \newline {\small \textcolor{red}{\cite{DBLP:conf/www/SterckxDDD15a}}}  \newline 
				{\small \textcolor{red}{\cite{DBLP:conf/starsem/DaneshSM15}}} \newline {\small \textcolor{red}{\cite{DBLP:conf/adc/WangLM15}}} & & \checkmark \newline \checkmark \newline \checkmark \newline \checkmark & & \checkmark \newline \checkmark & & & \vspace{0.92cm} \checkmark &   \\
				2017 & {\small \textcolor{red}{\cite{DBLP:conf/acl/TenevaC17}}} \newline  {\small \textcolor{red}{\cite{DBLP:conf/acl/FlorescuC17}}} \newline {\small \textcolor{red}{\cite{DBLP:journals/dase/ShiZYCZ17}}} & & \checkmark \newline \checkmark \newline \checkmark &  \vspace{0.48cm} \checkmark & \checkmark & & & \vspace{0.48cm} \checkmark&    \\
				2018 &  {\small \textcolor{red}{\cite{CamposSpringer}}} \newline  {\small \textcolor{red}{\cite{Boudin18Multipartite}}} \newline {\small \textcolor{red}{\cite{DBLP:journals/corr/abs-1801-04470}}} \newline {\small \textcolor{red}{\cite{papagiannopoulou2018local}}} \newline {\small \textcolor{red}{\cite{key2vec2018Mahata}}} \newline {\small \textcolor{red}{\cite{wikirank2018yu}}} & \checkmark \newline \newline \newline \checkmark  &  \vspace{0.02cm} \checkmark \newline \newline \newline \newline \checkmark \newline \checkmark & \vspace{0.02cm} \checkmark &   & \vspace{2.3cm} \checkmark & & \vspace{0.48cm} \checkmark \newline  \checkmark \newline \newline \checkmark \newline \checkmark &    \\
				2019 &  {\small \textcolor{red}{\cite{wonautomatic}}} & \checkmark &  &  &   &  & &  &    \\ \hline
		\end{tabular}}
		\caption{A time line of all the unsupervised keyphrase extraction methods discussed in this review, along with their key-characteristics.}
		\label{tbl:unsupervisedCategorization1}
	\end{table}

	\subsection{Statistics-based Methods}
	\label{sec:statistical_appr}
	
	\textit{TfIdf} is the common baseline on the task. This method scores and ranks the phrases according to the formula:
	$$\textit{TfIdf} = \textit{Tf} \times \textit{Idf}$$
	where \textit{Tf} is the raw phrase frequency and $\textit{Idf}=\textit{log}_2{\frac{N}{1+|{d \in \textit{D: phrase} \in d}|}}$, with \textit{N} the number of documents in the document set \textit{D} and $|{d \in \textit{D: phrase} \in d}|$ the number of documents in which the phrase appears. Effective variations of TfIdf are also implemented, such as taking the logarithm of the phrase frequency instead of the raw frequency to saturate the increase for high frequency words. Phrase frequency also contributes to alternative scoring schemes such as the one recently proposed by \cite{DBLP:conf/ecir/FlorescuC17}. Specifically, the score of the phrase is calculated as:
	$$\textit{mean} \times \textit{Tf}$$
	where \textit{mean} corresponds to the mean of the words' scores which constitute the phrase and \textit{Tf} is the phrase frequency in the text document. This is not a free-standing scoring scheme, but it can be considered as an intermediate stage of an unsupervised method.
	
	\textit{KP-Miner} \citep{DBLP:journals/is/El-BeltagyR09} is a keyphrase extraction system that exploits various types of statistical information beyond the \textit{Tf} and \textit{Idf} scores. It follows a quite effective filtering process of candidate phrases and uses a scoring function similar to TfIdf. Particularly, the system keeps as candidates those that are not be separated by punctuation marks/stopwords, considering at the same time the \textit{least allowable seen frequency (lasf)} factor and a cutoff constant \textit{(CutOff)} that is defined in terms of a number of words after which a phrase appears for the first time. Then, the system ranks the candidate phrases taking into account the \textit{Tf} and \textit{Idf} scores as well as the term position and a boosting factor for compound terms over the single terms.

	Around the same time, co-occurrence statistics and statistical metrics based on external resources started to be used for \textit{semantic} similarity calculation among the document's candidate terms. \textit{KeyCluster} \citep{liu2009clustering} tries to extract keyphrases that cover all the major topics of a document. First, it removes stopwords and selects the candidate terms. Then, it utilizes a variety of measures (co-occurrence-based or Wikipedia-based) to calculate the semantic relatedness of the candidate terms and groups them into clusters (using spectral clustering). Finally, it finds the exemplar terms of each cluster in order to extract keyphrases from the document. 
	
	The great importance of using both statistics and contexts info is confirmed by recent methods such as YAKE \citep{CamposSpringer} and the method proposed by \cite{wonautomatic}. YAKE, besides the term's position/frequency, also uses new statistical metrics that capture context information and the spread of the terms into the document. First, YAKE preprocesses the text by splitting it into individual terms. Second, a set of 5 features is calculated for each individual term: Casing ($W_{\textit{case}}$ that reflects the casing aspect of a word), Word Positional ($W_{\textit{Position}}$ that values more those words occurring at the beginning of a document), Word Frequency ($W_{\textit{Freq}}$), Word Relatedness to Context ($W_{\textit{Rel}}$ that computes the number of different terms that occur to the left/right side of the candidate word), and Word DifSentence ($W_{\textit{DifSentence}}$ quantifies how often a candidate word appears within different sentences). Then, all these features are used for the computation of the \textit{S(w)} score for each term (the smaller the value, the more important the word \textit{w} would be).
	$$S(w)=\frac{W_{\textit{Rel}}*W_{\textit{Position}}}{W_\textit{{case}}+\frac{W_{\textit{Freq}}}{W_{\textit{Rel}}}+\frac{W_{\textit{DifSentence}}}{W_{\textit{Rel}}}}$$
	Finally, a contiguous sequence of 1, 2 and 3-gram candidate keywords is generated using a sliding window of 3-grams. For each candidate keyword \textit{kw} the following score is assigned:
	$$S(\textit{kw})=\frac{\prod_{w \in \textit{kw}}\textit{S(w)}}{\textit{Tf(kw)}*(1+\sum_{w \in \textit{kw}}\textit{S(w)})}$$
	The smaller the score the more meaningful the keyword will be. In addition, the method of \cite{wonautomatic} shows that using a combination of simple textual statistical features is possible to achieve results that compete with state-of-the-art methods. The first step of this method is the selection of the candidate phrases using morphosyntactic patterns. Then, for each candidate the following features are calculated: \textit{Term Frequency}, i.e., the sum of each word
	frequency of the candidate phrase, \textit{Inverse Document Frequency (Idf)}, \textit{Relative First Occurrence}, i.e., the cumulative probability of the type $(1-a)^k$ where $a \in [0, 1]$
	measures the position of the first occurrence and $k$ the candidate frequency, and,  \textit{Length}, i.e., a simple rule that scores 1 for unigrams and 2 for the remaining sizes. The final score of each candidate is the result of the product of these 4 features. Moreover, with respect to the observation that the larger document datasets are associated with a higher number of keyphrases per document, the top-$N$ candidates are extracted for each document with $n = 2.5\times \textit{log}_{10}(\textit{doc size})$ (the $2.5$ parameter was found experimentally).

	\subsection{Graph-based Ranking Methods}
	\label{sec:graph_appr}
	
	The basic idea in graph-based ranking is to create a graph from a document that has as nodes the candidate phrases from the document and each edge of this graph connects related candidate keyphrases. The final goal is the ranking of the nodes using a graph-based ranking method, such as Google's PageRank \citep{grin+page1998}, Positional Function \citep{herings+van+talman} and HITS \citep{kleinberg1999} or generally solving a proposed optimization problem on the graph. 
	Based on the methods described below, PageRank has been successfully used for graph-based
	keyphrase extraction. PageRank is based on eigenvector centrality and recursively defines the weight of a vertex as a measure of its influence inside the graph-of-words, regardless of how cohesive its neighborhood is. However, \cite{DBLP:conf/ecir/RousseauV15} consider the vertices of the main core as the set of keywords to extract from the document, as it corresponds to the most cohesive connected component of the graph. For this reason, the vertices are intuitively appropriate candidates.

	\textit{TextRank} was the first graph-based keyphrase extraction method proposed by \cite{mihalcea+tatau2004} which inspired researchers to build upon it, leading to well-known state-of-the-art methods. First of all, the text is tokenized, and annotated with POS tags. Then, syntactic filters are applied to the text units, i.e., only nouns and adjectives are kept. Next, the lexical units that pass the filters mentioned above, i.e., the candidates, are added to the graph as nodes and an edge is added between the nodes that co-occur within a window of $M$ words. The graph is undirected and unweighted. The initial score assigned to each node is equal to 1 and then the PageRank algorithm runs until it converges. Specifically, for a node $V_i$ the corresponding score function that is repeatedly computed  is:
	$$S(V_i) = (1-\lambda)+\lambda*\sum_{j \in N(V_i)}\frac{1}{N(V_j)}S(V_j)$$
	where $N(V_i)$ is the set of neighbors of $V_i$, $N(V_j)$ is the set of neighbors of $V_j$ and $\lambda$ is the probability of jumping from one node to another. Once the algorithm converges, the nodes are sorted by decreasing score.
	
	\textit{SingleRank} \citep{wan+xiao2008} is an extension of TextRank which incorporates weights to edges. Similarly, to the statistical-based methods, the co-occurrence statistics are crucial information regarding the contexts. Hence, each edge weight is equal to the number of co-occurrences of the two corresponding words. Then, the score function for a node $V_i$ is computed in a similar way:
	$$\textit{WS}(V_i) = (1-\lambda)+\lambda*\sum_{j \in N(V_i)}\frac{\#_c{ij}}{\sum_{V_k \in N(V_j)}\#_c{jk}}\textit{WS}(V_j)$$
	where $\#_c{ij}$ is the the number of co-occurrences of word $i$ and word $j$. In a post-processing stage, for each continuous sequence of nouns and adjectives in the text document, the scores of the constituent words are summed up and the $T$ top-ranked candidates are returned as keyphrases. Co-occurrences are also used by various graph-based methods such as the one of \cite{rose2010automatic}, called \textit{RAKE (Rapid Automatic Keyword Extraction)}, that utilizes both word frequency and word degree to assign scores to phrases. RAKE take as input parameters a list of stopwords, a set of phrase delimiters and a set of word delimiters to partition the text into candidate phrases. Then, a graph of word-word co-occurrences is created and a score (word frequency or the word degree or the ratio of degree to frequency) is assigned for each candidate phrase which is the sum of the scores of the words that comprise the phrase. In addition, RAKE is able to identify keyphrases that contain interior stopwords, by detecting pairs of words that adjoin one another at least twice in the same document, in the same order. Finally, the top $T$ ranked candidate phrases are selected as keyphrases for the document.
	
	In this vein, the more recent methods \textit{SGRank} \citep{DBLP:conf/starsem/DaneshSM15} and \textit{PositionRank (PR)} \citep{DBLP:conf/acl/FlorescuC17} utilize statistical, positional, and, word co-occurrence information, thus improving the overall performance. In particular, \textit{SGRank} \citep{DBLP:conf/starsem/DaneshSM15}, first, extracts all possible n-grams from the input text, eliminating those that contain punctuation marks or whose words are anything different than noun, adjective or verb. Furthermore, it takes into account term frequency conditions. In the second stage, the candidate n-grams are ranked based on a modified version of TfIdf (similar to KP-Miner). In the third stage, the top ranking candidates are re-ranked based on additional statistical heuristics, such as position of first occurrence and term length. Finally, the ranking produced in stage three is incorporated into a graph-based algorithm which produces the final ranking of keyphrase candidates. \textit{PositionRank (PR)} \citep{DBLP:conf/acl/FlorescuC17} is a graph-based unsupervised method that tries to capture frequent phrases considering the word-word co-occurrences and their corresponding position in the text. Specifically, it incorporates all positions of a word into a biased weighted PageRank. Finally, the keyphrases are scored and ranked. 
	
	The rest of the graph-based methods are grouped into three main categories, i.e., the methods that incorporate information from similar documents or citation networks (Section \ref{sec:graph_citation_appr}), the topic-based methods (Section \ref{sec:graph_topic_appr}), and the graph-based methods that utilize semantics (Section \ref{sec:graph_semantics_appr}), which are discussed in the following sections. 
	

	\subsubsection{Incorporating Information from Similar Documents/Citation Networks}
	\label{sec:graph_citation_appr}

	The graph-based methods discussed earlier assume that the documents are independent of each other. Hence, only the information included in the target document, i.e., the phrase’s \text{TfIdf}, position etc., is used during the keyphrase extraction process. However, related documents have mutual influences that help to extract keyphrases. \textit{ExpandRank} \citep{wan+xiao2008} is an extension of SingleRank that takes into consideration information from neighboring documents to the target document. It constructs an appropriate \textit{knowledge context} for the target document that is used in the keyphrase extraction process and helps to extract important keyphrases from it. According to this method each document is represented by a vector with TfIdf scores. For a target document $d_0$, its $k$ nearest neighbors are identified, and a larger document set $D$ of $\textit{k+1}$ documents is created. Based on this document set, a graph is constructed, where each node corresponds to a candidate word in $D$, and edges are added between two nodes, $(v_i,v_j)$, that co-occur within a window of $M$ words in the document set. The weight of an edge, $e(v_i,v_j)$ is computed as follows:
	$$e(v_i,v_j) = \sum_{d_k \in D}\textit{sim}(d_0,d_k) \times \textit{freq}_{d_k}(v_i,v_j)$$
	where $\textit{sim}(d_0,d_k)$ is  the  cosine  similarity  between $d_0$ and $d_k$, and
	$\textit{freq}_{d_k}(v_i,v_j)$ is  the  co-occurrence frequency of $v_i,v_j$ in document $d_k$. Once the graph is constructed, the rest of the procedure is identical to SingleRank.

	A more related knowledge context to the target document can be also found via citation networks. In a citation network, information flows from one paper to another via the citation relation. In other words, the influence
	of one paper on another is captured through \textit{citation contexts} (i.e., short text segments surrounding a paper's mention). In this vein, \textit{CiteTextRank} \citep{gollapalli2014extracting} incorporates information from citation networks for the keyphrase extraction process capturing the information from such citation contexts. In particular, given a target document $d$ and a citation network $C$ ($d \in C$), a \textit{cited context} for $d$ is a context in which $d$ is cited by a paper $d_i$, and a \textit{citing context} for $d$ is a context in which $d$ is citing another paper $d_j$. The content of $d$ is dubbed as \textit{global context}. As a first step, they construct an undirected graph $G = (V, E)$ for $d$, with nodes the words from all types of contexts of $d$ and edges between the nodes ($v_i,v_j$), in case of co-occurrence within a window of $M$ continuous tokens in any of the contexts. The weight of an edge is set equal to:
	$$e_{\textit{ij}}=\sum_{t \in TC} \sum_{c \in C_t} \lambda_t \cdot \textit{sim}(c,d) \cdot \#_c(v_i,v_j)$$ where the $\textit{TC}$ are the available types of contexts in $d$ (global, citing, cited), $\textit{sim}(c,d)$ is the cosine similarity between the \textit{TfIdf} vectors of any context $c$ of $d$ and $d$, $\#_c(v_i,v_j)$ is the number of co-occurrences of $v_i,v_j$ in context $c$, $C_t$ is the set of contexts of type $t \in \textit{TC}$ and $\lambda_t$ is the weight for contexts of type $t$. Finally, they score the vertices using the PageRank algorithm.

	\subsubsection{Topic-based Methods}
	\label{sec:graph_topic_appr}
	
	Except the previous methods that use classic statistical heuristics (\textit{Tf}, \textit{Idf}, position) as well as context-aware statistics such as word-word co-occurrence information, there are also methods that try to return keyphrases related to the \textit{topics} discussed in the document. Specifically, the topic-based methods try to extract keyphrases that are representative for a text document in terms of the topics it covers. Such methods usually apply clustering techniques or Latent Dirichlet Allocation (LDA) \citep{blei2003latent} to detect the main topics discussed. 
	
	\noindent \textbf{Clustering-based Methods}
	
	\textit{TopicRank (TR)} \citep{bougouin2013topicrank}, first, preprocesses the text to extract the candidate phrases. Then, the candidate phrases are grouped into separate topics using hierarchical agglomerative clustering. In the next stage, a graph of topics is constructed whose edges are
	weighted based on a measure that considers phrases' offset positions in the text. Then, TextRank is used to rank the topics and one keyphrase candidate is selected from each of the N most important topics (first occurring keyphrase candidate). A more recent very similar method to TopicRank but more advanced is \textit{MultipartiteRank (MR)} \citep{Boudin18Multipartite} which introduces an in-between step where edge weights are adjusted to capture position information giving bias towards keyphrase candidates occurring earlier in the document. Note that the heuristic to promote a specific group of candidates, e.g., those that appear earlier in the text, can be adapted to satisfy other conditions/needs.

	\noindent \textbf{LDA-based Methods}
	
	\textit{Topical PageRank (TPR)} \citep{liu2010automatic} is a topic-based method upon which various topic-based methods have been built. TPR uses LDA to obtain the topic distribution \textit{pr(z|w)} of each word $w$, for topic $z \in K$, where $K$ is the number of topics and the topic distribution \textit{pr(z|d)} of a new document $d$, for each topic $z \in K$. Then, for a document $d$, it constructs a word graph based on word-word co-occurrences by adding only the adjectives and nouns. The idea of TPR is to run a \textit{Biased PageRank} for each topic separately. So, for every topic $z$, the topic-specific PageRank word scores are calculated as follows:
	$$S_z(w_i) = \lambda \sum_{j:w_j \rightarrow w_i} \frac{e(w_j,w_i)}{O(w_j)}S_z(w_j)+(1-\lambda)p_z(w_i)$$ where $p_z(w_i)$ is equal to $pr(z|w)$ which is the probability of topic $z$ given word $w$, $\lambda$ is a damping factor range from $0$ to $1$, $e(w_j,w_i)$ is the weight of link $e(w_j,w_i)$, and $O(w_j)$ is the out-degree of vertex $w_j$. The final PageRank word scores are obtained by the equation given above, iteratively, until convergence. Using the topic-specific importance scores of words, the ranking of candidate phrases with respect to each topic $z$ separately is:
	$$S_z(p)=\sum_{w_i \in p}S_z(w_i)$$ where $p$ is a candidate phrase. Finally, the topic distribution of the document $p(z|d)$ is considered by integrating the topic-based rankings of candidates: 
	$$S(p)=\sum_1^K S_z(p) \times \textit{pr}(z|d)$$
	
	\textit{Single Topical PageRank (Single TPR)} \citep{DBLP:conf/www/SterckxDDD15} is an alternative method to avoid the large cost of TPR by running only one PageRank for each document. According to this method, the sum of the $K$ topic-specific values of each word $w_i$ is replaced by the concept of topical importance $W(w_i)$. Particularly, first, the cosine similarity ($W(w_i)$) is calculated between the vector of word-topic probabilities and the document-topic probabilities. Finally, the Single PageRank $S(w_i)$ becomes
	$$S(w_i) = \lambda \sum_{j:w_j \rightarrow w_i} \frac{e(w_j,w_i)}{O(w_j)}S(w_j)+(1-\lambda)\frac{W(w_i)}{\sum_{w \in V}W(w)}$$ where $V=\{w_1, \dots, w_N\}$ is the set of graph nodes. Moreover, in a related study, \cite{DBLP:conf/www/SterckxDDD15a} propose the utilization of \textit{multiple topic models} for the keyphrase extraction task to show the benefit from a combination of models. Particularly, the models disagree when they are trained on different corpora, as there is a difference in contexts between the corpora. This leads to different topic models and disagreement about the word importance. So, this disagreement is leveraged by computing a combined topical word importance value which is used as a weight in a Topical PageRank, improving the performance, in cases where the topic models differ substantially.
	
	In this spirit, \textit{Salience Rank} \citep{DBLP:conf/acl/TenevaC17} is quite close to Single TPR as it runs only once PageRank, incorporating in it a word metric called \textit{word salience} $S_{\alpha}(w)$, which is a linear combination of the \textit{topic specificity} \citep{DBLP:conf/avi/ChuangMH12} and \textit{corpus specificity} of a word (the last can be calculated counting word frequencies in a specific corpus). Intuitively, topic specificity measures how much a word is shared across topics (the less the word is shared across topics,
	the higher its topic specificity). Users can balance topic specificity and corpus specificity of the extracted keyphrases and can tune the results according to particular cases. So, the Single PageRank becomes:
	$$S(w_i) = \lambda \sum_{j:w_j \rightarrow w_i} \frac{e(w_j,w_i)}{O(w_j)}S(w_j)+(1-\lambda)\times S_{\alpha}(w_i)$$
	
	An interesting and probably effective direction in LDA-based methods would be the utilization of a topic model similar to the one proposed by \cite{DBLP:journals/pvldb/El-KishkySWVH14}. Indeed, most topic modeling algorithms model text corpora with unigrams, whereas human interpretation often relies on inherent grouping of terms into phrases. Particularly, \cite{DBLP:journals/pvldb/El-KishkySWVH14} first propose an efficient phrase mining technique to extract frequent significant phrases and segment the text
	at the same time, which uses frequent phrase mining and a statistical significance measure. Then, they introduce a simple but effective topic model that restricts all constituent terms within a phrase to share the same latent topic, and assigns the phrase to the topic of its constituent words. Besides, the first part of the method (phrase mining) could be exploited to filter out false candidate keyphrases in the context of a keyphrase extraction pipeline.

	\subsubsection{Graph-based Methods with Semantics}
	\label{sec:graph_semantics_appr}
	
	\textbf{Semantics from Knowledge Graphs/Bases}
	
	The main problems of the topic-based methods are that the topics are too general and vague. In addition, the co-occurrence-based methods suffer from \textit{information loss}, i.e., if two words never co-occur within a window size in a document, there will be no edges to connect them in the corresponding graph-of-words even though they are semantically related, whereas the statistics-based methods suffer from \textit{information overload}, i.e., the real meanings of words in the document may be overwhelmed by the large amount of external texts used for the computation of statistical information. To deal with such problems and incorporate semantics for keyphrase extraction, \cite{DBLP:journals/dase/ShiZYCZ17} propose a keyphrase extraction system that uses knowledge graphs. First, nouns and named entities (\textit{keyterms}) are selected and grouped based on semantic similarity by applying clustering. Then, the keyterms of each cluster are connected to entities of DBpedia. For each cluster, the relations between the keyterms are detected by extracting the \textit{h-hop keyterm graph} from the knowledge graph, i.e., the subgraph of DBpedia that includes all paths of length no longer than $h$ between two different nodes of the cluster. Then, all the extracted keyterm graphs of the clusters are integrated into one and a Personalized PageRank (PPR) \citep{DBLP:conf/www/Haveliwala02} is applied on it to get the ranking score of each keyterm. The final ranking scheme of the candidate phrases uses the PPR score which is the sum of the PPR scores of the keyterms in it, as well as, the frequency and first occurrence position of the phrase. Moreover, this method could be categorized to the topic-based methods using clustering (see Table \ref{tbl:unsupervisedCategorization1}).  
	
	Similarly, \cite{wikirank2018yu} propose WikiRank, an unsupervised automatic keyphrase extraction method that tries to link semantic meaning to text. First, they use TAGME \citep{WikiRankFerragina2010}, which is a tool for topic/concept annotation that detects meaningful text phrases and matches them to a relevant Wikipedia page. Additionally, they extract noun groups whose pattern is zero or more adjectives followed by one or more nouns as candidate keyphrases. Then, a semantic graph is built whose vertices are the union of the concept set and the candidate keyphrase set. In case the candidate keyphrase contains a concept according to the annotation of TAGME an edge is added between the corresponding nodes. The weight of a concept is equal to the frequency of the concept in the full-text document. Moreover, they propose the score of a concept $c$ in a subgraph of $G$ to be:
	$$S(c) = \sum_{i=0}^{\textit{deg}(c)}\frac{w_c}{2^i}$$
	where $w_c$ is the weight of the concept $c$, and $deg(c)$ is the degree of $c$ in the corresponding subgraph. The goal is to find the candidate keyphrase set $\Omega$ with the best coverage, i.e., the maximum sum of scores of the concepts annotated from the phrases in $\Omega$.

	\noindent \textbf{Semantics from Pretrained Word Embeddings}
	
	Although the methods that utilize semantics from knowledge graphs/bases have shown their improvements, the keyphrase extraction process requires more background knowledge than just semantic relation information. Thus, \cite{Wang2014} propose a graph-based ranking model that considers semantic information coming from distributed word representations as background knowledge. Again, a graph of words is created with edges that represent the co-existence between the words within a window of \textit{M} consecutive words. Then, a weight, called \textit{word attraction score} is assigned to every edge, which is the product of two individual scores; a) the \textit{attraction force} between two words that uses the frequencies of the words and the euclidean distance between the corresponding word embeddings and b) the \textit{dice coefficient} \citep{dice1945measures, Stubbs2003} to measure the
	probability of two words co-occurring in a pattern, or by
	chance. Particularly, given a document $d$ as a sequence of words $w$, i.e., $d = {w_1,w_2,\dots,w_n}$ the dice coefficient is computed as:
	$$\textit{dice}(w_i,w_j) = \frac{2 \times \textit{freq}(w_i,w_j)}{\textit{freq}(w_i)+\textit{freq}(w_j)}$$
	where $\textit{freq}(w_i,w_j)$ is the co-occurrence frequency of words
	$w_i$ and $w_j$ , and $\textit{freq}(w_i)$ and $\textit{freq}(w_j)$ are the occurrence
	frequencies of $w_i$ and $w_j$ in $d$.
	Once more, a weighted PageRank algorithm is utilized to rank the words. \cite{DBLP:conf/adc/WangLM15} propose an improved method that uses a \textit{Personalized} weighted PageRank model with pretrained word embeddings and also more effective edge weights. Particularly, the strength of relation of a pair of words is calculated as
	the product of the \textit{semantic relatedness} and \textit{local co-occurrence co-efficient}, as:
	$$\textit{sr}(w_i,w_j) = \textit{semantic}(w_i,w_j)\times \textit{cooccur}(w_i,w_j)$$
	where $\textit{semantic}(w_i,w_j)=\frac{1}{1-\textit{cosine}(w_i,w_j)}$ with $\textit{cosine}(w_i,w_j)$ to be the cosine similarity between the corresponding vectors. Furthermore, as co-occurrence co-efficient is used the Point-wise Mutual Information. Finally, the score of $w_i$ is calculated as follows: 
	$$S(w_i)=(1-\lambda)\textit{pr}(w_i)+\lambda \times \sum_{w_j in N(w_i)}\frac{\textit{sr}(w_i,w_j)}{|\textit{out}(w_j)|}S(w_j)$$
	where $\textit{sr}(w_i, w_j)$ is the strength of relatedness score between the two words calculated previously, $\textit{pr}(w_i)$ is the probability distribution of word
	$w_i$, calculated as $\frac{\textit{freq}(w_i)}{N}$ ($\textit{freq}(w_i)$ is the occurrence frequency and $N$ the number of total words), and $N(w_i)$ is the set of vertices incident to $w_i$ 
	
	However, \cite{DBLP:conf/adc/WangLM15} do not use domain-specific word embeddings and notice that training them might lead to improvements. This motivated \cite{key2vec2018Mahata} to present \textit{Key2Vec}, an unsupervised keyphrase extraction technique from scientific articles that represents candidate keyphrases of a document by domain specific phrase embeddings and ranks them using a \textit{theme-weighted} PageRank algorithm \citep{pagerankLangville2003}. After exhaustive text preprocessing on a corpus of scientific abstracts, which is well-described in their work, Fasttext \citep{DBLP:journals/tacl/BojanowskiGJM17} is utilized for training multiword phrase embeddings. First, the same text preprocessing is applied to the target document in order to get a set of unique \textit{candidate keyphrases}. Then, a \textit{theme excerpt}, i.e., the first sentence(s), is extracted from the document. Afterwards, a unique set of thematic phrases, i.e., named entities, noun phrases and unigram words, are also extracted from the theme excerpt. Next, they get the vector representation of each thematic phrase using the trained phrase embedding model and perform vector addition to get the final \textit{theme vector}. Moreover, the phrase embedding model is also used to get the vector representation for each candidate keyphrase. Then, they calculate the cosine distance between the theme vector and the vector for each candidate keyphrase, assigning a score to each candidate (\textit{thematic weight}). Then, a directed graph is constructed with the candidate keyphrases as the vertices. Two candidate keyphrases are connected if they co-occur within a window size of 5 (bidirectional edges are used). In addition, weights are calculated for the edges using the semantic similarity between the candidate keyphrases obtained from the phrase embedding model and their frequency of co-occurrence, as in \cite{DBLP:conf/adc/WangLM15}. For the final ranking of the candidate keyphrases, a weighted personalized PageRank algorithm is used.

	\subsection{Keyphrase Extraction based on Embeddings}
	\label{embeddings_appr}
	
	Many methods for representation of words have been proposed. Representative techniques, which are based on a co-occurrence matrix, are Latent Dirichlet Allocation (LDA) \citep{blei2003latent} and Latent Semantic Analysis (LSA) \citep{deerwester1990indexing}. However, word embeddings came to the foreground by \cite{DBLP:journals/corr/abs-1301-3781}, who presented the popular Continuous Bag-of-Words model (CBOW) and the Continuous Skip-gram model. Additionally, sentence embeddings (Doc2Vec \citep{DBLP:conf/rep4nlp/LauB16} or Sent2vec \citep{DBLP:conf/naacl/PagliardiniGJ18}) as well as the popular GloVe (Global Vectors) \citep{Pennington14glove:global} method are utilized by keyphrase extraction methods.
	
	\textit{EmbedRank}  \citep{DBLP:journals/corr/abs-1801-04470} extracts candidate phrases based on POS sequences (phrases that consist of zero or more adjectives followed by one or multiple nouns). EmbedRank uses sentence embeddings (Doc2Vec or Sent2vec) to represent both the candidate phrases and the document in the same high-dimensional vector space. Finally, the system ranks the candidate phrases using the cosine similarity between the embedding of the candidate phrase and the document embedding.
	
	Moreover, \cite{papagiannopoulou2018local} present the \textit{Reference Vector Algorithm (RVA)}, a method for keyphrase extraction, whose main innovation is the use of local word embeddings/semantics (in particular GloVe vectors), i.e., embeddings trained from the single document under consideration. The employed local training of GloVe on a single document and the graph-based family
	of methods can be considered as two alternative views of the same information source, as both methods utilize the statistics
	of word-word co-occurrence in a text. Then, the mean (\textit{reference}) vector of the words in the document's title and abstract is computed. This mean vector is a vector representation of the semantics of the whole publication. Finally, candidate keyphrases are extracted from the title and abstract, and are ranked in terms of their cosine similarity with the reference vector, assuming that the closer to the reference vector is a word vector, the more representative is the corresponding word for the publication.
	
	
	\subsection{Language Model-based Methods}
	\label{sec:language_appr}
	
	Language modeling plays an important role in natural language processing tasks \citep{DBLP:journals/csl/ChenG99}. Generally, an $N$-gram language model assigns a probability value to every sequence of words $\boldmath{w}=w_1w_2\dots w_n$, i.e., the probability $P(w)$ can be decomposed as
	$$P(w)=\prod_{i=1}^{n}P(w_i|w_1,w_2,\dots,w_{i-1})$$
	For example a trigram language model is the following:
	$$P(w)=\prod_{i=1}^{n}P(w_i|w_{i-1},w_{i-2})$$ 
	
	Besides the $N$-gram language models there are also various types of models, such as the popular neural language models that use neural networks to learn the context of the words \cite{DBLP:journals/corr/abs-1803-08240}.
	
	\textbf{Keyphrase extraction with N-gram language models}
	\cite{tomokiyo2003language} create both unigram and n-gram language models
	on a foreground corpus (target document) and a background corpus (document set). Their main idea is based on the fact that the loss between two language models can be measured using the Kullback-Leibler divergence. Particularly, at phrase level, for each phrase, the \textit{phraseness} is computed as the divergence between the unigram and n-gram language models on the foreground corpus and the \textit{informativeness} is calculated as the divergence between the n-gram language models on the foreground and the background corpus. Then, the phraseness and informativeness are summed as a final score for each phrase. Finally, phrases are sorted based on this score.

	\section{Supervised Methods}
	\label{supervised}
\begin{figure}[H]
		\centering
		\includegraphics[width=1.0\linewidth]{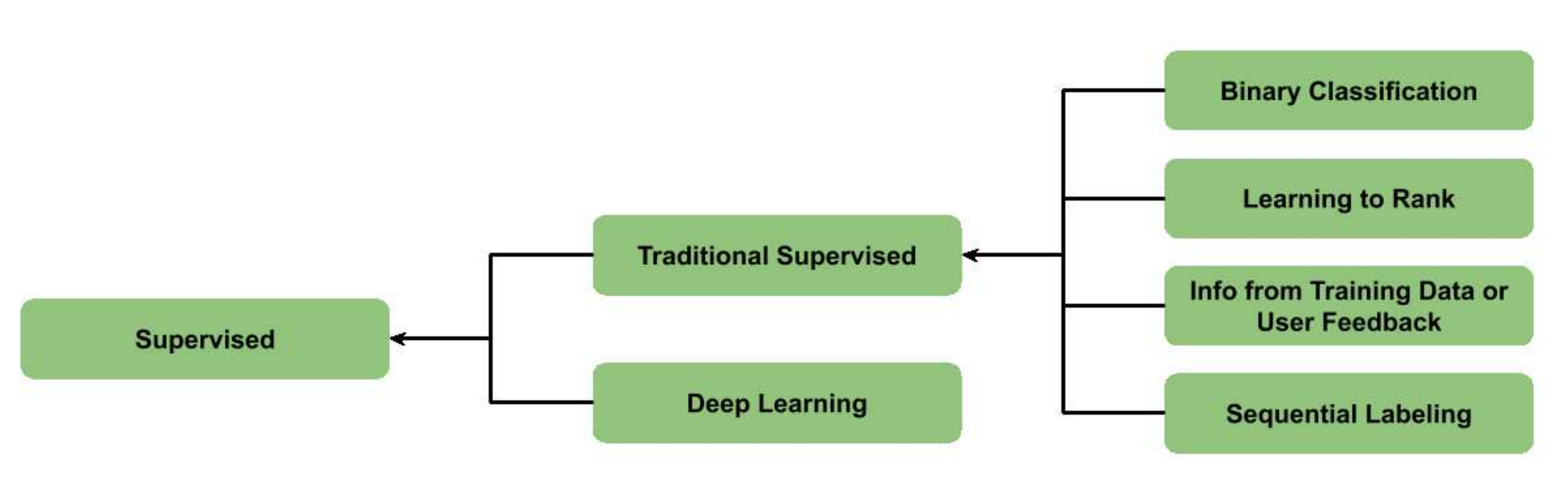}
		\caption{Presentation structure of supervised keyphrase extraction methods.}
		\label{fig:general_taxonomy_supervised}
	\end{figure}

	In this section, we present traditional supervised methods (Section \ref{sec:traditional_sup_methods}) as well as deep learning methods (Section \ref{sec:deep_learning_methods}) along with the main categories of features that are used (Section \ref{sec:types_features}). Section \ref{sec:exp_comparison_supervised} discusses the performance of some earlier, state-of-the-art and more recent supervised methods on 3 popular keyphrase extraction datasets. Finally, we discuss the main problems of the supervised learning methods along with the proposed solutions (Section \ref{imbalanced}). Figure \ref{fig:general_taxonomy_supervised} shows the presentation structure of the supervised keyphrase extraction methods that is also consistent with their corresponding taxonomy. 


	\subsection{Traditional Supervised Methods}
	\label{sec:traditional_sup_methods}
	
	\noindent \textbf{Keyphrase Extraction as Binary Classification Task} 
	
	In general, in supervised learning, a classifier is trained on annotated with keyphrases documents in order to determine whether a candidate phrase is a keyphrase or not. These keyphrases and non-keyphrases are used to generate positive and negative examples. In this section, we discuss the most representative methods of this category in terms of the learning process they employ.
	
	One of the first keyphrase extraction systems is \textit{KEA} \citep{witten1999kea} which calculates for each candidate phrase its TfIdf score and its first occurrence, i.e., the position (offset in words from the start of the document) of the phrase's first appearance, and uses as learning algorithm Naive Bayes. \cite{hulth2003improved} proposes a system that utilizes linguistic knowledge. Particularly, for each candidate phrase of the training set four features are calculated: the within-document frequency, the collection frequency, the relative position of the first occurrence and part-of-speech (POS) tag sequence. In this case the machine learning approach is a rule induction system with bagging. An extension of KEA is proposed by \cite{DBLP:conf/icadl/NguyenK07}. The existing feature set (TfIdf and first occurrence) is enhanced with additional features to incorporate position information (a section occurrence vector, i.e., a vector of frequency features for 14 generic section headers) and additional morphological/linguistic characteristics (POS tag sequence, suffix sequence and acronym status) of the keyphrases. Once more, as learning method they use Naive Bayes. Later, \cite{medelyan2009human} propose \textit{Maui}, which extends KEA by introducing an alternative set of new features. Maui uses as features in the classification model the TfIdf score, the first occurrence, the \textit{keyphraseness} which quantifies how often a candidate phrase appears as a tag in the training corpus, the phrase length (measured in words), the \textit{Wikipedia-based keyphraseness} which is the probability of a phrase being a link in the Wikipedia, the \textit{spread} of a phrase, i.e., the distance between its first and last occurrence of the phrase as well as additional features that utilize Wikipedia as a source of language usage statistics such as the \textit{node degree}, the \textit{semantic relatedness} and the \textit{inverse Wikipedia linkage}. Maui uses bagged decision trees as classifier to capture interactions between features. Afterwards, \cite{DBLP:conf/semeval/NguyenL10} propose an method (\textit{WINGNUS}) that extracts candidates using the regular expression rules used in \cite{DBLP:conf/mwe/KimK09}. In their work, they also study the keyphrase distribution on the training data over the sections of documents, concluding that the best choice is either the full-text or a text segment comprising of title, headers, abstract, introduction, related work, conclusion and the first sentence of each paragraph. They experimented with various combinations of features from a large feature set using the Naive Bayes algorithm. Finally, they propose as the best features the TfIdf, the term frequency of the phrase substrings, the first occurrence and the length of the phrase.
	
	\cite{caragea2014citation} propose \textit{CeKE}, a binary classification model (Naive Bayes classifier with decision threshold 0.9) that utilizes novel features from information of citation contexts (boolean features that are true if the candidate phrase occurs in cited/citing contexts and the TfIdf score of each phrase computed from the citation contexts), existing features from previous works (phrase TfIdf, first position, relative position and POS tags), and extended existing features (a boolean feature which is true if the TfIdf score is greater than a threshold and a boolean feature which is true if the distance of the first occurrence of a phrase from the beginning of a target paper is below some value). \cite{Wiki2008shi} use Wikipedia as an additional feature source for their supervised keyphrase extraction method. The first step is the detection of the candidate keyphrases via mapping to the Wikipedia concepts that appear in the titles of Wikipedia articles, redirect/disambiguation pages and anchor text of Wikipedia articles. As each Wikipedia concept is related to at least one category, candidate topics can be defined for a document by collecting the corresponding concepts' categories. Then, a semantic bipartite graph is constructed where candidate keyphrases are connected with the document topics based on various semantic relation types, e.g., the synonym and hypernym relation. After the graph construction, a feature weight ($w_e=\lambda^l \cdot f_p$, where $\lambda$ is a parameter that lies in (0,1) for the level of category $l$, and $f_p$ is the frequency of the phrase in the document) is assigned to the candidate keyphrases and a proposed variant of the HITS algorithm that considers the link weights is used to compute the final \textit{semantic feature weights} for each phrase. Finally, additional statistical, positional and linguistics features are computed, such as the appearance in the title, the phrase frequency, the phrase position and the phrase length, and a logistic regression model is used to predict the document's keyphrases.
	
	Afterwards, \cite{DBLP:conf/semeval/WangL17} propose a system (PCU-ICL) that incorporates features, such as stemmed unigrams, Tf, Idf, TfIdf, POS tags for every word in the phrase, phrase length, previous/next token of the phrase, POS of previous/next token of the phrase, distance between the phrase and the citation, a boolean feature that encodes whether the phrase is in IEEE taxonomy list, Wikipedia-based Idf, GloVe embedding of the phrase, and unsupervised model based features. An ensemble of unsupervised models, random forest (RF) and linear models are used for candidate keyphrase ranking. Recently, \cite{DBLP:conf/aaai/2018} have proposed \textit{SurfKE}, a supervised method that learns feature representations based on the document's graph of words. The Gaussian Naive Bayes classifier is used to train their model.

	\noindent \textbf{Keyphrase Extraction as Learning to Rank Task}
	
	Learning to rank approaches 
	learn a ranking function that sorts the candidate phrases based on their score of being a keyphrase, whereas classification methods have to make hard decisions.
	A representative method in this category is \textit{Ranking SVM} \citep{jiang2009ranking} which first constructs an appropriate training set by following a particular process. Suppose there is a training set of $L$ documents $D = {d_1, d_2, \dots, d_L}$ and for each document $k$ there are $N_k$ candidate keyphrases, each of them is represented by a feature vector $x_{\textit{ki}}$ (the $i$th candidate keyphrase of the $k$th document) and the corresponding rank $y_{\textit{ki}}$. In order to train a ranking function $f^*$, the training data is transformed into a set of ordered phrase pairs $(x_{\textit{ki}}-x_{\textit{kj}}, z_{k,\textit{ij}})$ where $z_{k,\textit{ij}}$ shows their preference relationships, $z_{k,\textit{ij}}=+1$ if $y_{k,i} \succ y_{k,j}$ and $z_{k,\textit{ij}}=-1$, otherwise. Then, an SVM model is trained on this training set that ends up to an optimization problem whose optimal solution are the $\omega$ weights, denoted as $\omega^*$ and the corresponding ranking function becomes: $$ f^*(x)= \langle \omega^*, x \rangle$$
	
	\textit{MIKE} \citep{DBLP:conf/cikm/ZhangCLG0X17} is a more advanced learning to rank method. First, it selects the candidate words by applying stopword and POS filters. Then, a graph of words is constructed based on the co-occurrence relations between the candidate words within the target document. Afterwards, features of candidate words and their co-occurrences (i.e., \textit{node features}, such as TfIdf, TfIdf over a certain threshold, first occurrence, relative position, POS tags, and citation TfIdf as well as \textit{edge features} which are based on the co-occurrence frequency between candidate word pairs in the word graph), \textit{topic distributions} of candidates and \textit{relative importance relation} between candidates (i.e., prior knowledge based on some documents that defines a partial ordering between keyphrases - non keyphrases pairs) are collected and integrated into a random-walk parametric model. Then, the model defines a loss function, which is optimized using gradient descent, in order to learn the parameters, and computes the ranking score of each candidate word. Finally, consecutive words, phrases or n-grams are scored by using the sum of scores of individual words that comprise the phrase.
	
	\noindent \textbf{Keyphrase Extraction Using Supervision}
	
	\cite{GraphCoR2016Bougouin} (TopicCoRank) extend the unsupervised method TopicRank, making it capable of assigning domain-specific keyphrases that do not necessarily occur within
	the document, by unifying a second graph with the domain to the basic topic graph. Particularly, the keyphrases manually assigned to the training documents are considered as controlled keyphrases. In addition, these controlled keyphrases are not further clustered as they are supposed to be non-redundant. The unified graph is denoted as $G = (V = T \cup K, E = E_{\textit{in}} \cup E_{\textit{out}})$, where V are the graph vertices that comprise the set of topics $T = {t_1, t_2, \dots, t_n}$ and the set of controlled keyphrases $K = {k_1, k_2, \dots, k_m}$. Furthermore, the set $E_{\textit{in}}$ contains edges of type $<t_i, t_j>$ or $<k_i, k_j>$, whereas the set $E_{\textit{out}}$ contains edges of type $<k_i, t_j>$. An edge connects a controlled
	keyphrase $k_i$ with a topic $t_j$ if the controlled keyphrase is a member of the topic. Moreover, two topics $t_i$ and $t_j$ or two
	controlled keyphrases $k_i$ and $k_j$ are connected in case they co-occur within a sentence of the document or as keyphrases of a training document, respectively. The weight of the edge $<t_i, t_j>$ is the number of times ($w_{i,j}$) topics $t_i$ and $t_j$ co-occur in the same sentence within the document. Similarly, a weight is assigned to the edge $<k_i, k_j>$ equal to the number of times ($w_{i,j}$) keyphrases $k_i$ and $k_j$ are associated to the same document among the training documents. TopicCoRank assigns a score to every topic ($S(t_i)$) or controlled keyphrase ($S(k_i)$) using
	graph co-ranking, simulating the voting concept based on inner ($R_{\textit{in}}$) and outer recommendations ($R_{\textit{out}}$):
	$$S(t_i)=(1-\lambda_t)R_{\textit{out}}(t_i)+\lambda_tR_{in}(t_i)$$
	$$S(k_i)=(1-\lambda_k)R_{\textit{out}}(k_i)+\lambda_kR_{in}(k_i)$$
	where
	$$R_{\textit{in}}(v_i)=\sum_{v_j \in E_{\textit{in}}(v_i)}\frac{w_{\textit{ij}}S(v_j)}{\sum_{v_k \in E_{\textit{in}}(v_j)}w_{\textit{jk}}}$$
	$$R_{\textit{out}}(v_i)=\sum_{v_j \in E_{\textit{out}}(v_i)}\frac{S(v_j)}{|E_{\textit{out}}(v_j)|}$$
	where $v_i$ is a node of a keyphrase or a topic, $\lambda_t$ and $\lambda_k$ are parameters that control the influence of the inner recommendation over the outer recommendation.

	\cite{DBLP:journals/mta/YangLZXZQ18} propose a model to extract a set of keyphrases that correspond to particular events. In the first phase, they segment each document into  phrases. In the second phase, they use a Task-oriented Latent Dirichlet Allocation model (ToLDA) to extract candidate phrases for specific events. More precisely, they treat each event as a topic in their topic model. The ToLDA model utilizes a small set of \textit{seed} keyphrases (defined by the user) to guide the topics' generation process. In this way, the extracted phrases are more related/oriented to the seed keyphrases. In the last phase, they apply the PMI-IR algorithm \cite{DBLP:conf/ecml/Turney01} to get the synonyms of the candidate phrases. Finally, the candidate phrases are evaluated by inspectors, and the selected phrases are the final phrases which can be treated as seed keyphrases. The last two phases can be run iteratively.
	
	\noindent \textbf{Sequential Labeling} 
	
	We should not omit the recent work on keyphrase extraction where the task has been treated as a sequence tagging task using Conditional Random Fields (CRFs) by \cite{DBLP:conf/aaai/GollapalliLY17}. The features that are used represent linguistic, orthographic, and structure information from the document. Furthermore, they investigate feature-labeling and posterior regularization in order to integrate expert/domain-knowledge throughout the keyphrase extraction process. In addition, \cite{SogaardB17} exploit the idea of multi-task learning approaches \citep{DBLP:conf/icml/Caruana93} and joint modeling of tasks \citep{DBLP:conf/acl/MiwaB16, DBLP:conf/emnlp/HashimotoXTS17, bekoulis:18b, bekoulis:18a}
	to study the performance effectiveness of the keyphase detection task (on the  SemEval 2017 Task 10 dataset) when training along with other sequence labeling tasks (e.g., POS tagging, chunking) in a multi-task learning fashion.

	\subsection{Deep Learning Methods}
	\label{sec:deep_learning_methods}
	
	\cite{DBLP:conf/emnlp/ZhangWGH16} investigate the task of keyphrase extraction from tweets. They propose a deep recurrent neural network (RNN) model with two hidden layers. In the first hidden layer, they capture the keyword information, while in the second one, they extract the keyphrase based on the keyword information using a sequence labeling approach. Later, \cite{meng2017deep} (seq2seq) proposed a generative model for keyphrase prediction using an encoder-decoder framework that captures the semantic meaning of the content via a deep learning method (CopyRNN). First, the data is transformed into text-keyphrase pairs and then an RNN Encoder-Decoder model is applied to learn the mapping from the source sequence to target sequence. The key idea here is the compression of the source text into a hidden-layer representation with an encoder and the keyphrase generation with a decoder, based on the previous hidden-layer representation. However, the main drawbacks of this approach is that the generated keyphrases suffer from \textit{duplication} (at least two phrases expressing the same meaning) and \textit{coverage} (important topics in the document are not covered) issues, as it ignores the correlation among the target keyphrases. Hence,  \cite{CorrRNN2018Chen} propose CorrRNN, an alternative sequence-to-sequence architecture, capable of capturing such type of correlation constraints, i.e., (1) keyphrases should cover all document's topics and (2) differ from each other. Particularly, a coverage mechanism \citep{CoverageMechTu2016} is utilized to memorize which text parts of the document have been covered by previous phrases. Additionally, a review mechanism is proposed to satisfy the second constraint. More specifically, the review mechanism models the correlation between the already returned keyphrases and the keyphrase that is going to be returned. 
	
	From another point of view, both the latter methods \citep{meng2017deep, CorrRNN2018Chen} are mainly based on massive amounts of labeled documents for the training process. \cite{semi2018YE} try to deal with this problem by proposing two approaches for leveraging unlabeled data. The most effective way that is adopted in the context of the first approach (based on the experimental results) is the construction of synthetic labeled data by assigning keyphrases obtained by 2 unsupervised keyphrase extraction methods, the TfIdf and the TextRank \citep{mihalcea+tatau2004}, to unlabeled documents. Particularly, they run the 2 methods separately, and then they get the union of the two returned keyphrases' sets. After the construction of the synthetic data, the labeled data is mixed with the synthetic ones to train the sequence-to-sequence model. For the second approach to leverage unlabeled documents, a multi-task learning framework combines the basic task of keyphrase generation with an auxiliary task. The auxiliary task that has been chosen is the title generation which has been studied as a summarization problem \citep{summarizationRush2015}. Specifically, the 2 tasks share an encoder network but have different decoders. In the multi-task learning approach, the following training strategy is adopted: first, the parameters are estimated on the auxiliary task, with the dataset which is assigned with titles for unlabeled data for one epoch, then, the model is trained on the main task with the labeled dataset for 3 epochs. This training strategy is followed up to the convergence of the model of the main task. 
	
	However, \cite{DBLP:conf/icdm/WangLQXWCX18} improve the performance of keyphrase extraction in the unlabeled/insufficiently labeled target domain by transferring knowledge from resource-rich source domain (cross-domain perspective). More specifically, a Topic-based Adversarial Neural Network (TANN) is proposed that makes use of both labeled data in the related resource-rich domain and unlabeled data in source/target domain. The TANN model consists of 4 components:
	\begin{enumerate}
		\item[i)] Topic-based encoder: first, the input sentence is read by a Bidirectional Long Short-Term Memory (Bi-LSTM) Network encoder, which is used to capture the sequential information and then more attention is paid to the words related to the documents' topics by considering the correlation	score between the topic vector and the text words (topic correlation mechanism).
		\item[ii)] Domain discriminator: as there are some common words like
		``we introduce/we propose'' across domains, which are good indicators for the following keyphrases, adversarial learning is used to ensure that the extracted features by the encoder are invariant to the change of domains.
		\item[iii)] Target bidirectional decoder: to balance the effort of the adversarial process to eliminate all the domain-private information in the target domain,
		a bidirectional decoder with bidirectional reconstruction loss is used in the target domain.	
		\item[iv)] Keyphrase tagger: the output representation of the topic-based encoder is given as input to the keyphrase tagger to predict the label of each word in the source text.
	\end{enumerate}
	Finally, \cite{DBLP:conf/ircdl/BasaldellaAST18} propose a Bi-LSTM RNN which is able to exploit previous and future context of a given word. First, the document is split into sentences that are tokenized in words. Then, each word is associated to a word embedding. Finally, word embeddings are fed into a Bi-LSTM RNN. In this vein, \cite{DBLP:conf/www/AlzaidyCG19} combine a Bi-LSTM layer to model the sequential text data with a Conditional Random Field (CRF) layer to model dependencies in the output. Particularly, the first layer of the model is a Bi-LSTM network that captures the semantics of the input text sequence. The output of the Bi-LSTM layer is passed to a CRF layer that gives a probability distribution over the tag sequence using the dependencies among labels (i.e., KP: keyphrase word, Non-KP: not keyphrase word) of the entire sequence. 

	\subsection{Types of Features}
	\label{sec:types_features}
	
	Supervised keyphrase extraction methods employ different types of features to discover the importance of documents' terms. 
	 Table \ref{fig:supervised_features} gives an overview of the most popular features used by supervised methods. We also provide a categorization of the features that belong in the same family. Each category of features is discussed in a separate section. Table \ref{tbl:supervisedCategorizationMethods} gives an overview of the most representative supervised keyphrase extraction methods in terms of the learning algorithm they employ and the type of their input features/the knowledge they incorporate: statistical (Stat.), positional (Posit.), linguistic (Ling.), context (Cont.), stacking (Stack.) and external knowledge (Ext.). Once again, the categorization scheme and the table can be extended as new methods are developed with new features, feature categories and additional methods.

	\subsubsection{Statistical Features}
	\label{sec:statistical_features}
	
	Features, such as the \textit{term frequency (Tf)}, the \textit{inverse document frequency (Idf)}, the \textit{TfIdf} score as well as statistical scores, such as the \textit{phrase entropy} are very popular and utilized by many methods, e.g., \cite{witten1999kea} (KEA), \cite{medelyan2009human} (MAUI), \cite{caragea2014citation} (CeKE), \cite{hulth2003improved}, \cite{DBLP:conf/aaai/2018} (SurfKE), \cite{DBLP:conf/semeval/WangL17} (PCU-ICL), \cite{jiang2009ranking} (Ranking SVM), \cite{DBLP:conf/icadl/NguyenK07}, \cite{DBLP:conf/cikm/ZhangCLG0X17} (MIKE), \cite{Wiki2008shi}, \cite{GraphCoR2016Bougouin}, and \cite{DBLP:conf/semeval/NguyenL10} (WINGNUS). In some cases, thresholds are used for the scores mentioned above for the creation of boolean features, ending up to values such as \textit{high}, \textit{low} etc. Moreover, \cite{DBLP:conf/aaai/GollapalliLY17} detect the features that co-occur with the keyphrases very often and incorporate this information via extra features, whereas \cite{DBLP:journals/mta/YangLZXZQ18} adopt a user-guided Task-oriented Latent Dirichlet Allocation model (ToLDA) to generate topic distributions.
	
	\subsubsection{Positional Features}
	\label{sec:positional_features}
	
	This category includes features that indicate the \textit{appearance of phrases in specific positions (e.g. the 1$^{st}$ occurrence of the phrase in the text), sections, titles, abstracts, citation contexts}. The part(s) of the document, i.e., section, title, abstract where the phrases occur, imply the importance of the corresponding phrases. For instance, phrases that appear in title or in early parts of the document indicate that these phrases play an important role. Examples of such methods that use this type of information are \cite{jiang2009ranking} (Ranking SVM), \cite{DBLP:conf/cikm/ZhangCLG0X17} (MIKE), \cite{Wiki2008shi} and \cite{DBLP:conf/aaai/GollapalliLY17}. \cite{DBLP:conf/icadl/NguyenK07} utilize a vector of frequency features for 14 generic section headers (e.g., Abstract, Categories and Subject Descriptors, General Terms, Introduction, Conclusions etc.). A maximum entropy-based classifier is used to infer the above generic section header. 
	
	The \textit{exact position of the 1st/last occurrence of a phrase} or even though an average score of its positions in the text gives useful information for the role of a phrase as keyphrase. Popular supervised models are trained incorporating the above features, e.g., \cite{witten1999kea} (KEA), \cite{medelyan2009human}, (MAUI), \cite{caragea2014citation} (CeKE), \cite{jiang2009ranking} (Ranking SVM), \cite{DBLP:conf/cikm/ZhangCLG0X17} (MIKE), \cite{Wiki2008shi} and \cite{DBLP:conf/semeval/NguyenL10} (WINGNUS). The \textit{spread} is another feature related to positional information which is calculated as the number of words between the first and last occurrences of a phrase in the document \citep{medelyan2009human} (MAUI). In the same spirit, features, such as the \textit{distance between phrase and citation} \citep{DBLP:conf/semeval/WangL17} (PCU-ICL) seem to help the learning process, whereas sequential labeling approaches benefit from information for the internal structure of a text, e.g., the annotation of \textit{sentence boundaries} \citep{DBLP:conf/aaai/GollapalliLY17}.
	
	\subsubsection{Linguistic Features}
	\label{sec:linguistic_features}
	
	\cite{hulth2003improved} has shown for the first time, through an extended empirical study, that the use of POS tag(s) as a feature improves the performance of a supervised keyphrase extraction system regardless the selection approach of the candidate phrases. \textit{Phrase POS features} or \textit{Noun Phrase (NP)-chunking} to discover phrases that match a predefined POS pattern have been also utilized by \cite{caragea2014citation} (CeKE), \cite{DBLP:conf/semeval/WangL17} (PCU-ICL), \cite{DBLP:conf/cikm/ZhangCLG0X17} (MIKE) and \cite{DBLP:conf/aaai/GollapalliLY17}. Besides the POS tag sequence feature, \cite{DBLP:conf/icadl/NguyenK07} consider morphological features, i.e., the \textit{suffix sequence} and the \textit{acronym status}.  
	
	Other types of linguistic features, such as the \textit{stemmed} form of the words \citep{DBLP:conf/semeval/WangL17, DBLP:conf/aaai/GollapalliLY17} or boolean features that indicate whether the term \textit{is a stopword/consists of punctuation/has capitalized letter(s)} are also met in the bibliography \citep{DBLP:conf/semeval/WangL17} (PCU-ICL). Finally, a quite popular feature is the \textit{phrase length} which is adopted by various methods, such as \cite{DBLP:conf/semeval/WangL17} (PCU-ICL), \cite{medelyan2009human} (Maui), \cite{jiang2009ranking} (Ranking SVM), \cite{Wiki2008shi} and \cite{DBLP:conf/semeval/NguyenL10} (WINGNUS).

	\subsubsection{Context Features}
	\label{sec:context_features}
	
	The context of a word shows its meaning. In particular, even though we do not know a term, we can guess its meaning based on the context where it is used. Context-based features are quite useful for the keyphrase extraction task which is a process that is inseparable from extracting the central meaning of a text document and the expression of the topics discussed in it. The context of a phrase can also be simply expressed by providing the \textit{previous/next token of the phrase} of interest \citep{DBLP:conf/semeval/WangL17} (PCU-ICL). This type of features often appears along with \textit{POS/syntactic features of the previous/next token of the target phrase} \citep{DBLP:conf/semeval/WangL17} (PCU-ICL). The same paradigm is followed by the sequential labeling approach in \cite{DBLP:conf/aaai/GollapalliLY17} where similar \textit{bigram/skipgram/compound features} related to stems and POS tag(s) are exploited.
	
	Alternative types of context meaning can be captured by positional information incorporating the \textit{relative position of the phrase in a given text}, i.e, the position of the first occurrence of a phrase normalized by the length of the target publication \citep{caragea2014citation} (CeKE), \citep{hulth2003improved}, \citep{DBLP:conf/cikm/ZhangCLG0X17} (MIKE), and \citep{DBLP:conf/semeval/WangL17} (PCU-ICL). Recently, \cite{DBLP:conf/aaai/2018} (SurfKE) as well as \cite{meng2017deep}, \cite{CorrRNN2018Chen}, \cite{DBLP:conf/icdm/WangLQXWCX18}, \cite{DBLP:conf/www/AlzaidyCG19} and \cite{semi2018YE} go one step beyond the context-based features by \textit{learning features/embeddings} using graph structures and neural networks, respectively.
	
		\begin{table}[H]
		\renewcommand\arraystretch{1.4}\arrayrulecolor{black}
		\centering
		
		\begin{tabu}{p{3cm}<{\hskip 2pt} !{\fooo} >{\raggedright\arraybackslash}p{8cm}}
			\hiderowcolors
			\toprule
			\addlinespace[1.5ex]
			\textbf{Feature Category} & \textbf{Description} \\ \hline
			Statistical & {\small\textcolor{black}{\tabitem} Tf/Idf/TfIdf} \newline {\small\textcolor{black}{\tabitem} Number of sentences containing the phrase} {\small\newline  \textcolor{black}{\tabitem} Words or phrase entropy} \newline {\small\textcolor{black}{\tabitem} Correlations between features and the phrase} {\small\newline \textcolor{black}{\tabitem} Topic distributions (LDA)}\\
			Positional & {\small\textcolor{black}{\tabitem} Appearance in
				specific parts of the fulltext, e.g., sections, titles etc.} \newline {\small\textcolor{black}{\tabitem} Position of the (1st or last) occurrence} \newline {\small\textcolor{black}{\tabitem} Distance between phrase and citation} \newline {\small\textcolor{black}{\tabitem} Section occurrence vector} \newline {\small\textcolor{black}{\tabitem} Sentence boundaries} \newline {\small\textcolor{black}{\tabitem} Spread}\\
			Linguistics & {\small\textcolor{black}{\tabitem} Stemmed unigram} \newline {\small\textcolor{black}{\tabitem} Boolean features: IsCapilazed, IsStopword, AllPunctuation} \newline {\small\textcolor{black}{\tabitem} POS tags, NP-chunking} \newline  {\small\textcolor{black}{\tabitem} Phrase length} \newline {\small\textcolor{black}{\tabitem} Suffix sequence} \newline {\small\textcolor{black}{\tabitem} Acronym status}\\
			Context & {\small\textcolor{black}{\tabitem} Previous/next token of the phrase}  \newline {\small\textcolor{black}{\tabitem} POS/syntactic features of previous/next token of phrase} \newline {\small\textcolor{black}{\tabitem} Relative position of the phrase in given text} \newline {\small\textcolor{black}{\tabitem} Learning embeddings/features (using NN/graph)} \newline {\small\textcolor{black}{\tabitem} Bigram, skipgram, compound features (stem, pos, chunk etc.)} \\
			Stacking & {\small\textcolor{black}{\tabitem} Unsupervised methods output} \newline {\small\textcolor{black}{\tabitem} Supervised methods output}   \\
			External Knowledge & {\small\textcolor{black}{\tabitem} Existence of the phrase in ontologies (IEEE) or as a Wikipedia link} \newline {\small\textcolor{black}{\tabitem} Wikipedia based Idf/phraseness} \newline {\small\textcolor{black}{\tabitem} (Pretrained) Word embedding of the phrase} \newline {\small\textcolor{black}{\tabitem} Supervised keyphraseness} \newline {\small\textcolor{black}{\tabitem} Bias based on previous research} \newline {\small\textcolor{black}{\tabitem} TitleOverlap} \newline {\small\textcolor{black}{\tabitem} Semantic feature weight (returned by HITS with Wikipedia Info)}\\ \hline
			
		\end{tabu}
		\caption{A categorization of the most popular features that are used by supervised keyphrase extraction methods.}
		\label{fig:supervised_features}
	\end{table}

	\subsubsection{Stacking}
	\label{sec:stacking_features}
	
	Stacking is a way to ensemble models. The idea is that we build various learners and then use them to build another model (i.e., their predictions are used as features for a second level model), and so on. We usually end up with a model that has better performance than any individual intermediate model. \cite{DBLP:conf/semeval/WangL17} (PCU-ICL) stack trees upon output from unsupervised keyphrase extraction methods, i.e., TextRank and SGRank, whereas \cite{DBLP:conf/aaai/GollapalliLY17} also incorporate information from the output of TfIdf, TextRank, SingleRank, and ExpandRank.
	
	Moreover, models are stacked upon supervised approaches. Specifically, at a second layer \cite{DBLP:conf/semeval/WangL17} (PCU-ICL) stack a linear model (linear SVM) upon random forest. Finally, \cite{DBLP:conf/aaai/GollapalliLY17} imply a high probability of a word to be a keyphrase when the phrase is identified by CeKE and Maui through special labeled features.

	\subsubsection{External Knowledge}
	\label{sec:external_features}
	
	There is a great deal of features that are calculated based on external knowledge sources/corpora. In this category, there are boolean features that indicate the \textit{existence of the phrase in an ontology/knowledge base}, e.g., IEEE, \citep{DBLP:conf/semeval/WangL17} (PCU-ICL) as well as numeric features based on statistics, such as the \textit{Wikipedia based Idf} calculation \citep{DBLP:conf/semeval/WangL17} (PCU-ICL) and the \textit{Wikipedia based keyphraseness} which is the likelihood of a phrase being a link in the
	Wikipedia corpus \citep{medelyan2009human} (Maui). Similar to these features is the \textit{supervised keyphraseness} which counts the number of times a phrase appears as a keyphrase in the training data \citep{medelyan2009human} (Maui), the \textit{semantic feature weight} for each candidate phrase that is returned by a graph-based ranking approach incorporating semantic information from Wikipedia \citep{Wiki2008shi}, and the \textit{TitleOverlap} which is the number of times
	a phrase appears in the title of other documents in DBLP database \citep{DBLP:conf/semeval/NguyenL10} (WINGNUS). Another source of external knowledge is  previous research results or expert knowledge that can affect the output of the keyphrase extraction system with the introduction of \textit{bias} \citep{DBLP:conf/aaai/GollapalliLY17} via labeled features whose values are restricted using a posterior regularization framework. In the same spirit, the interaction with the user in \cite{DBLP:journals/mta/YangLZXZQ18} as well as the utilization of reference keyphrases from the training data in \cite{GraphCoR2016Bougouin} can be considered as introducing external knowledge into the system.

	Recently, features based on pretrained word embeddings are widely used. Particularly, deep learning methods \citep{DBLP:conf/ircdl/BasaldellaAST18, DBLP:conf/emnlp/ZhangWGH16,DBLP:conf/icdm/WangLQXWCX18,DBLP:conf/www/AlzaidyCG19} use such embeddings as input. Additionally, GloVe pretrained word embeddings are used along with an Idf-weighted scheme for each phrase representation by \cite{DBLP:conf/semeval/WangL17} (PCU-ICL).
	
	\begin{table}
		\renewcommand\arraystretch{1.4}\arrayrulecolor{black}
		\centering
		\scalebox{0.95}{
			\begin{tabu}{p{0.75cm}<{\hskip 1pt} !{\foooo} >{\raggedright\arraybackslash}p{2.5cm}<{\hskip 1pt} !{\foooo} >{\raggedright\arraybackslash}p{2.5cm} <{\hskip 1pt} !{\foooo} >{\raggedright\arraybackslash} p{0.75cm} p{0.75cm} p{0.75cm} p{0.75cm} p{0.75cm} p{0.75cm} p{0.75cm}}
				\hiderowcolors
				\hline
				\addlinespace[1.5ex]
				\textbf{Year} & \textbf{Method} & \textbf{ML Algorithm}   & \textbf{Stat.} & \textbf{Posit.} & \textbf{Ling.} & \textbf{Cont.} & \textbf{Stack.} & \textbf{Ext.}  \\ \hline
				1999	& {\small \textcolor{red}{\cite{witten1999kea}}}     &   \textit{Naive Bayes}        & \checkmark          & \checkmark          &            &         &          &                    \\ 
				2003    & {\small \textcolor{red}{\cite{hulth2003improved}}} &   \textit{Rule Induction/Bagging}      & \checkmark           &            & \checkmark       & \checkmark       &          &                    \\ 
				2007	& {\small \textcolor{red}{\cite{DBLP:conf/icadl/NguyenK07}}} & \textit{Naive Bayes} & \checkmark           & \checkmark          &  \checkmark         &        &         &                    \\
				2008 & {\small \textcolor{red}{\cite{Wiki2008shi}}} & \textit{Logistic Regression} & \checkmark           & \checkmark          & \checkmark          &        &         & \checkmark                 \\	
				2009 & {\small \textcolor{red}{\cite{medelyan2009human}}} \newline \newline {\small \textcolor{red}{\cite{jiang2009ranking}}}    &  \textit{Bagged Decision Trees}  \newline \textit{SVM}      &  \checkmark  \newline \newline \checkmark         & \checkmark  \newline \newline  \checkmark          & \checkmark  \newline \newline \checkmark          &         &          & \checkmark                  \\ 
				2010 & {\small \textcolor{red}{\cite{DBLP:conf/semeval/NguyenL10}}} & \textit{Naive Bayes} & \checkmark           & \checkmark          & \checkmark          &        &         & \checkmark                  \\
				2014 & {\small \textcolor{red}{\cite{caragea2014citation}}}      &  \textit{Naive Bayes}       & \checkmark           & \checkmark          & \checkmark          & \checkmark       &          &                    \\ 
				2016 & {\small \textcolor{red}{\cite{DBLP:conf/emnlp/ZhangWGH16}}} \newline {\small \textcolor{red}{\cite{GraphCoR2016Bougouin}}} & \textit{RNN} \newline \textit{Graph-based Method}  &  \vspace{0.3cm} \checkmark           &            &            &         &          & \checkmark  \newline  \checkmark               \\ 
				2017 & {\small \textcolor{red}{\cite{DBLP:conf/semeval/WangL17}}} \newline {\small \textcolor{red}{\cite{meng2017deep}}} \newline {\small \textcolor{red}{\cite{DBLP:conf/aaai/GollapalliLY17}}} \newline {\small \textcolor{red}{\cite{DBLP:conf/cikm/ZhangCLG0X17}}} & \textit{Ensemble (RF/SVM)} \newline \textit{seq2seq Learning} \newline \textit{CRFs} \newline \textit{Random-walk Parametric} Model  & \checkmark   \newline \newline \checkmark \newline \checkmark         & \checkmark  \newline \newline \checkmark \newline \checkmark          & \checkmark  \newline \newline \checkmark \newline \checkmark         & \checkmark \newline \checkmark  \newline \checkmark \newline \checkmark  & \checkmark \newline \newline \checkmark       & \checkmark \newline \newline \checkmark                \\ 

				2018 & {\small \textcolor{red}{\cite{DBLP:conf/aaai/2018}}} \newline {\small \textcolor{red}{\cite{CorrRNN2018Chen}}} \newline {\small \textcolor{red}{\cite{semi2018YE}}} \newline \newline {\small \textcolor{red}{\cite{DBLP:conf/ircdl/BasaldellaAST18}}} \newline {\small \textcolor{red}{\cite{DBLP:journals/mta/YangLZXZQ18}}} \newline \newline {\small \textcolor{red}{\cite{DBLP:conf/icdm/WangLQXWCX18}}} & \textit{Naive Bayes} \newline  \newline \textit{seq2seq Learning} \newline \textit{Multi-task Learning (seq2seq Model)}) \newline \textit{Bi-LSTM RNN} \newline \textit{Task-oriented LDA Model} \newline \textit{Topic-based Adversarial Neural Network} & \checkmark \newline \newline \newline \newline \newline \newline \checkmark         &            &            & \checkmark  \newline \newline \checkmark \newline \checkmark  \newline \newline \newline \newline \newline \checkmark   &          &  \vspace{2.1cm} \checkmark  \newline \checkmark \newline \newline \checkmark               \\ 
				
				2019 & {\small \textcolor{red}{\cite{DBLP:conf/www/AlzaidyCG19}}}  & \textit{Bi-LSTM-CRF Sequence Labeling} & & & & \checkmark & & \checkmark \\
				
				\hline
		\end{tabu}}
		\caption{A time line of all the supervised keyphrase extraction methods discussed in this review, along with their key-characteristics.}
		\label{tbl:supervisedCategorizationMethods}
	\end{table}

	
	\subsection{Comparative Experimental Results}
	\label{sec:exp_comparison_supervised}
	
		 As we don't conduct experiments ourselves for supervised methods, we here present results for supervised methods, taking experimental results from the scientific publications that are included in our study. Table \ref{tbl:supervised_approaches_results} shows the performance of some earlier, state-of-the-art and more recent supervised methods according to the F$_1$-measure on 3 popular keyphrase extraction datasets (see Section \ref{data} for further details about the datasets) at the top 5 (F$_1$@5) and top 10 (F$_1$@10). The performance evaluation is based on the \textit{exact match evaluation} where the number of correctly matched phrases with the golden ones are determined based on string matching. We see that the performances of the two state-of-the-art supervised models, i.e., Maui and KEA, are unstable on some datasets, but Maui achieves better performance than the KEA baseline model. Undoubtedly, the deep learning methods CopyRNN and CorrRNN achieve better $F_1$ scores compared to the 2 baselines (Maui and KEA), indicating that their models learn the mapping patterns from source text to target keyphrases to a satisfying degree. In addition, the limited results of WINGNUS method on Semeval dataset seem to be quite promising, as the method achieves quite high $F_1$ scores compared to the baselines. However, the deep learning methods, except for RNN, outperform all the other methods. Finally, the coverage and the review mechanisms that are used by CorrRNN make the difference regarding its high performance in all datasets.

	\begin{table}[H]
		\centering
		\scalebox{0.9}{
			\begin{tabu}{|c|c|c|c|c|c|c|}
				\hiderowcolors
				\hline
				\multirow{2}{*} & \multicolumn{2}{c|}{Semeval} & \multicolumn{2}{c|}{Krapivin} & \multicolumn{2}{c|}{NUS} \\ \hline
				Methods   & F$_1$@5          & F$_1$@10        & F$_1$@5          & F$_1$@10         & F$_1$@5        & F$_1$@10      \\ \hline
				KEA                      & 0.025         & 0.026        & 0.110         & 0.152         & 0.069       & 0.084      \\ 
				MAUI                     & 0.044         & 0.039        & 0.249         & 0.216         & 0.249       & 0.268      \\ 
				CopyRNN                  & 0.291         & 0.296        & 0.302         & 0.252         & 0.342       & 0.317      \\ 
				CorrRNN                  & \textbf{0.320}         & \textbf{0.320}        & \textbf{0.318}         & \textbf{0.278}         & \textbf{0.358}       & \textbf{0.330}      \\
				RNN                      & 0.157         & 0.124        & 0.135         & 0.088         & 0.169       & 0.127      \\ 
				WINGNUS                  & 0.205         & 0.247        & -             & -             & -           & -          \\ \hline
		\end{tabu}}
		\caption{Performance of supervised keyphrase extraction methods. There are no available results for WINGNUS regarding the Krapivin and NUS datasets in the scientific literature. 
		}
		\label{tbl:supervised_approaches_results}
	\end{table}

	\subsection{Subjectivity and Class Imbalance}
	\label{imbalanced}
	
	Unbalanced training data is a very common problem in supervised keyphrase extraction because candidate phrases that are not annotated by humans as keyphrases are consider as negative training examples. This problem occurs for many reasons (e.g., authors select as keyphrases those that promote their work in a particular way or those that are popular regarding concept drift etc.) which can be summarized into one, subjectivity. \cite{sterckx2016supervised} conduct an interesting study where they conclude that unlabeled keyphrase candidates are not reliable as negative examples. 
	
	To deal with this problem, \cite{sterckx2016supervised} propose to treat supervised keyphrase extraction as Positive Unlabeled Learning based on \cite{elkan2008learning} by assigning weights to training examples depending on the document and the candidate. Particularly, they try to model the annotations by multiple annotators and the uncertainty of negative examples. Firstly, they train a classifier on a single annotator's data and use the predictions on the negative/unlabeled phrases as weights. Then, a second classifier is trained on the re-weighted data to predict a final ranking/labels of the candidates. It is worth to see in depth the process discussed above step by step. A weight equal to 1 is assigned to every positive example and the unlabeled examples are duplicated such that one copy is considered as positive (weight $w(x) = P(\textit{keyphrase}|x,s=0)$, s indicates whether x is labeled or not) and the other copy as negative (weight $1 - w(x)$). According to \cite{sterckx2016supervised}, this weight is not just a factor of the prediction of the initial classifier as proposed in \cite{elkan2008learning}. Actually, they normalize the predictions and they include a measure for pairwise co-reference between unlabeled candidates and known keyphrases in a function $\textit{Coref(candidate, keyphrase)} \in \{0,1\}$ returning 1 if one of the binary indicator features, presented in \cite{bengtson2008understanding} is present. 
	
	To sum up, the problem of subjectivity could be partially addressed using multiple
	annotators or treating supervised keyphrase extraction as Positive Unlabeled Learning \citep{DBLP:journals/tochi/ChuangMH12, sterckx2016supervised}. In this direction, \cite{DBLP:journals/lre/SterckxDDD18} propose as solution the creation of new collections for the evaluation of keyphrase extraction methods, derived from various sources with multiple annotations. The golden set of keyphrases, which is utilized for evaluation reasons, also incorporates subjectivity issues. Therefore, the need for reliable semantic evaluation approaches is also evident. Although there is a variety of evaluation approaches in the keyphrase extraction task, the most preferred one is the more classic and strict exact (string) matching. 
	
	\section{Evaluation Approaches}
	\label{evaluation}
	
	Undoubtedly, it is a legitimate objective to have evaluation measures or approaches that offer a fair comparison of the keyphrase extraction methods, i.e., a ranking of the performance of the methods in order to determine which one is the most effective. However, we are also interested in knowing the method's accuracy, i.e, which percentage of the golden keyphrases the method returns or an indicative score related to the success rate. In this section, we briefly describe the most well-known evaluation measures that are used to evaluate the performance of the methods included in the literature review:
	\begin{enumerate}
		\item \textit{Precision/Recall/$F_1$-measure}: 
		
		Precision is defined as:
		$$\textit{precision} = \frac{\textit{number of correctly matched}}{\textit{total number of extracted}}=\frac{\textit{TP}}{\textit{TP+FP}}$$ where $\textit{TP}$ is the number of true positives and $\textit{FP}$ the number of false positives, respectively.
		
		Recall is defined as:
		$$recall = \frac{\textit{number of correctly matched}}{\textit{total number of assigned}}=\frac{\textit{TP}}{\textit{TP}+\textit{FN}}$$ where $\textit{TP}$ is the number of true positives and $\textit{FN}$ the number of false negatives, respectively.
		
		F$_1$-measure is defined as:
		$$F_\textit{1-measure} = 2\times\frac{\textit{precision} \times \textit{recall}}{\textit{precision} + \textit{recall}}$$
		\item \textit{Ranking Quality Measures}  are used to measure the ranked extracted phrases. Such measures take into account the relative order of the phrases extracted by the keyphrase extraction systems. Popular ranking evaluation measures in the keyphrases extraction task are the following: 
		\begin{itemize}
			\item \textit{Mean Reciprocal Rank (MRR)} \citep{voorhees1999trec} that is defined as follows: $$\textit{MRR} = \frac{1}{|D|}\sum_{d \in D}\frac{1}{\textit{rank}_d}$$ where $\textit{rank}_d$ is denoted as the rank of the first correct keyphrase with all extracted
			keyphrases, $D$ is the document set for keyphrase extraction and $d$ is a specific document.
			\item \textit{Mean Average Precision (MAP)} \citep{DBLP:reference/db/2009} that also takes the ordering of a particular returned list of keyphrases into account. The average precision (AP) of the list is defined as follows: $$\textit{AP}=\frac{\sum_{r=1}^{|L|}P(r)\cdot \textit{rel}(r)}{|L_R|}$$
			where $|L|$ is the number of items in the list, $|L_R|$ is the number of relevant items, $P(r)$ is the precision when the returned list is treated as containing only its first $r$ items, and $\textit{rel}(r)$ equals 1 if the $\textit{r}^\textit{th}$ item
			of the list is in the golden set and 0 otherwise.
			
			By averaging AP over a set of $n$ document cases, we obtain the Mean Average
			Precision (MAP), defined as follows:
			$$\textit{MAP}=\frac{1}{n}\cdot \sum_{i=1}^{n}\textit{AP}_i$$ where $\textit{AP}_i$ is the average precision of the extracted keyphrases list returned for a document.
		\end{itemize}
		\item \textit{Binary preference measure (Bpref)} \citep{buckley2004retrieval} is a summation-based measure of how many relevant phrases are ranked before irrelevant ones and it is defined as follows:
		$$\textit{Bpref} = \frac{1}{R}\sum_{r \in R}1-\frac{|\textit{n ranked higher than r}|}{M}$$ where $R$ are the correct keyphrases within $M$ extracted keyphrases in which $r$ is a correct keyphrase and $n$ is an incorrect keyphrase for a document.
		\item \textit{Average of Correctly Extracted Keyphrases - (ACEK)} is the average number of the extracted keyphrases that also belong to the document's golden set of keyphrases. This was the first type of performance evaluation that is used in the keyphrase extraction task but it is not widely used anymore, as precision and recall offer a more complete view for a system's performance in terms of the set of the extracted keyphrases. 
	\end{enumerate}
	
	The above measures are usually calculated following one of the following directions (approaches):  
	\begin{itemize}
		\item \textit{Exact match evaluation} where the number of correctly matched phrases with the golden ones are determined based on string matching. In most cases, stemming is a preprocessing step to determine the match of two keyphrases.
		\item \textit{Manual evaluation} where experts decide whether the returned keyphrases by a system are wrong or right. However, this type of evaluation requires the investment of time and money and is characterized by great subjectivity \citep{zesch2009approximate}
		\item \textit{Partial match evaluation}, a looser evaluation process that is proposed by \cite{DBLP:conf/ecir/RousseauV15}, which calculates the Precision, Recall and F$_1$-measure between the set of words found in all golden keyphrases and the set of words found in all extracted keyphrases.  Again, stemming is a required preprocessing step. However, such type of evaluation cannot evaluate the syntactic correctness of the phrases or deal with more complex issues such as over-generation problems and overlapping keyphrase candidates. 
	\end{itemize}

	Furthermore, \cite{kim2010evaluating} study the issue of n-gram-based evaluation measures for automatic keyphrase extraction. In this study various evaluation measures developed for machine translation and summarization are included as well as the R-precision evaluation measure. 
	However, such kind of evaluation, 
	is not widely adopted by the keyphrase extraction community, as it is not found in any of the scientific publications included in the current survey. 
	
	\begin{figure}
		\centering
		\includegraphics[width=0.8\linewidth]{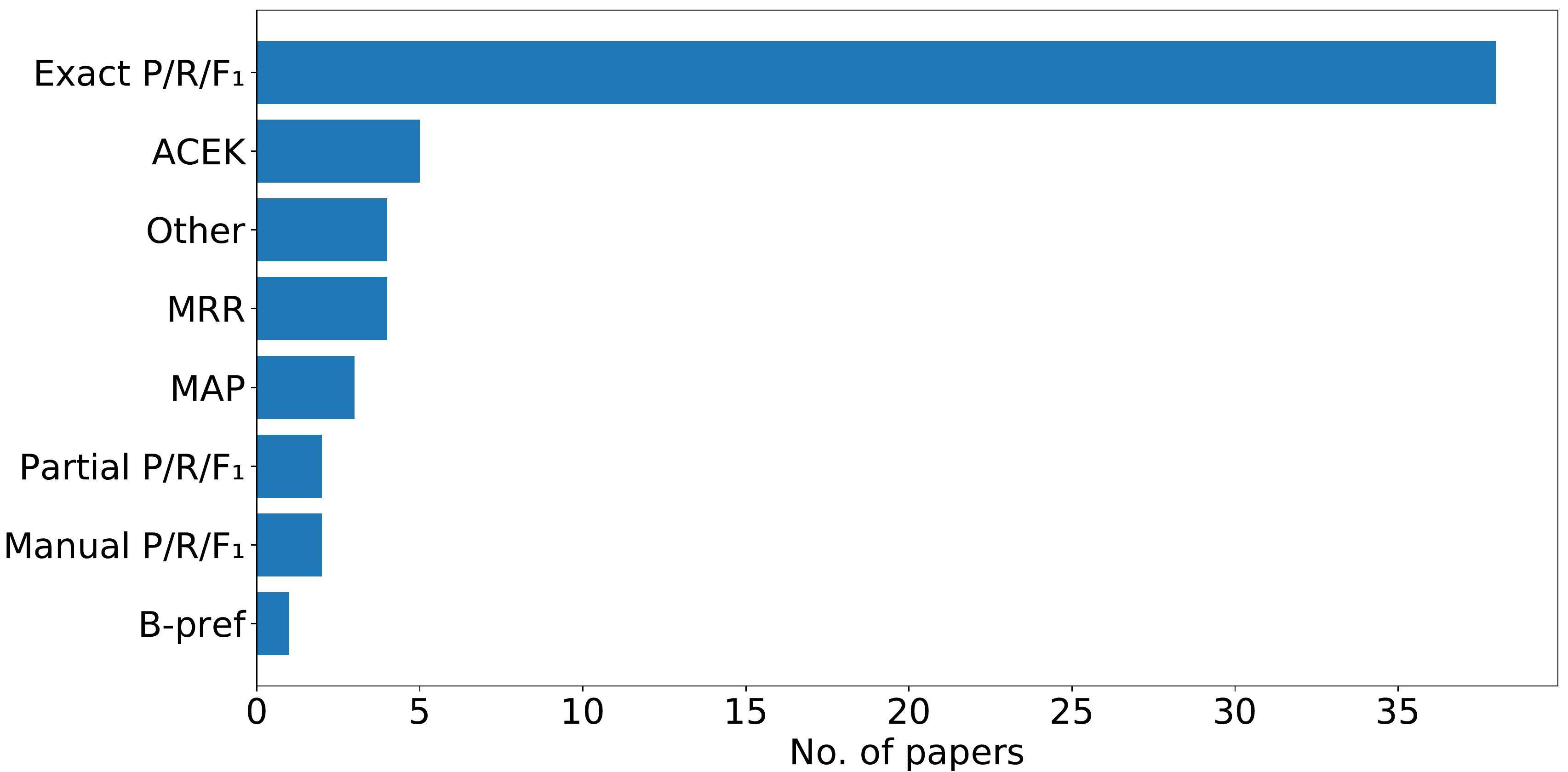}
		\caption{Popularity of evaluation approaches. ACEK, MRR, MAP and B-pref are used along with the exact match evaluation approach.}
		\label{fig:popularity_evaluation}
	\end{figure}

	In Fig. \ref{fig:popularity_evaluation}, we show the popularity of the most well-known evaluation measures that have been used in the keyphrase extraction task, which is based on our bibliographic study. Some works have used more than one types of evaluation measures/approaches. Infrequent as well as insignificant evaluation approaches are categorized to the ``Other'' group of methods. Precision/Recall/F$_1$-measure has indeed been used in the majority of the related work. However, their calculation based on the exact phrase matching, which is very popular as well, is too strict and equally penalizes a predicted keyphrase that is completely different from a golden keyphrase and a predicted keyphrase whose words are a subset or superset of the words of a golden keyphrase. Furthermore, this type of evaluation cannot identify the semantic similarity of two different phrases. Additionally, another issue of this approach is that even small variants in the keyphrases have considerable impact on the evaluation results \citep{kim2010evaluating}. Suppose we would like to find the keyphrases of the 628247.txt file from the Krapivin dataset whose golden set of keyphrases includes the phrases ``approximate search'' and ``similarity search'' and the keyphrase extraction method returns as output the keyphrase ``approximate similarity search'', i.e., a correct keyphrase that appears in the text. However, the exact match evaluation approach will consider this keyphrase as wrong. For this reason, this type of evaluation could be considered as suboptimal. 
	Moreover, Precision/Recall/F$_1$-measure at the top $N$ phrases, where $N=5, 10, 15, 20$, prevail over the ranking quality measures, as the success rate at top of the ranking seems to more important compared to the actual ordering of the keyphrases in most applications.

	We believe that this discussion will eventually be more fruitful for the research community and stir more research. A solution to the above problems is the development of semantic evaluation approaches for the keyphrase extraction task. In this direction, \cite{papagiannopoulou2018local} proposed an alternative evaluation, focused on the ``gold'' keyphrases' comparison with the returned keyphrases of a system, that exploits the representation of the words as vectors. Particularly, the cosine similarity is computed between the mean word vector derived from the ground truth's keyphrases and the mean word vector of the system's phrases.

	\section{Datasets and Software}
	\label{data-comp-software}
	
	In this section, we refer to popular datasets that are used for the development and evaluation of the keyphrase extraction methods (Section \ref{data}). Moreover, in Section \ref{software}, we present available free and commercial keyphrase extraction software providing a short description with their corresponding features.
	
	\subsection{Datasets}
	\label{data}
	Keyphrase extraction systems have been evaluated on datasets from various text sources, i.e., full-text scientific publications, paper abstracts and news documents. Some of the most well-known datasets are given in Table \ref{KP_datasets} grouped by
	their type of source. 
	For each dataset, we give its (i) name (Dataset), (ii) creator(s) (Created By), (iii) number of text documents (Docs), (iv) text language (Language), (v) annotators (Annotation Type), which can be author-assigned keyphrases, reader-assigned keyphrases or controlled/uncontrolled keyphrases assigned by professional indexers, and vi) the number of times that the dataset is used in the evaluation process with respect to the scientific articles that are included in the current survey. There are also 16 papers that use other, not so well-known datasets, that are not publicly available or are created for the evaluation of their methods. 
	
	A number of datasets for evaluating keyphrase extraction algorithms are available on Github\footnote{\url{https://github.com/zelandiya/keyword-extraction-datasets}}\textsuperscript{,}\footnote{\url{https://github.com/snkim/AutomaticKeyphraseExtraction}}, as well as on the website of Sujatha Das Gollapalli\footnote{\url{https://sites.google.com/site/sujathadas/home/datasets}}.
	
	\begin{table}[H]
		\centering
		\scalebox{0.92}{
			\begin{tabu}{|p{1.3cm}|c|l|c|c|c|l|}
				\hiderowcolors
				\hline
				Type                         & Dataset       & \multicolumn{1}{c|}{Created By}                          & Docs & Language   & Annotation Type & Freq. \\ \hline
				\multirow{4}{*}{\parbox{1.3cm}{Full-text Papers}} & NUS           & \cite{DBLP:conf/icadl/NguyenK07}        & 211         & English    & Authors/Readers & 9 \\ 
				& Krapivin      & \cite{krapivin2009}                     & 2304        & English    & Authors   & 6      \\ 
				& Semeval2010   & \cite{Kim:semeval2010}                  & 244         & English    & Authors/Readers & 12 \\ 
				& Citeulike-180 & \cite{medelyan2009human}               & 180         & English    & Readers  & 1       \\ \hline
				\multirow{3}{*}{\parbox{1.2cm}{Paper Abstracts}}  & Inspec        & \cite{hulth2003improved}                & 2000        & English    & Indexers  & 21 \\ 
				& KDD           & \cite{gollapalli2014extracting}         & 755         & English    & Authors & 6        \\ 
				& KP20k           & \cite{meng2017deep}         & 567830         & English    & Authors & 3        \\ 
				& WWW           & \cite{gollapalli2014extracting}         & 1330        & English    & Authors  & 6       \\ \hline
				\multirow{4}{*}{News}             & DUC-2001      & \cite{wan+xiao2008}                     & 308         & English    & Readers  & 10       \\ 
				& 500N-KPCrowd  & \cite{newsMarujo2012}                   & 500         & English    & Readers  & 4        \\ 
				& 110-PT-BN-KP  & \cite{DBLP:conf/interspeech/MarujoVN11} & 110         & Portuguese & Readers  & 1       \\ 
				& Wikinews      & \cite{bougouin2013topicrank}            & 100         & French     & Readers   & 1      \\ \hline
		\end{tabu}}
		\caption{Popular evaluation datasets grouped by their type of source (Type).}
		\label{KP_datasets}
	\end{table}

	Furthermore, several approaches to the creation of such datasets have been proposed, in the past \citep{medelyan2009human, DBLP:journals/lre/SterckxDDD18}. Recently, \cite{DBLP:journals/lre/SterckxDDD18} have created new keyphrase extraction collections consisting of on-line news/sports, lifestyle magazines as well as newspaper articles, annotated by multiple annotators. These collections are available for research purposes upon request to the article's authors. 
	
	\subsection{Keyphrase Extraction Software} 
	\label{software}
	
	Both commercial and free software is developed for keyphrase extraction. In this section, we briefly present available free keyphrase extraction software (Section \ref{free_software}) as well as some well-known commercial keyphrase extraction APIs (Section \ref{commercial_software}).  
	
	\subsubsection{Free Software}  
	\label{free_software}
	
	Table \ref{tbl:software} presents free keyphrase extraction software along with some useful features, such as the implemented methods, supported languages and implementation language. There are software packages that include only one keyphrase extraction method (Maui, KEA, seq2seq, TextRank, YAKE, RAKE, TopicCoRank) and others that have more than one methods implemented (PKE - Python Keyphrase Extraction \citep{DBLP:conf/coling/Boudin16}, the KE package, Sequential Labeling, CiteTextRank). As far as the supported languages by each software concerned, RAKE supports English, French and Spanish, TopicCorank supports English and French, KEA and Maui support English, French, German and Spanish, whereas TextRank supports additional languages,  e.g., Swedish, Danish, Dutch, Italian etc. Although the default language in PKE and YAKE is English, keyphrases can also be extracted in other languages by setting the language parameter to the desired language. 
	
	\subsubsection{Commercial Software}  
	\label{commercial_software}
	
	As far as the commercial keyphrase extraction software concerned, to the best of our knowledge, Google Cloud Natural Language API\footnote{\url{https://cloud.google.com/natural-language}} has not included a service devoted to keyphrase extraction. However, an interesting feature of this API is the one of \textit{entity recognition} which identifies entities and labels by types, such as person, organization, location, event, product, and media. Furthermore, our empirical analysis presented in Table \ref{tbl:exact_match_results_APIs} shows that the entity recognition feature provided by Google's API is an alternative option, as it returns satisfactory results compared to the state-of-the-art methods of the keyphrase extraction task. 
	In the same direction, the TextRazor API\footnote{\url{https://www.textrazor.com/}} offers \textit{entity recognition} service that besides the classic confidence score (confidence related to the validity of the returned entity), it also gives the relevance score, which shows the relevance of the returned entity to the source text. 
	Microsoft offers its commercial software for keyphrase extraction via Microsoft's Text Analytics APIs\footnote{\url{https://azure.microsoft.com/en-us/services/cognitive-services/text-analytics}}. 
	Moreover, Aylien Text Analysis API\footnote{\url{https://docs.aylien.com/s}} performs a variety of complex NLP tasks on documents including keyphrase extraction, 
	whereas, IBM and Amazon offer their commercial solutions for advanced text analysis as well as keyphrase extraction via Watson Natural Language Understanding API\footnote{\url{https://www.ibm.com/watson/services/natural-language-understanding/}}. 
	and the Comprehend API\footnote{\url{https://aws.amazon.com/comprehend/}}. 
	
	Google\footnote{\url{https://cloud.google.com/natural-language/docs/languages}}, TextRazor\footnote{\url{https://www.textrazor.com/languages}}, IBM\footnote{\url{https://console.bluemix.net/docs/services/natural-language-understanding/language-support.html}}, and Microsoft Text Analytics\footnote{\url{https://docs.microsoft.com/en-us/azure/cognitive-services/text-analytics/text-analytics-supported-languages}} APIs have a wide range of supported languages, however, Amazon Comprehend API performs direct text analysis only on English and Spanish texts. Particularly, Amazon Comprehend proposes the conversion of the text of an unsupported language to English or Spanish via the Amazon Translate, and then uses Amazon Comprehend to perform text analysis. Finally, the Aylien Text Analysis\footnote{\url{https://docs.aylien.com/textapi/\#language-support}} API supports the following 6 languages regarding the keyphrase extraction feature: English, German, French, Italian, Spanish and Portuguese.
	
	In Section \ref{unsupevised-APIs}, Table \ref{tbl:exact_match_results_APIs} shows the performance of the commercial APIs discussed above on 5 popular datasets of the task. We should note that this empirical study is conducted in the context of the survey on the task using very domain-specific texts from keyphrase extraction data collections. Thus, \textit{such type of evaluation of commercial general purpose APIs, whose internal working is not actually known, should not be considered as a positive or negative attitude in favor of the APIs with high performance on the datasets}.

	\begin{table}[H]
		\centering
		\arrayrulecolor{black}\begin{threeparttable}
			\begin{tabular}{|>{\centering\arraybackslash}m{2cm}|>{\centering\arraybackslash}m{2cm}|>{\centering\arraybackslash}m{4cm}|>{\centering\arraybackslash}m{2cm}|}
				\hline
				\hiderowcolors
				Name              & Impl. Lang.                                                       & Methods & Languages                                                                                                                      \\ \hline
				\href{https://github.com/zelandiya/maui}{Maui}             & Java                                                           & \cite{medelyan2009human}   &   multilingual                                                                                     \\ \hline
				\href{https://github.com/LIAAD/yake}{YAKE}             & Python                                                           & \cite{YAKE2018Campos}   &   multilingual                                                                                     \\ \hline
				\href{https://github.com/adrien-bougouin/KeyBench}{TopicCoRank}             & Python                                                           & \cite{GraphCoR2016Bougouin}   &   English/French                                                                                     \\ \hline
				
				\href{https://github.com/zelandiya/RAKE-tutorial}{RAKE}             & Python                                                           & \cite{rose2010automatic}   &   multilingual                                                                                     \\ \hline
				
				\href{https://github.com/turian/kea-service}{KEA}               & \shortstack{Java\\Python wrapper} &  \cite{witten1999kea}   & multilingual                                                                                   \\ \hline
				\href{https://github.com/boudinfl/pke}{PKE}               & Python                                                         &  \shortstack{TfIdf, \cite{witten1999kea}\\ \cite{DBLP:conf/semeval/NguyenL10}\\\cite{DBLP:journals/is/El-BeltagyR09}\\ \cite{wan+xiao2008}\\ \cite{bougouin2013topicrank}\\ \cite{Boudin18Multipartite} \\ \cite{DBLP:conf/acl/FlorescuC17}\\\cite{DBLP:conf/www/SterckxDDD15}}     &  multilingual                                                                            \\ \hline
				\href{https://github.com/memray/seq2seq-keyphrase}{seq2seq} & Python                                                         &   \cite{meng2017deep}   &  English                                                                            \\ \hline
				\href{http://www.hlt.utdallas.edu/~saidul/code.html}{KE package}        & C++                                                            &   \shortstack{TfIdf, \cite{wan+xiao2008}\\ \cite{mihalcea+tatau2004} \\ \cite{wan+xiao2008}}    &    English\tnote{*}                                                                    \\ \hline
				\href{https://github.com/davidadamojr/TextRank}{TextRank}          & Python                                                         &  \cite{mihalcea+tatau2004}    &               multilingual        \\ \hline
				\href{https://www.dropbox.com/s/x8l7h2iatu54dne/aaai17distribv1.tgz?dl=0}{Sequential Labeling} & Java                                                         &  \shortstack{\cite{witten1999kea}\\ \cite{DBLP:conf/aaai/GollapalliLY17}\\\cite{caragea2014citation}\\\cite{medelyan2009human}}   &               English\tnote{*}        \\ \hline
				
				\href{https://www.dropbox.com/s/sb9nu817nyhbn9m/kpshare.tgz?dl=0}{CiteTextRank} & Java &  \shortstack{TfIdf,
					\cite{mihalcea+tatau2004} \\
					\citep{wan+xiao2008} \\
					\citep{gollapalli2014extracting}} & English\tnote{*}\\ \hline
			\end{tabular}
			
			\begin{tablenotes}
				\item[] \textit{The * symbol indicates that no other supported languages are explicitly mentioned.}
			\end{tablenotes}
		\end{threeparttable}
		
		\caption{Free keyphrase extraction software along with some useful features, such as the implementation language, implemented methods and supported languages.}
		\label{tbl:software}
	\end{table}

	\section{Empirical Evaluation Study}
	\label{comparative-eval}
	
	One of the biggest issues for keyphrase extraction remains the reliable evaluation measures/approaches and an analysis of the shortcomings by existing ones. In this section, we present an empirical study among commercial APIs  (Section \ref{unsupevised-APIs}) and state-of-the-art as well as popular/recent unsupervised methods (Section \ref{results}) using different evaluation approaches (exact, partial, manual evaluation) and measures (F$_1$-measure, MAP). We investigate their relation to see whether different evaluation approaches could lead to different conclusions concerning the performance comparison (Section \ref{exact_partial}). Furthermore, we investigate the way in which the different keyphrase extraction systems and measures behave with respect to the different gold standards (Section \ref{eval_gold_standards}).  We also present a qualitative analysis that highlights the differences between various evaluation approaches (Section \ref{qualitative-results}).

	\subsection{Experimental Setup}
	\label{setup}
	Popular commercial APIs commonly used to extract keyphrases from texts (discussed in Section \ref{commercial_software}) as well as unsupervised methods from all basic categories are included in our empirical study, in Sections \ref{unsupevised-APIs} and \ref{results}, respectively. However, the experimental results of the commercial general purpose APIs, should not be considered as a positive or negative attitude in favor of the APIs (see clarification in Section \ref{commercial_software}). The unsupervised methods that participate in the experimental comparison are (i) the statistical-based methods KP-Miner (KPM), YAKE and TfIdf, (ii) the graph-based methods SingleRank (SR), TopicRank (TR), MultipartiteRank (MR), PositionRank (PR), RAKE and (iii) the RVA method that utilizes word embeddings. 
	Finally, a variant of TfIdf (\textit{TfIdf2}) is developed for the evaluation purposes that combines popular statistical and positional heuristics of the task. Specifically, TfIdf2 has slight differences compared to TfIdf as it ranks the phrases in a particular document according to their TfIdf score by keeping only the candidates that consist of terms with term frequency greater than 3 and belong to the set of the first 100 nouns or adjectives that appear in the text.
	
	Many keyphrase extraction methods have been evaluated using texts from scientific publications (see Table \ref{KP_datasets}). We choose 3 popular datasets that contain full-text publications from the computer science domain, i.e, Krapivin, Semeval2010 (Semeval) and NUS. Additionally, we are interested in studying the performance of keyphrase extraction methods in short texts, hence, we experimented with the well-known Inspec dataset that contains abstracts from journal papers from the disciplines Computers and Control, and Information Technology. Finally, we use the 500N-KPCrowd news dataset, which contains broadcast news stories from 10 different categories, to see how the systems perform on texts of general domain. Since the SemEval and NUS datasets come with both author and reader assigned keyphrases, we use both sets of keyphrases as a gold standard in the experimental study of Sections \ref{unsupevised-APIs} and \ref{results}. We should note that all methods extract keyphrases from the full-text articles except for RVA which uses the full-text to create the vector representation of the words and then returns keyphrases only from the abstract. Furthermore, we give as input to the APIs only the abstracts of the articles in order to avoid additional costs. However, concerning the 500N-KPCrowd articles, whose average length is approximately 2607 characters, we give as input to the APIs the first 1500 characters of each text document where at least 75\% of the keyphrases appear. For all competing methods we use the default parameters proposed in the corresponding articles. 
	
	According to Fig. \ref{fig:popularity_evaluation}, Precision/Recall/F$_1$-measure have indeed been used in the majority of the related work. As most methods have been evaluated based on these measures, we choose the F$_1$-measure as the main measure of our study. We adopt two different automatic evaluation approaches for the calculation of the F$_1$-score: i) the strict exact match evaluation approach, which computes the F$_1$-score between golden keyphrases and candidate keyphrases, after stemming and removal of punctuation marks, and ii) the simple and more loose partial match evaluation, which calculates the F$_1$-measure between the set of words found in all golden keyphrases and the set of words found in all extracted keyphrases after stemming and removal of punctuation marks. We also provide the Mean Average Precision (MAP) score to evaluate the ranking of the returned keyphrases according to the exact match evaluation. 
	
	We compute F$_1$@10, F$_1$@20, MAP@10, MAP@20, as accuracy at the top of the ranking is more important in typical applications. We used the NLTK Python suite for preprocessing, and the PKE toolkit \citep{DBLP:conf/coling/Boudin16} for the implementations of most unsupervised keyphrase extraction methods. For YAKE\footnote{\url{https://github.com/LIAAD/yake}} and RAKE\footnote{\url{https://github.com/zelandiya/RAKE-tutorial}} methods we use their corresponding implementations that are available on Github.
	
	\subsection{Commercial APIs}
	\label{unsupevised-APIs}

	
	Comparing the commercial APIs' performance on the 4 datasets of scientific publications, i.e., Semeval, NUS, Krapivin, and Inspec, we find the IBM API in the first place followed by Google API in the second place except for Semeval at F$_1$@10 and NUS where Aylien API and Google API win, respectively. Aylien API and Textrazor API are quite competitive between each other with the second one to perform better on Krapivin dataset as well as at F$_1$@20 on NUS and Inspec datasets. Amazon API is the worst-performing among the commercial APIs in all datasets.
	
	Moreover, we compare the keyphrase extraction commercial APIs (including entity extraction services) based on their performance on the 500N-KPCrowd dataset, which includes hundreds of text documents of news stories, i.e., a domain other than that of scientific publications. Aylien API and Google API achieve high F$_1$-scores with values that lie quite close to each other. In the following ranking positions, we see IBM API, Textrazor API and Amazon API, respectively. For the sake of completeness, Appendix
	\ref{APIs_map_partial_eval} (Tables \ref{tbl:apis_map_results} and \ref{tbl:apis_partial_match_results}) reports the APIs' performance based on the MAP measure using the exact match evaluation and the F$_1$-score using the partial match evaluation.

	\begin{table}[H]
		\centering
		\scalebox{1.0}{
			\begin{tabu}{|c|c|c|c|c|c|c|c|c|c|c|}
				\hiderowcolors
				\hline
				\multirow{2}{*}{}  & \multicolumn{2}{c|}{Semeval}    & \multicolumn{2}{c|}{NUS}     & \multicolumn{2}{c|}{Krapivin}   & \multicolumn{2}{c|}{Inspec}      & \multicolumn{2}{c|}{500N-KPCrowd}     \\ \cline{1-11} 
				APIs & @10          & @20          & @10          & @20          & @10          & @20          & @10          & @20          & @10          & @20          \\ \hline
				IBM               & 0.100          & \textbf{0.118}          & 0.115          & 0.117          & \textbf{0.114}          & \textbf{0.106}          & \textbf{0.256}          & \textbf{0.270}          & 0.081          & 0.133          \\ 
				GOOGLE            & 0.089          & 0.106          & \textbf{0.135}          & \textbf{0.141}          & 0.106          & 0.096          & 0.168          & 0.192          & \textbf{0.143}          & 0.210          \\ 
				Amazon            & 0.037          & 0.062          & 0.035          & 0.063          & 0.034          & 0.054          & 0.063          & 0.109          & 0.058          & 0.093          \\ 
				Textrazor         & 0.073          & 0.084          & 0.099          & 0.113          & 0.096          & 0.097          & 0.116          & 0.140          & 0.062          & 0.098          \\ 
				Aylien            & \textbf{0.101}          & 0.092          & 0.121          & 0.108          & 0.080          & 0.062          & 0.123          & 0.132          & \textbf{0.143}          & \textbf{0.229}          \\
				\hline
		\end{tabu}}
		\caption{$F_1$@10 and $F_1$@20 according to the exact match evaluation on 5 datasets.}
		\label{tbl:exact_match_results_APIs}
	\end{table}

	\subsection{Unsupervised Keyphrase Extraction Methods}
	\label{results}
	
	\subsubsection{Exact Match Evaluation Approach}
	\label{f1_map}
	
	\noindent \textbf{F$_1$-measure}
	
	According to the exact match (Table \ref{tbl:exact_match_results}), KPM outperforms all the other methods in the first 3 datasets, i.e., Semeval, NUS, and Krapivin (full-text scientific publications), by a large margin, usually followed by TfIdf2 (2nd) and TfIdf (3rd) over all top @N rankings except for Semeval dataset at F$_1$@10 where YAKE is 3rd. We see that the modified version of TfIdf (TfIdf2) achieves considerably higher performance than the original TfIdf. MR is usually in the 4th position for all F$_1$@N scores in all datasets. PR, TR and RAKE follow, alternating their ranking positions. Particularly, PR achieves higher F$_1$@10, RAKE performs better at F$_1$@20, whereas TR has lower scores on the NUS dataset. In the next ranking positions, RVA achieves similar F$_1$@20 to PR and TR. SR is the worst-performing among the methods in all datasets.
	
	We benchmark the performance of all keyphrase extraction methods on the Inspec dataset, which contains short texts (abstracts of scientific articles). We exclude the RVA method from the comparison, as it needs the full-text of a document to create the vector representation of the words. We notice that the previous worst-performing SR method outperforms all the other methods. PR is the second one and MR, TR are in the third and fourth positions, respectively. Next in the ranking is RAKE that beats YAKE and TfIdf, which have equivalent performances. However, in this dataset TfIdf2 achieves lower F$_1$-scores than TfIdf.

	\begin{table}[H]
		\centering
		\scalebox{1.0}{
			\begin{tabu}{|c|c|c|c|c|c|c|c|c|c|c|}
				\hiderowcolors
				\hline
				\multirow{2}{*}{}  & \multicolumn{2}{c|}{Semeval}    & \multicolumn{2}{c|}{NUS}     & \multicolumn{2}{c|}{Krapivin}   & \multicolumn{2}{c|}{Inspec}      & \multicolumn{2}{c|}{500N-KPCrowd}     \\ \cline{2-11} 
				F$_1$  & @10          & @20          & @10          & @20          & @10          & @20          & @10          & @20          & @10          & @20          \\ \hline
				YAKE              & 0.160          & 0.169          & 0.188          & 0.180          & 0.124          & 0.109          & 0.197          & 0.212          & 0.107          & 0.168          \\ 
				TfIdf             & 0.154          & 0.176          & 0.201          & 0.205          & 0.126          & 0.113          & 0.197          & 0.212          & 0.179          & \textbf{0.243} \\ 
				TfIdf2            & 0.172          & 0.191          & 0.229          & 0.217          & 0.156          & 0.138          & 0.184          & 0.193          & \textbf{0.180} & 0.233          \\ 
				KPM           & \textbf{0.208} & \textbf{0.219} & \textbf{0.259} & \textbf{0.243} & \textbf{0.190} & \textbf{0.161} & 0.107          & 0.106          & 0.143          & 0.178          \\ \hline
				RAKE              & 0.114          & 0.147          & 0.134          & 0.142          & 0.091          & 0.096          & 0.216          & 0.233          & 0.064          & 0.066          \\ 
				MR                & 0.146          & 0.161          & 0.147          & 0.149          & 0.112          & 0.100          & 0.245          & 0.269          & 0.156          & 0.224          \\ 
				TR                & 0.134          & 0.142          & 0.126          & 0.118          & 0.099          & 0.086          & 0.235          & 0.249          & 0.148          & 0.209          \\ 
				PR                & 0.131          & 0.127          & 0.146          & 0.128          & 0.102          & 0.085          & 0.253          & 0.273          & 0.145          & 0.206          \\ 
				SR                & 0.036          & 0.053          & 0.044          & 0.063          & 0.026          & 0.036          & \textbf{0.278} & \textbf{0.295} & 0.096          & 0.164          \\ \hline
				RVA               & 0.096          & 0.125          & 0.096          & 0.115          & 0.093          & 0.099          & -          & -          & -          & -          \\ \hline
		\end{tabu}}
		\caption{F$_1$@10 and F$_1$@20 according to the exact match evaluation approach of all methods for all datasets.}
		\label{tbl:exact_match_results}
	\end{table}

	Moreover, we compare the keyphrase extraction methods based on their performance on the 500N-KPCrowd dataset, which includes hundreds of text documents of news stories, i.e., a domain other than that of scientific publications. Again, we exclude the RVA method from the comparison, as it needs an abstract/summary of the text document to complete the keyphrase extraction process. TfIdf2 and TfIdf outperform all the other methods followed by MR. TR and PR are in the 3rd position achieving F$_1$-scores with quite close values to each other. In the following ranking positions, there are KPM, YAKE and SR, respectively, whereas the worst-performing method is RAKE. 
	
	\noindent \textbf{MAP}
	
	In terms of MAP that is computed in the context of the exact match approach (Table \ref{tbl:map_results}), the non-commercial methods' ranking on the Semeval, NUS, and Krapivin datasets (full-text scientific publications) is quite similar to the one inferred by the F$_1$-scores described above. Once more, the Inspec dataset is used to benchmark the methods on short texts. Particulary, in the first 4 ranking positions we come across the same methods as before, i.e., based on the F$_1$-score, but in another order. MR and SR share the first position, followed by TR, and PR, respectively. The same holds for the next ranking positions where we find YAKE/RAKE, TfIdf, TfIdf2, and KPM, respectively. Furthermore, the keyphrase extraction methods follow a different ranking order based on their MAP scores on the 500N-KPCrowd dataset. Once again, TfIdf2 and TfIdf outperform all the other methods, whereas, MR, TR, PR follow in the next places but without notable differences. The lower performances are achieved by YAKE, RAKE and SR, respectively. 
	
	\noindent \textbf{Summary}
	
	To sum up, the results of F$_1$ and MAP scores on the datasets of the full-text scientific publications (Semeval, Krapivin, and NUS) according to the exact match evaluation approach confirm the superiority of the statistical methods (KPM, TfIdf2, TfIdf, and YAKE in that order) over the graph-based ones, as the full-texts contain sufficient statistical information to distinguish keyphrases from non-keyphrases. At the same time, the large amount of information makes harder the detection of keyphrases by the graph-based methods. Furthermore, graph-based methods (SR, MR, PR, TR and RAKE) outperform the statistical methods in the task of keyphrase extraction from short scientific texts (results on Inspec dataset). Finally, the classical statistical methods of TfIdf and its variation TfIdf2 ranked first with respect to F$_1$ and MAP scores in the task of keyphrase extraction from news articles, usually followed by the quite decent results of MR and KPM. We see that the modified version of TfIdf (TfIdf2) achieves considerably higher performance than the original TfIdf in all datasets except for the Inspect dataset where both methods do not perform well as the abstracts do not contain enough text to enable the separation of keyphrases from non-keyphrases. However, for the graph-based methods, the opposite is observed. In particular, abstracts capture adequately the co-occurrence (proximity) of words that is necessary for graph creation, avoiding at the same time the noise of the full-texts. Additionally, the threshold of TfIdf2 regarding the term frequency of a candidate keyword (greater than 3) impinges the keyphrase extraction process from short texts where the term frequencies range in low values. Once more, the high performance of the MR method compared to the rest graph-based methods in most datasets implies that the graph-based methods can benefit by integrating different types of information, such as topical, positional and statistical information.

	\begin{table}[H]
		\centering
		\scalebox{1.0}{
			\begin{tabu}{|c|c|c|c|c|c|c|c|c|c|c|}
				\hline
				\hiderowcolors
				\multirow{2}{*}{} & \multicolumn{2}{c|}{Semeval}    & \multicolumn{2}{c|}{NUS}     & \multicolumn{2}{c|}{Krapivin}   & \multicolumn{2}{c|}{Inspec}      & \multicolumn{2}{c|}{500N-KPCrowd}     \\ \cline{2-11} 
				MAP   & @10         & @20         & @10         & @20         & @10         & @20         & @10         & @20         & @10         & @20         \\ \hline
				YAKE              & 0.106          & 0.065          & 0.104          & 0.063          & 0.044          & 0.025          & 0.109          & 0.070          & 0.151          & 0.118          \\ 
				TfIdf             & 0.097          & 0.062          & 0.119          & 0.074          & 0.045          & 0.026          & 0.100          & 0.066          & 0.355          & \textbf{0.257} \\ 
				TfIdf2            & 0.110          & 0.070          & 0.136          & 0.083          & 0.056          & 0.033          & 0.094          & 0.056          & \textbf{0.365} & 0.253          \\ 
				KPM           & \textbf{0.144} & \textbf{0.090} & \textbf{0.160} & \textbf{0.098} & \textbf{0.073} & \textbf{0.042} & 0.058          & 0.029          & 0.302          & 0.197          \\ \hline
				RAKE              & 0.058          & 0.041          & 0.062          & 0.040          & 0.023          & 0.015          & 0.106          & 0.073          & 0.141          & 0.074          \\ 
				MR                & 0.093          & 0.058          & 0.076          & 0.047          & 0.040          & 0.023          & \textbf{0.153} & 0.098          & 0.253          & 0.194          \\ 
				TR                & 0.081          & 0.049          & 0.069          & 0.040          & 0.034          & 0.020          & 0.144          & 0.088          & 0.243          & 0.179          \\ 
				PR                & 0.073          & 0.042          & 0.068          & 0.039          & 0.030          & 0.017          & 0.136          & 0.091          & 0.241          & 0.176          \\ 
				SR                & 0.013          & 0.009          & 0.012          & 0.009          & 0.005          & 0.003          & \textbf{0.153}          & \textbf{0.103} & 0.096          & 0.091          \\ \hline
				RVA               & 0.050          & 0.035          & 0.036          & 0.026          & 0.026          & 0.017          & -          & -          & -          & -          \\ \hline
		\end{tabu}}
		\caption{MAP@10 and MAP@20 according to the exact match evaluation approach of all methods for all datasets.}
		\label{tbl:map_results}
	\end{table}

	\subsubsection{Partial Match Evaluation Approach}
	\label{pf1}
	
	According to the partial match evaluation approach proposed by \cite{DBLP:conf/ecir/RousseauV15} (Table \ref{tbl:partial_match_results}) KPM outperforms all the other methods in the first 3 datasets, i.e., Semeval, NUS, and Krapivin (full-text scientific publications), by a large margin, usually followed by TfIdf2 (2nd) in most cases. MR often follows in the next position in the ranking achieving quite high scores in Semeval and NUS at F$_1$@10, while YAKE performs better than MR in Krapivin and NUS at F$_1$@20. TfIdf, PR, TR, and RVA are quite competitive between each other and are usually in the next places alternating their ranking among the datasets. Finally, RAKE and SR are the worst-performing among the state-of-the-art methods in all datasets. 
	The experimental results regarding the Inspec dataset show that the previous worst-performing SR method outperforms all the other methods. Next in the ranking order are MR (2nd) and TR (3rd), followed by PR, RAKE, TfIdf, YAKE that have similar F$_1$-scores. The worst ranking scores are achieved by KPM and TfIdf2 methods. Furthermore, on the news dataset (500N-KPCrowd) the SR outperforms all the other methods, followed by TR (2nd) and MR (3rd). Next in the ranking order are PR, YAKE and TfIdf, respectively, without considerable differences. In the last ranking positions, we see the rest statistical methods (YAKE, TfIdf2, RAKE).

	\begin{table}[H]
		\centering
		\scalebox{1.0}{
			\begin{tabu}{|c|c|c|c|c|c|c|c|c|c|c|}
				\hline
				\hiderowcolors
				\multirow{2}{*}{} & \multicolumn{2}{c|}{Semeval}    & \multicolumn{2}{c|}{NUS}     & \multicolumn{2}{c|}{Krapivin}   & \multicolumn{2}{c|}{Inspec}      & \multicolumn{2}{c|}{500N-KPCrowd}     \\ \cline{2-11} 
				pF$_1$  & @10         & @20         & @10         & @20         & @10         & @20         & @10         & @20         & @10         & @20         \\ \hline
				YAKE              & 0.319          & 0.381          & 0.400          & 0.414          & 0.345          & 0.340          & 0.479          & 0.522          & 0.256          & 0.345          \\ 
				TfIdf             & 0.305          & 0.365          & 0.373          & 0.394          & 0.309          & 0.306          & 0.474          & 0.540          & 0.247          & 0.366 \\ 
				TfIdf2            & 0.345          & 0.413          & 0.436          & 0.457          & 0.392          & 0.390          & 0.443          & 0.478          & 0.217          & 0.288          \\ 
				KPM           & \textbf{0.389} & \textbf{0.451} & \textbf{0.468} & \textbf{0.482} & \textbf{0.422} & \textbf{0.402} & 0.305          & 0.311          & 0.169          & 0.210          \\ \hline
				RAKE              & 0.297          & 0.334          & 0.335          & 0.327          & 0.271          & 0.248          & 0.494          & 0.515          & 0.077          & 0.079          \\ 
				MR                & 0.362          & 0.401          & 0.407          & 0.383          & 0.342          & 0.303          & 0.515          & 0.553          & 0.315          & 0.421          \\ 
				TR                & 0.345          & 0.378          & 0.375          & 0.351          & 0.312          & 0.277          & 0.509          & 0.540          & 0.318          & 0.423          \\ 
				PR                & 0.294          & 0.317          & 0.371          & 0.350          & 0.342          & 0.302          & 0.494          & 0.538          & 0.264          & 0.377          \\ 
				SR                & 0.284          & 0.297          & 0.322          & 0.309          & 0.290          & 0.256          & \textbf{0.565}          & \textbf{0.586} & \textbf{0.328}          & \textbf{0.424}          \\ \hline
				RVA               & 0.332          & 0.365          & 0.374          & 0.380          & 0.348          & 0.337          & -          & -          & -          & -          \\ \hline
		\end{tabu}}
		\caption{F$_1$@10 and F$_1$@20 according to the partial match evaluation approach (pF$_1$) of all methods for all datasets.}
		\label{tbl:partial_match_results}
	\end{table}

	\noindent \textbf{Summary}
	
	To conclude, the results of F$_1$-scores with respect to the partial match evaluation on the datasets of the full-text scientific publications (Semeval, Krapivin, and NUS) confirm the superiority of the statistical methods (KPM, TfIdf2, and YAKE in that order) over the graph-based ones. The large amount of information in the full-texts benefits statistical methods, facilitating the separation between keyphrases and non-keyphrases, whereas the redundant information existing in full-texts does not facilitate the graph-based methods to model the correlations between the words. A text document contains hundreds/thousands of words whereas the number of keywords is very small. Hence, there is a number of words that are not keywords but play an important role in the description of the document’s topics (expressed by the keywords), impeding the keyphrase extraction process. However, once more, the high performance of the MR method compared to the rest of the graph-based methods implies the potential of the graph-based methods to exploit various types of information. Furthermore, RVA's performance witnesses the effectiveness of keyphrase extraction from scientific publications using only their titles and abstracts. \cite{papagiannopoulou2018local} conducted experiments using the same evaluation approach \citep{DBLP:conf/ecir/RousseauV15} and two versions of each  unsupervised keyphrase extraction method, one with the title/abstract and one with the full-text of each article. The abstract version of both RVA and the graph-based methods is better than the fulltext version according to the partial match evaluation approach, possibly due to the redundancy which is included in the fulltexts. The opposite holds for TfIdf, since abstracts do not contain enough text to enable the separation of keyphrases from non-keyphrases. In addition, the use of a local word vector representation that would be more oriented to the keyphrase detection than the GloVe (which is used by RVA) seems to be promising. Furthermore, similar to the exact match evaluation, graph-based methods (SR, MR, and TR) outperform the statistical methods in the task of keyphrase extraction from short scientific texts (results of Inspec dataset), since the abstracts seem to capture adequately the required context of words for the graph creation, avoiding the noise of the full-texts. Finally, the graph-based methods (SR, TR, MR, and PR) are highly ranked in the task of keyphrase extraction from news articles. We see that the modified version of TfIdf (TfIdf2) achieves considerably higher performance than the original TfIdf in the full-text publication datasets, whereas on the Inspect and news datasets (500N-KPCrowd) both methods do not perform well.

	\subsubsection{General Remarks}
	\label{general_remarks}
	Both evaluation strategies confirm the superiority of the statistical methods (KPM, TfIdf2) over the graph-based ones in the task of keyphrase extraction from full-text scientific publications. Yet, the partial evaluation benefits the graph-based methods ranking them in higher positions compared to the exact match evaluation. This is also evident from the results in the 500N-KPCrowd dataset where the graph-based methods are highly ranked with respect to the partial F$_1$-scores. At the same time, the partial evaluation strategy seems to give more reasonable results, as recent methods outperform the baseline of the task (TfIdf) in more cases. In addition, the high performance of the MR method compared to the rest graph-based methods is confirmed by both evaluation approaches. Finally, it would be interesting to see the development of a TfIdf variant, similar to TfIdf2, which would be universally adopted as the baseline of the task. 
	
	In addition, we provide some guidelines to practitioners. In case of keyphrase extraction from fulltexts of scientific articles, statistical methods are a decent option, as they offer a good balance between performance and computation time compared to the graph-based methods that achieve lower accuracy and have higher computation time (due to PageRank that runs in most of the graph-based methods). However, in case of keyphrase extraction from short texts, graph-based methods are suggested as more appropriate, since their computation cost is not considerable due to the short length of texts, and they perform better than the statistical methods. Particularly, SingleRank is the best option in comparison with the more recent methods of TR, PR and MR. It seems that in short texts the information coming from the graph-of-words and co-occurrence statistics is sufficient. There is no need for any additional positional or topical information, as the words do not appear in a long text (position does not play an important role in documents of a few lines) and the frequencies of similar keyphrase candidates (that are grouped in topics by TR, MR) are quite low due to the limited text. For text documents that are not from a specific domain both statistical and graph-based methods achieve good performance according to the exact and partial evaluation approaches. So, it is up to the practitioner's choice which category of methods will use, based on the computation cost that is related to the length of the text.
	
	In Section \ref{exact_partial}, we go one step further from describing intuitively the properties of the two evaluation approaches. More specifically, we investigate their properties (measuring (i) the correlation between the F$_1$-scores calculated with respect to the exact/partial evaluation approaches and the F$_1$-scores based on the manual evaluation, (ii) how close their actual values are regarding human computed F$_1$-scores, and (iii) testing the scores' distributions). In this way, we  recommend which evaluation strategy should be preferred in certain situations, providing more meaningful insights. 
	
\subsection{Exact vs Partial Matching}
\label{exact_partial}
	
The majority of the keyphrase extraction articles that we reviewed present experimental results based on exact matching of predicted and golden keyphrases. An important limitation of this evaluation approach, however, is that it penalizes methods even if they predict keyphrases that are semantically similar to the golden ones  \citep{DBLP:conf/ecir/RousseauV15,DBLP:conf/adc/WangLM15}. On the other hand, partial matching rewards methods when they predict words that appear in golden keyphrases, even when their predicted keyphrases are not themselves appropriate for the corresponding article. In order to obtain a deeper understanding of the pros and cons of exact and partial matching, this section empirically investigates their relationship to manual evaluation.  

We randomly selected 50 full-text articles from the Krapivin dataset and extracted keyphrases from them using one statistical method (KPM) and one graph-based method (MR). We calculated the F$_1$@10 of these methods for each article based on exact and partial matching, as well as based on manual evaluation according to the following process. Given an article, each of the top 10 predicted keyphrases was evaluated as true positive if both the authors of this review paper considered it relevant to the article, after reading the whole article, focusing on the abstract and introduction for articles outside our expertise. 

We use three different statistical tools to study how the F$_1$@10 scores obtained via exact and partial matching are related to the F$_1$@10 scores obtained via manual evaluation: (i) The Spearman coefficient in order to study their correlation, (ii) the Wilcoxon signed-rank non-parametric test at a significance level of 0.05 for assessing whether they come from the same distribution, and (iii) the Mean Squared Error (MSE), in order to measure their distance. Table \ref{tbl:exact_partial_manual} shows the results, including an additional evaluation approach that simply averages the F$_1$@10 scores of exact and partial matching. 

\begin{table}[H]
	\centering
	\scalebox{0.9}{
		\begin{tabu}{|c|c|c|c|c|c|c|c|c|c|}
			\hline
			\hiderowcolors
			\multirow{2}{*}{} & \multicolumn{3}{c|}{Spearman}              & \multicolumn{3}{c|}{Wilcoxon}            & \multicolumn{3}{c|}{MSE}                   \\ \cline{2-10} 
			& Exact            & Partial    & Average       & Exact & Partial             & Average       & Exact   & Partial          & Average          \\ \hline
			KPM           & \textbf{0.63692} & 0.34969 & \textit{0.49194} & $\approx$0.00000     & \textbf{0.09023} & \textit{0.00011} & 0.05981 & \textit{0.03122} & \textbf{0.02635} \\ 
			MR                & \textbf{0.34413} & 0.20348 & \textit{0.25634} & $\approx$0.00000     & \textbf{0.32243} & \textit{0.00097} & 0.05465 & \textit{0.03241} & \textbf{0.02814} \\ \hline
	\end{tabu}}
	\caption{Spearman correlation coefficient, Wilcoxon signed-rank test $p$ value, and Mean Squared Error (MSE) between the F$_1$@10 scores obtained via manual evaluation and those obtained via exact/partial matching along with their average. The second best values are emphasized in italic typeface whereas the best values are in bold.}
	\label{tbl:exact_partial_manual}
\end{table}

	
We first see that the F$_1$@10 scores obtained via manual evaluation are more correlated to the ones obtained via exact matching than those obtained via partial matching for both keyphrase extraction methods. This is expected to a certain extend, as exact matching is a lower bound of manual evaluation: a human would definitely evaluate as true positive an extracted keyphrase that matches a golden keyphrase and might also evaluate as true positive a predicted keyphrase that doesn't match a golden keyphrase but is semantically equivalent to it. Partial matching on the other hand may lead to both lower F$_1$@10 scores when predicted keyphrases are representative of the article, but don't contain words that appear in the golden keyphrases, and higher F$_1$@10 scores when predicted keyphrases are not related to the article, yet contain words that appear in golden keyphrases. 

\begin{figure}[H]
	\centering
	\begin{subfigure}[b]{0.9\textwidth}
		\centering
		\includegraphics[width=\linewidth]{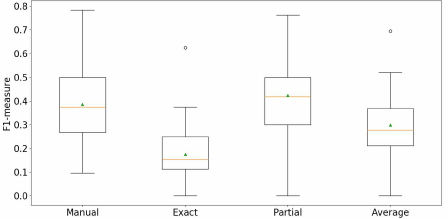}
		\caption{KPM}
	\end{subfigure}%
	\vspace{\floatsep}
	\begin{subfigure}[b]{0.9\textwidth}
		\centering
		\includegraphics[width=\linewidth]{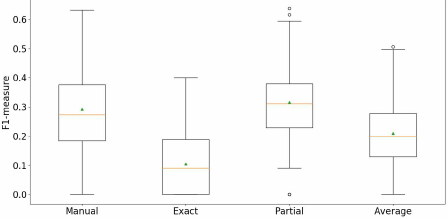}
		\caption{MR}
	\end{subfigure}
	\caption{Box plots of the F$_1$@10 scores obtained via manual evaluation, exact matching, partial matching and taking the average of exact and partial matching.}
	\label{fig:boxplots}
\end{figure}

The above reasoning further explains the results of the Wilcoxon test, which show us that the F$_1$@10 scores of exact matching do not come from the same distribution as those of manual evaluation, in contrast to the F$_1$@10 scores of partial matching. Fig. \ref{fig:boxplots} shows box plots of the F$_1$@10 scores obtained via manual evaluation, exact and partial matching, as well as their average. As expected, we notice that the scores of exact matching are coming from a distribution with a lower and more concentrated range of values, compared to those of manual evaluation. We also see that the scores of partial matching have similar statistics with that of the manual evaluation with a slightly higher median and interquartile range values. 

\begin{figure}[H]
	\centering
	\begin{subfigure}[b]{0.9\textwidth}
		\centering
		\includegraphics[width=\linewidth]{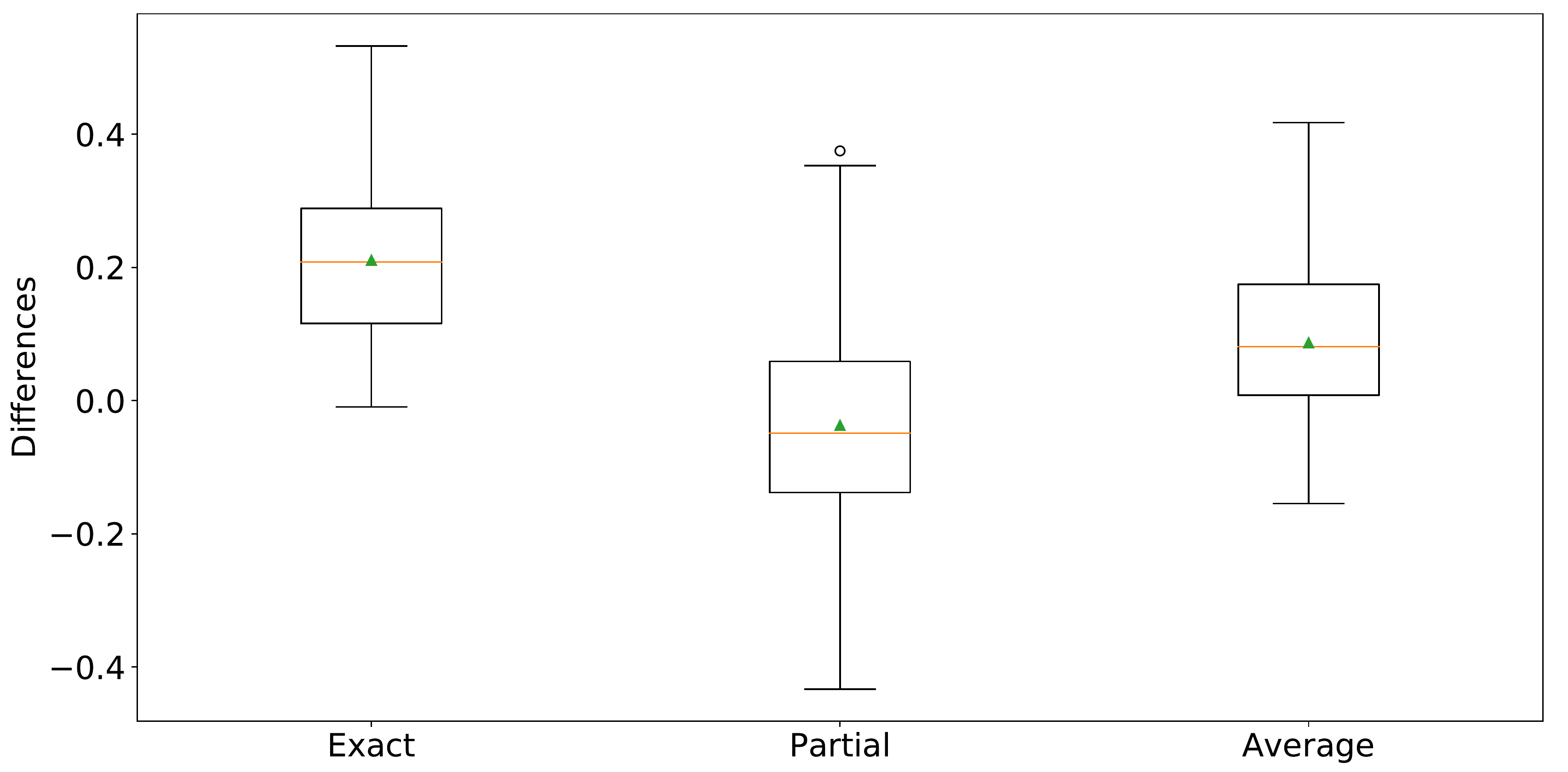}
		\caption{KPM}
		\label{subfig:kpm_plot}
	\end{subfigure}%
	\vspace{\floatsep}
	
	\begin{subfigure}[b]{0.9\textwidth}
		\centering
		\includegraphics[width=\linewidth]{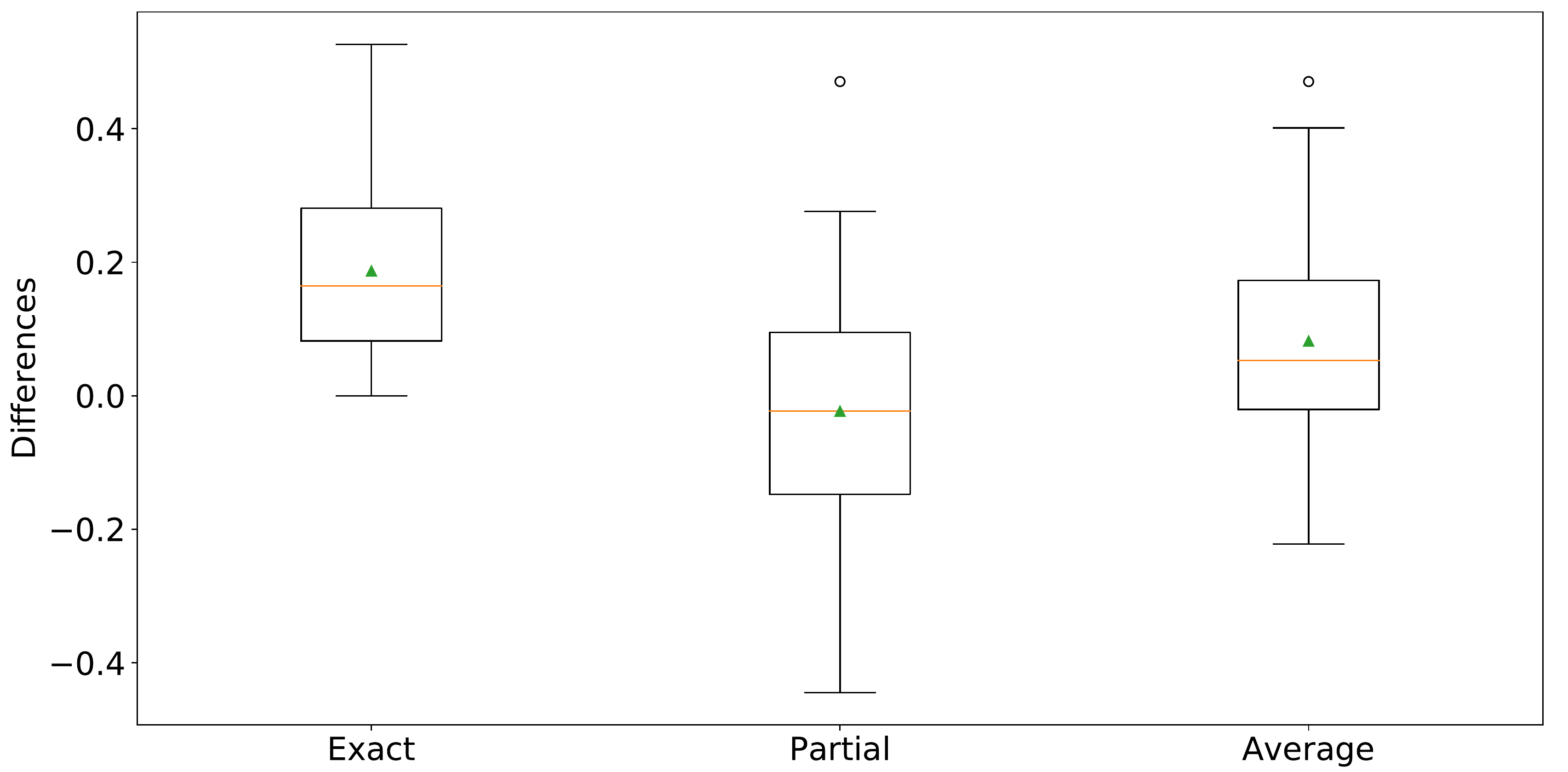}
		\caption{MR}
		\label{subfig:mr_plot}
	\end{subfigure}
	\caption{Distribution of differences between the F$_1$-scores based on the manual evaluation and the F$_1$-scores based on the exact (Exact), partial (Partial) and average (Average) evaluation approaches for the 50 manually evaluated documents given on the x axis.}
	\label{fig:differences}
\end{figure}

We now come to the most important aspect of our analysis, which concerns the MSE between manual evaluation and exact/partial matching. We see that partial matching has lower mean squared error than exact matching, while the lowest MSE is achieved by taking the average  F$_1$@10 scores of the two approaches. Figure \ref{fig:differences} shows box plots of the distribution of differences between the F$_1$-scores based on the manual evaluation and the F$_1$-scores based on the exact (Exact), partial (Partial) and average (Average) evaluation approaches for the 50 manually evaluated documents. As expected, we notice that the differences between manual and exact matching F$_1$-scores range to higher positive values as the exact matching is a lower bound of manual evaluation. We also see that the differences of between manual and partial matching F$_1$-scores have lower median/mean almost equal to zero and wide interquartile range values. Finally, the differences between manual and average matching F$_1$-scores have a slightly higher median/mean but more concentrated range of values around zero, compared to the rest types of differences (Exact, Partial). Figure \ref{fig:differences2} (see Appendix \ref{sec:differences}), shows an alternative view of the differences between the F$_1$-score values based on the manual evaluation and the F$_1$-score values calculated according to the exact (solid line), partial (dashed line) and average (dotted line) evaluation approaches for KPM and MR methods.

%
	
%

Our analysis suggests that researchers should consider the average of exact and partial matching for empirical comparison of keyphrase extraction methods. In addition, we stress the need for automatic evaluation approaches that take into account the semantic similarity between predicted and golden keyphrases. 


	\subsection{The Role of the Evaluation Gold Standards}
	\label{eval_gold_standards}
	
	There are keyphrase extraction datasets such as the Semeval and the NUS that come with both author and reader assigned keyphrases. In this section, we investigate the extent to which different gold evaluation standards affect the performance estimation of various keyphrase extraction systems, coming to completely different conclusions with respect to i) the rank of the methods regarding their performance or ii) the actual accuracy of a method. Table \ref{tbl:union_authors_readers} shows the F$_1$@10 scores using the exact match evaluation of the keyphrase extraction methods for the Semeval and NUS datasets based on 4 different evaluation gold standards: the union of the authors' and readers' keyphrases (Union), the authors' keyphrases (A), the readers' keyphrases (R) and the intersection of the authors' and readers' keyphrases (Inter.). In this part of the study, only the text documents that have both author and reader assigned keyphrases participate, i.e., 243 and 150 text documents from Semeval and NUS, respectively. 
	
	In all the keyphrase extraction methods, the highest F$_1$@10 scores appear when the union of the authors' and readers' keyphrases is regarded as the evaluation gold standard, as the higher the number of the golden keyphrases, the higher the probability of true positive cases in a system's output. On the contrary, the usage of the intersection between the authors' and readers' keyphrases as a gold standard leads to lower F$_1$@10 scores in almost all cases, as there are quite few keyphrases that are chosen by both authors and readers for a specific article. Authors' motivations differ from those of readers' ones, since authors usually aim not only to the summarization of their work via keyphrases but also to the promotion of their work considering phrases related to research trends etc. In addition, experimental results based on the F$_1$@20, MAP@10, and MAP@20 result in a similar conclusion to that one of the Table \ref{tbl:union_authors_readers}, i.e., the highest scores appear in the case of the evaluation gold standard that corresponds to the union of the authors' and readers' keyphrases.
	
	First of all, we observe differences in the ranking order of the methods based on the average F$_1$@10 scores across the documents of each dataset. Particularly, in the case of F$_1$@10 scores there are more evident differences in the NUS dataset. For example, the 5th position of the ranking belongs to MR regarding the union of the keyphrases as a gold evaluation standard. However, the gold evaluation standard of the authors' keyphrases gives the 5th ranking position to PR. A deeper insight to the correlation as well as the scores' distributions of the 4 evaluation gold standards is presented below.

	First, we utilize the Spearman's correlation to assess the monotonic relationship (linear or not) between the union of the authors'-readers' keyphrases and the authors' keyphrases (Union-A) as well as the readers' keyphrases (Union-R). Moreover, we investigate the correlation between the authors' and the readers' keyphrases (A-R) (see Tables \ref{tbl:semeval_union_authors_readers_spearman_wilcoxon} and \ref{tbl:nguyen_union_authors_readers_spearman_wilcoxon}). If there are no repeated data values, a perfect Spearman correlation of $+1$ or $-1$ occurs when each of the variables is a perfect monotone function of the other. The union of the keyphrases is moderately correlated to the authors' keyphrases evaluation gold standard for all keyphrase extraction methods in both datasets. Furthermore, the union of the keyphrases is strongly correlated to the readers' keyphrases evaluation gold standard for all keyphrase extraction methods in both datasets. Finally, the authors' keyphrases is weakly/moderately correlated to the readers' keyphrases evaluation gold standard in both datasets. We should not omit that all strong/moderate correlations are statistically significant with significance level 0.05.

	\begin{table}[H]
		\centering
		\begin{tabu}{|c|c|c|c|c|c|c|c|c|}
			\hline
			\hiderowcolors
			\multirow{3}{*}{} Exact Match & \multicolumn{4}{c|}{Semeval}              & \multicolumn{4}{c|}{NUS}        \\ \cline{1-9} 
			F$_1$@10       & Union            & A        & R  & Inter. & Union            & A & R & Inter.\\ \hline 
			YAKE              & \textbf{0.160} & 0.120          & 0.138  & 0.101    & \textbf{0.205} & 0.147   & 0.186  & 0.132 \\ 
			TfIdf             & \textbf{0.154} & 0.115          & 0.131  & 0.090    & \textbf{0.213} & 0.143   & 0.193   & 0.109 \\ 
			TfIdf2            & \textbf{0.172} & 0.144          & 0.144  & 0.119    & \textbf{0.240} & 0.171   & 0.222   & 0.151 \\ 
			KPM           & \textbf{0.208} & 0.168          & 0.173  & 0.124    & \textbf{0.274} & 0.194   & 0.246   & 0.157 \\ \hline
			RAKE              & \textbf{0.114} & 0.074          & 0.105  & 0.061    & \textbf{0.149} & 0.103   & 0.128 & 0.061 \\ 
			MR                & \textbf{0.146} & 0.099          & 0.126  & 0.091    & \textbf{0.164} & 0.118   & 0.150 & 0.122 \\ 
			TR                & \textbf{0.134} & 0.084          & 0.115  & 0.080    & \textbf{0.143} & 0.098   & 0.134  & 0.102 \\ 
			PR                & \textbf{0.131} & 0.108          & 0.110  & 0.081    & \textbf{0.155} & 0.123   & 0.131  & 0.103 \\ 
			SR                & \textbf{0.036} & 0.017          & 0.034  & 0.016    & \textbf{0.047} & 0.023   & 0.041  & 0.019 \\ \hline
			RVA               & \textbf{0.096} & 0.052          & 0.095  & 0.036    & \textbf{0.100} & 0.064   & 0.095  & 0.043 \\ \hline
		\end{tabu}
		\caption{F$_1$@10 of all methods for the Semeval and NUS datasets using various evaluation gold standards. The scores are calculated according to the exact match evaluation.}
		\label{tbl:union_authors_readers}
	\end{table}

	In this vein, we investigate the correlation (linear or not) between the intersection of the authors' and readers' keyphrases and the authors' keyphrases (Inter.-A) as well as the readers' keyphrases (Inter.-R). Moreover, we investigate the correlation between the intersection and the union of the authors' and the readers' keyphrases (Inter.-Union) (see Tables \ref{tbl:semeval_intersection_authors_readers_spearman_wilcoxon} and \ref{tbl:nguyen_intersection_authors_readers_spearman_wilcoxon}). The intersection of the keyphrases is moderately correlated to the authors' keyphrases evaluation gold standard for all keyphrase extraction methods in both datasets, except for the SR method at the NUS dataset that is weakly correlated. Furthermore, the intersection of the keyphrases is weakly/moderately correlated to the readers' keyphrases evaluation gold standard for all keyphrase extraction methods in both datasets. Finally, the intersection of the keyphrases gives weakly/moderately correlated scores to the union of keyphrases gold standard in both datasets.
	
	In addition, we are interested in investigating whether the F$_1$-scores based on the various evaluation gold standards come from the same distribution. For this reason, we use the Wilcoxon signed-rank test. The scores based on the union of the keyphrases and the scores using the authors' keyphrases evaluation gold standard come from different distributions for all keyphrase extraction methods in both datasets. Furthermore, for almost all methods at the Semeval dataset the scores based on the union of the keyphrases and the scores using the readers' keyphrases evaluation gold standard come from different distributions, while at the NUS dataset the corresponding scores come from either different or same distributions. Additionally, in most methods the scores based on the authors' keyphrases and the scores using the readers' keyphrases evaluation gold standard come from different distributions (see Tables \ref{tbl:semeval_union_authors_readers_spearman_wilcoxon} and \ref{tbl:nguyen_union_authors_readers_spearman_wilcoxon}).
	
	Tables \ref{tbl:semeval_intersection_authors_readers_spearman_wilcoxon} and \ref{tbl:nguyen_intersection_authors_readers_spearman_wilcoxon} show that the scores based on the intersection of the keyphrases and the scores using the authors' keyphrases evaluation gold standard come from different distributions for the most methods in both datasets. Furthermore, the scores based on the intersection of the keyphrases and the scores using the readers' keyphrases evaluation gold standard come from different distributions in both datasets. The same holds for the scores based on the intersection's keyphrases and the scores using the union's keyphrases evaluation gold standard.

	\begin{table}[H]
		\centering
		\begin{tabu}{|c|c|c|c|c|c|c|}
			\hline
			\hiderowcolors
			\multicolumn{7}{|c|}{Semeval}                                                        \\ \hline 
			Exact Match   & \multicolumn{3}{c|}{Spearman}               & \multicolumn{3}{c|}{Wilcoxon}               \\ \hline 
			F$_1$@10   & Union-A & Union-R & A-R & Union-A & Union-R & A-R \\ \hline
			YAKE              & 0.588       & 0.834       & 0.299           & $\approx$0.000       & $\approx$0.000       & 0.031           \\ 
			TfIdf             & 0.641       & 0.844       & 0.357           & $\approx$0.000       & $\approx$0.000       & 0.010           \\ 
			TfIdf2            & 0.652       & 0.832       & 0.330           & $\approx$0.000       & $\approx$0.000       & 0.708           \\ 
			KPM           & 0.614       & 0.802       & 0.278               & $\approx$0.000       & $\approx$0.000       & 0.542           \\ \hline
			RAKE              & 0.570       & 0.898       & 0.380           & $\approx$0.000       & 0.034       & $\approx$0.000           \\ 
			MR                & 0.590       & 0.854       & 0.342           & $\approx$0.000       & $\approx$0.000       & $\approx$0.000           \\ 
			TR                & 0.573       & 0.836       & 0.309           & $\approx$0.000       & 0.001       & $\approx$0.000           \\ 
			PR                & 0.559       & 0.845       & 0.299           & $\approx$0.000       & $\approx$0.000       & 0.760           \\ 
			SR                & 0.491       & 0.862       & 0.356           & $\approx$0.000       & 0.278       & $\approx$0.000           \\ \hline
			RVA               & 0.478       & 0.876       & 0.280           & $\approx$0.000       & 0.015       & $\approx$0.000           \\ \hline
		\end{tabu}
		\caption{Results of Spearman correlation coefficient and Wilcoxon signed-rank test between the scores calculated based on the evaluation using the union of the keyphrases and the authors' keyphrases (Union-A), the union of the keyphrases and the readers' keyphrases (Union-R), as well as the authors' keyphrases and the readers' keyphrases (A-R) for the Semeval dataset.}
		\label{tbl:semeval_union_authors_readers_spearman_wilcoxon}
	\end{table}

	\begin{table}[H]
		\centering
		\begin{tabu}{|c|c|c|c|c|c|c|}
			\hline
			\hiderowcolors
			\multicolumn{7}{|c|}{NUS}                                                        \\ \hline 
			Exact Match   & \multicolumn{3}{c|}{Spearman}               & \multicolumn{3}{c|}{Wilcoxon}               \\ \hline 
			F$_1$@10   & Union-A & Union-R & A-R & Union-A & Union-R & A-R \\ \hline
			YAKE              & 0.671       & 0.846       & 0.379           & $\approx$0.000       & 0.021       & 0.001           \\ 
			TfIdf             & 0.560       & 0.879       & 0.269           & $\approx$0.000       & 0.016       & $\approx$0.000           \\ 
			TfIdf2            & 0.593       & 0.844       & 0.255           & $\approx$0.000       & 0.075       & 0.001           \\ 
			KPM           & 0.523       & 0.864       & 0.200           & $\approx$0.000       & $\approx$0.000       & 0.001           \\ \hline
			RAKE              & 0.622       & 0.813       & 0.238           & $\approx$0.000       & 0.015       & 0.031           \\ 
			MR                & 0.579       & 0.880       & 0.312           & $\approx$0.000       & 0.153       & 0.006           \\ 
			TR                & 0.591       & 0.865       & 0.308           & $\approx$0.000       & 0.842       & 0.001           \\ 
			PR                & 0.633       & 0.801       & 0.296           & $\approx$0.000       & 0.001       & 0.465           \\ 
			SR                & 0.560       & 0.831       & 0.115           & $\approx$0.000       & 0.862       & 0.045           \\ \hline
			RVA               & 0.620       & 0.869       & 0.401           & $\approx$0.000       & 0.301       & $\approx$0.000           \\ \hline
		\end{tabu}
		\caption{Results of Spearman correlation coefficient and Wilcoxon signed-rank test between the scores calculated based on the evaluation using the union of the keyphrases and the authors' keyphrases (Union-A), the union of the keyphrases and the readers' keyphrases (Union-R), as well as the authors' keyphrases and the readers' keyphrases (A-R) for the NUS dataset.}
		\label{tbl:nguyen_union_authors_readers_spearman_wilcoxon}
	\end{table}
	
	\begin{table}[H]
		\centering
		\begin{tabu}{|c|c|c|c|c|c|c|}
			\hiderowcolors
			\hline
			\multicolumn{7}{|c|}{Semeval}                                                   \\ \hline
			Exact Match & \multicolumn{3}{c|}{Spearman}   & \multicolumn{3}{c|}{Wilcoxon}   \\ \hline
			F$_1$@10    & Inter.-A & Inter.-R & Inter.-Union & Inter.-A & Inter.-R & Inter.-Union \\ \hline
			YAKE        & 0.701   & 0.457   & 0.386       & $\approx$0.000   & $\approx$0.000   & $\approx$0.000       \\ 
			
			TfIdf       & 0.619   & 0.513   & 0.439       & $\approx$0.000   & $\approx$0.000   & $\approx$0.000       \\
			TfIdf2      & 0.641   & 0.489   & 0.465       & $\approx$0.000   & $\approx$0.000   & $\approx$0.000       \\
			KPM         & 0.697   & 0.570   & 0.551       & $\approx$0.000   & $\approx$0.000   & $\approx$0.000       \\\hline
			RAKE        & 0.781   & 0.553   & 0.472       & $\approx$0.000   & $\approx$0.000   & $\approx$0.000       \\
			MR          & 0.384   & 0.312   & 0.212       & 0.156   & $\approx$0.000   & $\approx$0.000       \\  
			TR          & 0.341   & 0.221   & 0.127       & 0.364   & $\approx$0.000   & $\approx$0.000       \\
			PR          & 0.442   & 0.349   & 0.194       & 0.001   & $\approx$0.000   & $\approx$0.000       \\
			SR          & 0.553   & 0.312   & 0.231       & 0.949   & 0.002   & 0.001       \\ \hline
			RVA         & 0.357   & 0.310   & 0.269       & 0.003   & $\approx$0.000   & $\approx$0.000      \\\hline
		\end{tabu}
		\caption{Results of Spearman correlation coefficient and Wilcoxon signed-rank test between the scores calculated based on the evaluation using the intersection of the keyphrases and the authors' keyphrases (Inter.-A), the intersection of the keyphrases and the readers' keyphrases (Inter.-R), as well as the intersection and the union of authors' and readers' keyphrases (Inter.-Union) for the Semeval dataset.}
		\label{tbl:semeval_intersection_authors_readers_spearman_wilcoxon}
	\end{table}

	\begin{table}[H]
		\centering
		\begin{tabu}{|c|c|c|c|c|c|c|}
			\hline
			\hiderowcolors
			\multicolumn{7}{|c|}{NUS}                                                    \\ \hline
			Exact Match & \multicolumn{3}{c|}{Spearman}   & \multicolumn{3}{c|}{Wilcoxon}   \\ \hline
			F$_1$@10       & Inter.-A & Inter.-R & Inter.-Union & Inter.-A & Inter.-R & Inter.-Union \\ \hline
			YAKE        & 0.685   & 0.477   & 0.383       & 0.040   & $\approx$0.000   & $\approx$0.000       \\ 
			
			TfIdf       & 0.607   & 0.383   & 0.370       & $\approx$0.000   & $\approx$0.000   & $\approx$0.000       \\ 
			TfIdf2      & 0.612   & 0.440   & 0.424       & 0.002   & $\approx$0.000   & $\approx$0.000       \\ 
			KPM         & 0.610   & 0.327   & 0.331       & $\approx$0.000   & $\approx$0.000   & $\approx$0.000       \\ \hline
			RAKE        & 0.650   & 0.411   & 0.350       & $\approx$0.000   & $\approx$0.000   & $\approx$0.000       \\ 
			MR          & 0.508   & 0.243   & 0.248       & 0.908   & 0.001   & $\approx$0.000       \\ 
			TR          & 0.367   & 0.192   & 0.197       & 0.856   & 0.001   & $\approx$0.000       \\ 
			PR          & 0.498   & 0.361   & 0.267       & 0.191   & $\approx$0.000   & $\approx$0.000       \\ 
			SR          & 0.214   & 0.023   & 0.009       & 0.567   & 0.002   & $\approx$0.000       \\ \hline
			RVA         & 0.332   & 0.236   & 0.169       & 0.019   & $\approx$0.000   & $\approx$0.000       \\ \hline
		\end{tabu}
		\caption{Results of Spearman correlation coefficient and Wilcoxon signed-rank test between the scores calculated based on the evaluation using the intersection of the keyphrases and the authors' keyphrases (Inter.-A), the intersection of the keyphrases and the readers' keyphrases (Inter.-R), as well as the intersection and the union of authors' and readers' keyphrases (Inter.-Union) for the NUS dataset.}
		\label{tbl:nguyen_intersection_authors_readers_spearman_wilcoxon}
	\end{table}

	To sum up, it is strongly recommended to the research community to explicitly mention which set of keyphrases is regarded as a gold evaluation standard in the experiments, since different evaluation gold standards give different performance ranking and score estimations. Particularly, the union of the keyphrases evaluation gold standard is strongly correlated to the readers' evaluation gold standard for all keyphrase extraction methods in both datasets. Moreover, the average F$_1$@10 scores with respect to readers' evaluation gold standard seem to be closer to the scores with respect to the union's evaluation gold standard, even though in most cases at the Semeval dataset and in half cases of the NUS dataset the union's and the readers' evaluation gold standards give scores coming from different distributions. Using both types of annotations, i.e., authors' keyphrases and multiple annotators' (readers') keyphrases, we have at our disposal a quite expanded set of keyphrases. However, using only the multiple annotators' keyphrases, we end up to a decent number of gold unbiased keyphrases certainly depending  on the methodology followed for the annotation process and the collective effort. On the other hand, authors' sets of keyphrases contain fewer but sufficient phrases to cover the topics of the target document. Such type of evaluation standard usually gives lower performance scores, moderately correlated with those resulting from the union keyphrases' gold evaluation standard. Finally, the intersection's  evaluation gold standard is not recommended, as it is quite strict, contains a very small number of keyphrases, and does not guarantee that all topics discussed in the target documents are covered by the intersection of the keyphrases, since authors and readers may use different expressive means/vocabularies and do not share same annotation motivations.

	\subsection{Qualitative Analysis}
	\label{qualitative-results}
	
	Via this qualitative study, we would like to highlight the limitations of the exact match evaluation, especially in cases where we are interested in the actual performance (success rate) of a method. Additionally, we show a case where the ``looser'' partial match strategy gives lower scores than the exact match one. Finally, we give an example that the partial match evaluation can be harmful, failing to assess properly the syntactic correctness of the return keyphrases. We use MR to extract the keyphrases from 3 publication full-texts of the Krapivin dataset collection.
	
	First, we present an example where the $F_1$-score with respect to the partial evaluation is greater than the corresponding exact match score and closer to the corresponding score with respect to the manual evaluation, which intuitively is the expected case.
	We quote the publication's title and abstract below in order to get a sense of its content:
	\\\\
	{\centering\fbox{\begin{minipage}{45em} \small{
					Title: Exact algorithms for finding minimum transversals in rank-3 hypergraphs.
					
					Abstract: We present two algorithms for the problem of finding a minimum transversal in a hypergraph of rank 3, also known as the 3-Hitting Set problem. This problem is a natural extension of the vertex cover problem for ordinary graphs. The first algorithm runs in time O(1.6538n) for a hypergraph with n vertices, and needs polynomial space. The second algorithm uses exponential space and runs in time O(1.6316n).
				}
	\end{minipage}}}
	\\\\
	
	The corresponding set of the ``gold'' keyphrases are: 
	\{\textit{hypergraph, 3-hitting set, exact algorithm, minimum transversal}\}. For evaluation purposes, we transform the set of ``gold'' keyphrases into the following one (after stemming and removal of punctuation marks, such as dashes and hyphens):
	
	$\begin{aligned}
	\{(hypergraph), (3hit, set), (exact, algorithm), (minimum, transvers)\}
	\end{aligned}$
	
	The MR's result set is given in the first box below, followed by its stemmed version in the second box. The words that are both in the golden set and in the set of our candidates are highlighted with bold typeface:\\
	
	{\centering\fbox{\begin{minipage}{45em} { \{problem, case, branching, 3hitting set problem, time, edges, minimum transversals, algorithms, number, rank3 hypergraphs\}
				}
	\end{minipage}}}
	
	{\centering\fbox{\begin{minipage}{45em} { \{(problem), (case), (branch), (\textbf{3hit}, \textbf{set}, problem), (time), (edg), (\textbf{minimum}, \textbf{transvers}), (\textbf{algorithm}), (number), (rank3, \textbf{hypergraph})]
					\}
				}
	\end{minipage}}}
	\\\\
	
	According to the exact match evaluation approach, the top-10 returned candidate keyphrases by MR include 1 True Positive (TP), the bigram phrase \textit{minimum transversals}, 9 False Positives (FPs) and 3 False Negatives (FNs). Hence,
	
	\begin{table}[H]
		\centering
		
		\begin{tabular}{ccc}
			\hiderowcolors
			$\textit{precision} = \frac{\textit{TPs}}{\textit{TPs}+\textit{FPs}} = 0.10$ & $\textit{recall} = \frac{\textit{TPs}}{\textit{TPs}+\textit{FNs}} = 0.25$ &   $F_1 = 2\times\frac{\textit{precision} \times \textit{recall}}{\textit{precision} + \textit{recall}} = 0.14$
		\end{tabular}
	\end{table}
	
	
	However, the stemmed set of words found in all golden keyphrases used by the partial match evaluation approach is the following:
	
	$\begin{aligned}
	\{(hypergraph), (3hit), (set), (exact), (algorithm), (minimum), (transvers)\}
	\end{aligned}$
	
	and the set of words found in all extracted keyphrases by MR is given below:
	\\\\
	{\centering\fbox{\begin{minipage}{45em} { \{(problem), (case), (branch), (\textbf{3hit}), (\textbf{set}), (time), (edg), (\textbf{minimum}), (\textbf{transvers}), (\textbf{algorithm}), (number), (rank3), (\textbf{hypergraph})]
					\}
				}
	\end{minipage}}}
	\\\\
	According to the partial match evaluation approach, the top-10 returned candidate keyphrases by MR include 6 TPs, 7 FPs and 1 FNs. Consequently,

	\begin{table}[H]
		\centering
		\begin{tabular}{ccc}
			\hiderowcolors
			$\textit{precision} = 0.46$ & $\textit{recall} = 0.86$ &   $F_1 = 0.60$
		\end{tabular}
	\end{table}
	
	
	In this case it is confirmed that the exact match evaluation seems to be too strict  whereas the looser partial match approach is closer to the actual performance of the MR method. However, partial match evaluation considers as TPs unigram phrases that have trivial meaning without their context words, e.g., the unigram \textit{algorithm} instead of the bigram \textit{exact algorithm}.  
	
	We give an example where the exact match evaluation corresponds to a higher $F_1$-score than the partial match evaluation. The results are based on the output of the MR method. We quote the publication's title and abstract below in order to get a sense of its content:
	\\\\
	{\centering\fbox{\begin{minipage}{45em} \small{
					Title: Finite state machines for strings over infinite alphabets.
					
					Abstract:Motivated by formal models recently proposed in the context of XML, we study automata and logics on strings over infinite alphabets. These are conservative extensions of classical automata and logics defining the regular languages on finite alphabets. Specifically, we consider register and pebble automata, and extensions of first-order logic and monadic second-order logic. For each type of automaton we consider one-way and two-way variants, as well as deterministic, nondeterministic, and alternating control. We investigate the expressiveness and complexity of the automata and their connection to the logics, as well as standard decision problems. Some of our results answer open questions of Kaminski and Francez on register automata.
					
				}
	\end{minipage}}}
	\\\\
	The corresponding set of the ``gold'' keyphrases are: 
	\{\textit{xml, automata, first-order logic, expressiveness, pebbles, monadic second-order logic, infinite alphabets, registers}\}. For evaluation purposes, we transform the set of ``gold'' keyphrases into the following one (after stemming and removal of punctuation marks, such as dashes and hyphens):
	
	$\begin{aligned}
	\{(xml), (automata), (firstord, logic), (express), (pebbl)\\ (monad, secondord, logic),(infinit, alphabet), (regist)\}
	\end{aligned}$
	
	The MR's result set is given in the first box below, followed by its stemmed version in the second box. The words that are both in the golden set and in the set of our candidates are highlighted with bold typeface:\\
	
	{\centering\fbox{\begin{minipage}{45em} { \{pebbles, mso, finite set, set, proof, strings, pebble automata, automata, positions, data values\}
				}
	\end{minipage}}}
	
	{\centering\fbox{\begin{minipage}{45em} { \{(\textbf{pebbl}), (mso), (finit, set), (set), (proof), (string), (\textbf{pebbl}, \textbf{automata}), (\textbf{automata}), (posit), (data, valu)\}
				}
	\end{minipage}}}
	\\\\
	
	According to the exact match evaluation, the top-10 returned candidate keyphrases by MR include 2 True Positives (TPs), the unigram phrases \textit{pebbles} and \textit{automata}, 8 False Positives (FPs) and 6 False Negatives (FNs). Hence,\\
	
	\begin{table}[H]
		\centering
		\begin{tabular}{ccc}
			$\textit{precision} = 0.20$ & $\textit{recall} = 0.25$ &   $F_1 = 0.22$
		\end{tabular}
	\end{table}
	
	
	However, partial match evaluation approach uses the stemmed set of words found in all golden keyphrases, i.e.:
	
	$\begin{aligned}
	\{(xml), (automata), (firstord), (logic), (express), (pebbl), (monad), (secondord), (infinit), (alphabet), (regist)\}
	\end{aligned}$
	
	and the set of words found in all extracted keyphrases by MR, i.e.:
	\\\\
	{\centering\fbox{\begin{minipage}{45em} { \{(\textbf{pebbl}), (mso), (finit), (set), (proof), (string), (\textbf{automata}), (posit), (data), (valu)\}
				}
	\end{minipage}}}
	\\\\
	According to the partial match evaluation, the top-10 returned candidate keyphrases by MR include 2 TPs, \textit{pebbles} and \textit{automata}, 8 FPs and 9 FNs. Hence,
	\begin{table}[H]
		\centering
		\begin{tabular}{ccc}
			\hiderowcolors
			$\textit{precision} = 0.20$ & $\textit{recall} = 0.18$ &   $F_1 = 0.19$
		\end{tabular}
	\end{table}
	
	
	So, it is quite usual in cases where there are unigram keyphrases as TPs and multiword keyphrases as FPs/FNs, the exact match $F_1$-score to be higher than the partial match $F_1$-score. Note that both evaluation approaches fail to recognize as TP the unigram keyphrase \textit{mso} which is an abbreviation of the multiword keyphrase \textit{monadic second-order} logic. Finally, in Appendix we give \ref{quality} an example where the partial match evaluation can be considered as harmful compared to the strict exact match evaluation.

\section{Conclusions and Future Directions}
\label{sec:conclusions}
	
Keyphrases are multi-purpose knowledge gems. They constitute a concise summary of documents that is extremely useful both for human inspection and machine consumption, in support of tasks such as faceted search, document classification and clustering, query expansion and document recommendation. Our article reviews the existing body of work on keyphrase extraction and presents a comprehensive organization of the material that aims to help newcomers and veterans alike navigate the large amount of prior art and grasp its evolution.  

We present a large number of both unsupervised and supervised keyphrase extraction methods, including recent deep learning methods, categorizing them according to their main features and properties, and highlighting their strengths and weaknesses. We discuss the challenges that supervised methods face, namely the subjectivity that characterizes the existing annotated datasets and the imbalance of keyphrases versus non-keyphrases. In addition, we discuss how keyphrase extraction methods are currently evaluated, and present a long list of free and commercial keyphrase extraction software and APIs, as well as the main collections of documents with associated keyphrases that are used for obtaining experimental results. 

Our review includes an extensive empirical evaluation study of keyphrase extraction. We compare  commercial APIs, as well as unsupervised methods ourselves, while for supervised methods we include a table with results collected from the corresponding papers. The results show that simple unsupervised methods, such as TfIdf2, are strong baselines that should be considered in empirical studies and that deep learning methods achieve state-of-the-art results. Among unsupervised methods, we notice that graph-based methods work better for short, while statistical methods work better for long documents. 

Our evaluation study presents a thorough analysis of the exact and partial matching approaches, concluding with the recommendation of considering their average, and highlighting the need for approaches that take the semantic similarity of predicted and golden keyphrases. In addition, our study investigates how the different golden keyphrase sources (authors and readers) affect the evaluation of keyphrase extraction methods, concluding that they play a significant role and should be explicitly considered and reported in empirical studies.

A lot of progress still remains to be done in this challenging task, as the accuracy of state-of-the-art systems has not reached satisfactory levels yet. At the moment, the most exciting developments in mastering language are coming from the frontier of deep learning and unsupervised language models \citep{DBLP:conf/naacl/DevlinCLT19, DBLP:journals/corr/abs-1906-08237, DBLP:journals/corr/abs-1803-11175}. The exploitation of such models for keyphrase extraction and/or generation appears as the most interesting future direction.

	\section*{Funding Information}
	
	This work was partially funded by Atypon Systems,  LLC\footnote{\url{https://www.atypon.com/}}.

\clearpage
	\appendix
	
	\section{Qualitative Results}
	\label{quality}
	We give an example where the partial match evaluation can be considered as harmful compared to the strict exact match evaluation. The results are based on the output of the MR method. Once again, we quote the publication's title and abstract below in order to get a sense of its content:
	\\\\
	{\centering\fbox{\begin{minipage}{45em} \small{
					Title: Programming and Verifying Real-Time Systems by Means of the Synchronous Data-Flow Language LUSTRE.
					
					Abstract: The benefits of using a synchronous data-flow language for programming critical real-time systems are investigated. These benefits concern ergonomy (since the dataflow approach meets traditional description tools used in this domain) and ability to support formal design and verification methods. It is shown, using a simple example, how the language LUSTRE and its associated verification tool LESAR, can be used to design a program, to specify its critical properties, and to verify these properties. As the language LUSTRE and its uses have already been discussed in several papers, emphasis is put on program verification.

				}
	\end{minipage}}}
	\\\\
	The corresponding set of the ``gold'' keyphrases are: 
	\{\textit{sampling, matrix algorithms, low-rank approximation}\}. For evaluation purposes, we transform the set of ``gold'' keyphrases into the following one (after stemming and removal of punctuation marks, such as dashes and hyphens):
	
	$\begin{aligned}
	\{(sampl), (matrix, algorithm), (lowrank, approxim)\}
	\end{aligned}$
	
	The MR's result set is given in the first box below, followed by its stemmed version in the second box. The words that are both in the golden set and in the set of our candidates are highlighted with bold typeface:\\
	
	{\centering\fbox{\begin{minipage}{45em} { \{low-rank approximations, algorithm, singular vectors, rows, lemma, assumption a1, entries, problem, modern applications, matrix\}
				}
	\end{minipage}}}
	
	{\centering\fbox{\begin{minipage}{45em} { \{(\textbf{lowrank}, \textbf{approxim}), (algorithm), (singular, vectort), (row), (lemma), (assumpt, a1), (entri), (problem), (modern, applic), (matrix)\}
				}
	\end{minipage}}}
	\\\\
	
	According to the exact match evaluation, the top-10 returned candidate keyphrases by MR include 1 True Positive (TP), the bigram phrase \textit{low-rank approximations}, 9 False Positives (FPs) and 2 False Negatives (FNs). We should notice that this example belongs to the less usual case where the exact match evaluation is close to the manual evaluation, i.e., in a right direction, indicating the actual success rate of the method. Hence, $\textit{precision} = 0.10, \textit{recall} = 0.33, F_1 = 0.15$
	
	
	
	However, partial match evaluation approach uses the stemmed set of words found in all golden keyphrases, i.e.:
	
	$\begin{aligned}
	\{(sampl), (matrix), (algorithm), (lowrank), (approxim)\}
	\end{aligned}$
	
	and the set of words found in all extracted keyphrases by MR, i.e.:
	\\\\
	{\centering\fbox{\begin{minipage}{45em} { \{(\textbf{lowrank}), (\textbf{approxim}), (\textbf{algorithm}), (singular), (vector), (row), (lemma), (assumpt), (a1), (entri), (problem), (modern), (appli), (\textbf{matrix})\}
				}
	\end{minipage}}}
	\\\\
	
	According to the partial match evaluation, the top-10 returned candidate keyphrases by MR include 4 TPs, \textit{lowrank}, \textit{approxim}, \textit{algorithm} and \textit{matrix}, 10 FPs and 1 FNs. Hence, $\textit{precision} = 0.29, \textit{recall} = 0.80, F_1 = 0.42$
	
	Despite the fact that the partial match evaluation gives $F_1$-scores closer to the manual ones, we see here that the $F_1$-score based on the partial match evaluation is quite far from the actual performance of the method on the specific document. Particularly, the partial match strategy fails to evaluate the syntactic correctness of the returned phrases \textit{algorithm} and \textit{matrix}, which have very general meaning when they returned as unigrams by MR compared to the bigram gold keyphrase \textit{matrix algorithm}.

	\section{Commercial APIs Evaluation}
	\label{APIs_map_partial_eval}
	
	
	\begin{table}[H]
		\centering
		\scalebox{1.0}{
			\begin{tabu}{|c|c|c|c|c|c|c|c|c|c|c|}
				\hline
				\hiderowcolors
				\multirow{2}{*}{} & \multicolumn{2}{c|}{Semeval}    & \multicolumn{2}{c|}{NUS}     & \multicolumn{2}{c|}{Krapivin}   & \multicolumn{2}{c|}{Inspec}      & \multicolumn{2}{c|}{500N-KPCrowd}     \\ \cline{2-11} 
				MAP   & @10         & @20         & @10         & @20         & @10         & @20         & @10         & @20         & @10         & @20         \\ \hline
				IBM               & 0.066          & 0.041          & 0.071          & 0.042          & 0.047          & 0.028          & \textbf{0.146}          & \textbf{0.093}          & 0.095          & 0.078          \\ 
				GOOGLE            & 0.060          & 0.038          & 0.088          & 0.056          & 0.042          & 0.025          & 0.090          & 0.060          & \textbf{0.258}          & \textbf{0.209}          \\ 
				Amazon            & 0.017          & 0.013          & 0.011          & 0.010          & 0.008          & 0.007          & 0.018          & 0.016          & 0.059          & 0.047          \\ 
				Textrazor         & \textbf{0.109}          & \textbf{0.062}          & \textbf{0.113}          & \textbf{0.069}          & \textbf{0.062}          & \textbf{0.035}          & 0.084          & 0.051          & 0.202          & 0.143          \\ 
				Aylien            & 0.061          & 0.034          & 0.063          & 0.036          & 0.027          & 0.015          & 0.054          & 0.034          & 0.170          & 0.165          \\ 
				\hline
		\end{tabu}}
		\caption{MAP@10 and MAP@20 according to the exact match evaluation approach of all APIs for all datasets.}
		\label{tbl:apis_map_results}
	\end{table}
	
	
	\begin{table}[H]
		\centering
		\scalebox{1.0}{
			\begin{tabu}{|c|c|c|c|c|c|c|c|c|c|c|}
				\hline
				\hiderowcolors
				\multirow{2}{*}{} & \multicolumn{2}{c|}{Semeval}    & \multicolumn{2}{c|}{NUS}     & \multicolumn{2}{c|}{Krapivin}   & \multicolumn{2}{c|}{Inspec}      & \multicolumn{2}{c|}{500N-KPCrowd}     \\ \cline{2-11} 
				pF$_1$  & @10         & @20         & @10         & @20         & @10         & @20         & @10         & @20         & @10         & @20         \\ \hline
				IBM               & \textbf{0.282}          & \textbf{0.307}          & \textbf{0.375}          & 0.344          & 0.348          & 0.305          & \textbf{0.569} & \textbf{0.583}          & \textbf{0.348} & \textbf{0.445}\\ 
				GOOGLE            & 0.233          & 0.253          & \textbf{0.375}          & \textbf{0.374}          & \textbf{0.368}          & \textbf{0.336}          & 0.444          & 0.501          & 0.267          & 0.362          \\ 
				Amazon            & 0.190          & 0.268          & 0.188          & 0.267          & 0.180          & 0.227          & 0.261          & 0.388          & 0.239          & 0.356          \\ 
				Textrazor         & 0.212          & 0.258          & 0.278          & 0.316          & 0.289          & 0.302          & 0.348          & 0.411          & 0.167          & 0.240          \\ 
				Aylien            & 0.278          & 0.305          & 0.339          & 0.341          & 0.296          & 0.295          & 0.401          & 0.480          & 0.237          & 0.336          \\ 
				\hline
		\end{tabu}}
		\caption{F$_1$@10 and F$_1$@20 according to the partial match evaluation approach (pF$_1$) of all APIs for all datasets.}
		\label{tbl:apis_partial_match_results}
	\end{table}

	\section{Bibliographic Sources}
	\label{sources}

	\begin{table}[H]
		\centering
		\scalebox{0.88}{
		\begin{tabu}{|c|c|}
			\hiderowcolors
			\hline
			\textbf{Journal}                               & \textbf{Publisher} \\ \hline
			ACM Transactions on Computer-Human Interaction & ACM                \\ 
			Information Systems                            & ELSEVIER           \\ 
			Decision Support Systems                       & ELSEVIER           \\ 
			Information Processing \& Management           & ELSEVIER           \\ 
			Language Resources and Evaluation              & Springer           \\ 
			Data Science and Engineering                   & Springer           \\ 
			Text Mining: Applications and Theory           & Wiley              \\ \hline
%
%
				\textbf{Conference (Abbreviation)}                                                                                                  & \textbf{Proceedings Publisher} \\ \hline
				Annual Meeting of the Association for Computational Linguistics (ACL)                                                          & ACL                            \\ 
				Conference on Empirical Methods in Natural Language Processing (EMNLP)                                                         & ACL                            \\ 
				International Conference on Computational Linguistics (COLING)                                                                 & ACL                            \\ 
				Conference of the North American Chapter of the Association for Computational Linguistics (NAACL) & ACL                            \\ 
				International Joint Conference on Natural Language Processing (IJCNLP)                                                         & ACL                            \\ 
				European Conference on Information Retrieval (ECIR)                                                                            & Springer                       \\ 
				ACM International Conference on Information and Knowledge Management (CIKM)                                                    & ACM                            \\ 
				International World Wide Web Conference (WWW)                                                                                  & ACM                            \\ 
				ACM International Conference on Research and Development in Information Retrieval (SIGIR)                                      & ACM                            \\ 
				International Conference on Language Resources and Evaluation (LREC)                                                           & ELRA                              \\ 
				Annual Conference of the International Speech Communication Association (Interspeech)                                          & ISCA                              \\ 
				IEEE International Conference on Data Mining (ICDM)                                                                            & IEEE                           \\ 
				AAAI Conference on Artificial Intelligence (AAAI)                                                                              & AAAI                           \\ \hline
		\end{tabu}}
		\caption{The main journals and conferences that were used as sources of related articles in this survey.}
		\label{tbl:journals_conferences}
	\end{table}
	
\section{Automatic vs Manual Scores}
\label{sec:differences}


\begin{figure}[H]
	\centering
	\begin{subfigure}[b]{0.9\textwidth}
		\centering
		\includegraphics[width=\linewidth]{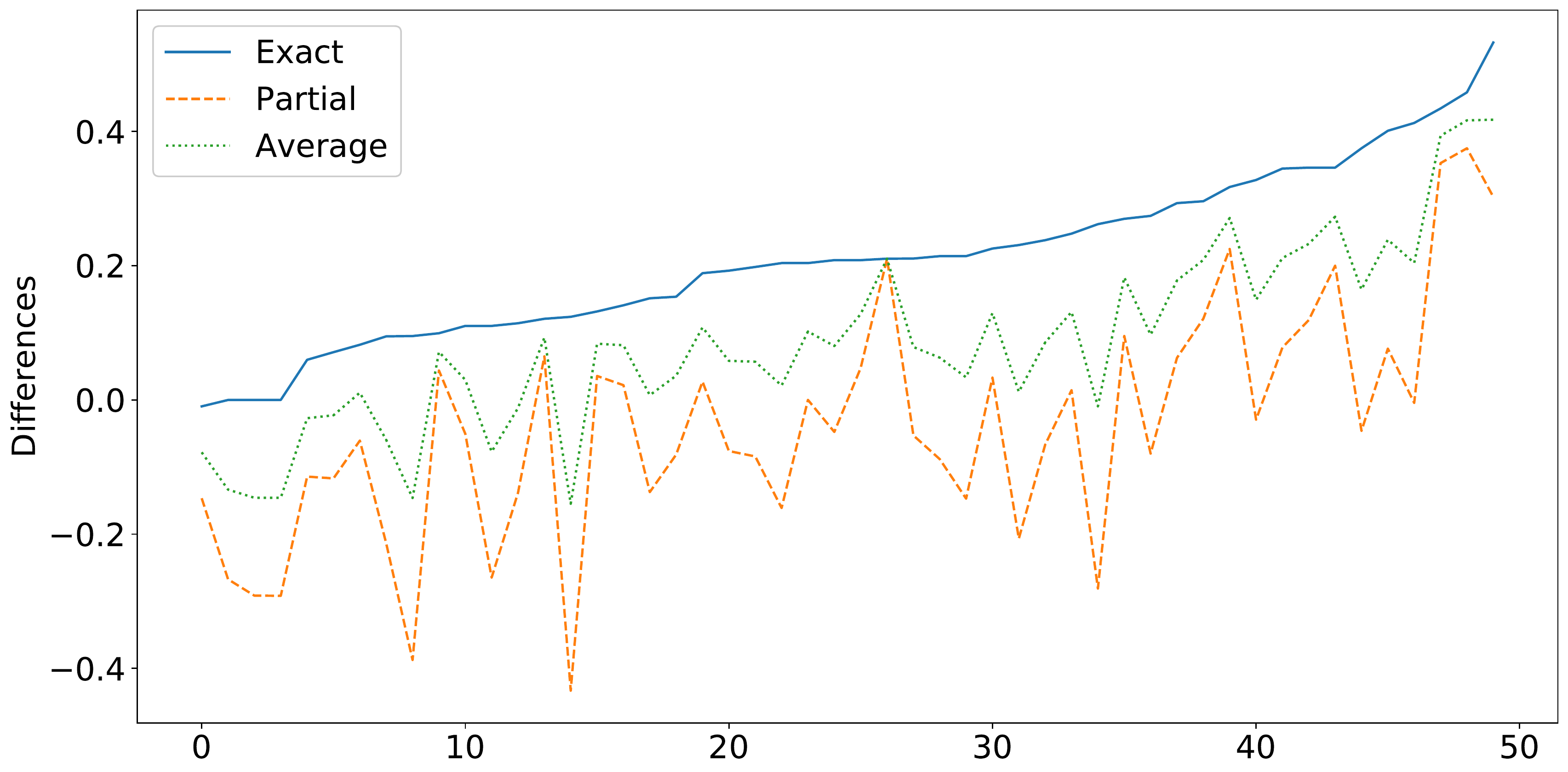}
		\caption{KPM}
		\label{subfig:kpm_plot_diff}
	\end{subfigure}%
	\vspace{\floatsep}
	
	\begin{subfigure}[b]{0.9\textwidth}
		\centering
		\includegraphics[width=\linewidth]{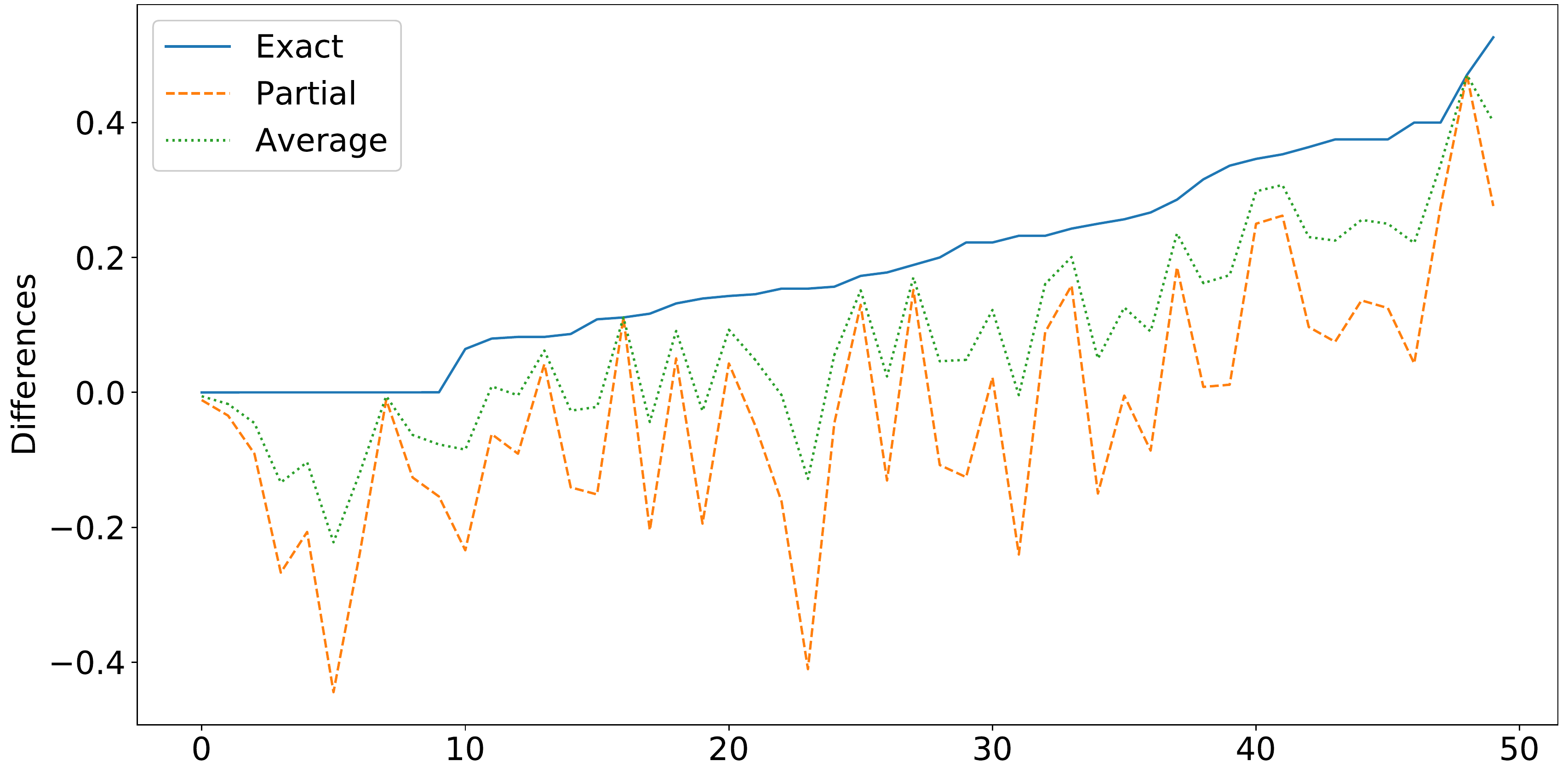}
		\caption{MR}
		\label{subfig:mr_plot_diff}
	\end{subfigure}
	\caption{Differences between the F$_1$-scores based on the manual evaluation and the F$_1$-scores based on the exact (solid line), partial (dashed line) and average (dotted line) evaluation approaches for the 50 manually evaluated documents given on the x axis.}
	\label{fig:differences2}
\end{figure}

In Fig. \ref{fig:differences2} we plot the differences between the F$_1$-score values based on the manual evaluation and the F$_1$-score values calculated according to the exact (solid line), partial (dashed line) and average (dotted line) evaluation approaches for KPM and MR methods. The differences between manual and average scores range closer to 0 in comparison with the differences between manual-exact and manual-partial scores, respectively. Particularly, most differences between manual-exact scores are positive (manual scores are greater than the exact match scores), whereas there are many differences between manual-partial scores that are negative (manual scores are lower than the partial match scores). 



\begin{thebibliography}{111}
		\expandafter\ifx\csname natexlab\endcsname\relax\def\natexlab#1{#1}\fi
		\expandafter\ifx\csname url\endcsname\relax
		\def\url#1{\texttt{#1}}\fi
		\expandafter\ifx\csname urlprefix\endcsname\relax\def\urlprefix{URL: }\fi
		
		\bibitem[{Agarwal and Lavie(2008)}]{agarwal2008meteor}
		Agarwal, A. and Lavie, A. (2008) Meteor, {M-BLEU} and {M-TER:} evaluation
		metrics for high-correlation with human rankings of machine translation
		output.
		\newblock In \textit{Proceedings of the Third Workshop on Statistical Machine
			Translation, WMT@ACL 2008, Columbus, Ohio, USA, June 19, 2008}, 115--118.
		\newblock \urlprefix\url{https://aclanthology.info/papers/W08-0312/w08-0312}.
		
		\bibitem[{Alzaidy et~al.(2019)Alzaidy, Caragea and
			Giles}]{DBLP:conf/www/AlzaidyCG19}
		Alzaidy, R., Caragea, C. and Giles, C.~L. (2019) Bi-lstm-crf sequence labeling
		for keyphrase extraction from scholarly documents.
		\newblock In \textit{The World Wide Web Conference, {WWW} 2019, San Francisco,
			CA, USA, May 13-17, 2019}, 2551--2557.
		\newblock \urlprefix\url{https://doi.org/10.1145/3308558.3313642}.
		
		\bibitem[{Augenstein et~al.(2017)Augenstein, Das, Riedel, Vikraman and
			McCallum}]{augenstein2017semeval}
		Augenstein, I., Das, M., Riedel, S., Vikraman, L. and McCallum, A. (2017)
		Semeval 2017 task 10: Scienceie - extracting keyphrases and relations from
		scientific publications.
		\newblock In \textit{Proceedings of the 11th International Workshop on Semantic
			Evaluation, SemEval@ACL 2017, Vancouver, Canada, August 3-4, 2017}, 546--555.
		\newblock \urlprefix\url{https://doi.org/10.18653/v1/S17-2091}.
		
		\bibitem[{Basaldella et~al.(2018)Basaldella, Antolli, Serra and
			Tasso}]{DBLP:conf/ircdl/BasaldellaAST18}
		Basaldella, M., Antolli, E., Serra, G. and Tasso, C. (2018) Bidirectional
		{LSTM} recurrent neural network for keyphrase extraction.
		\newblock In \textit{Proceedings of the 14th Italian Research Conference on
			Digital Libraries, {IRCDL} 2018, Udine, Italy, January 25-26, 2018},
		180--187.
		\newblock \urlprefix\url{https://doi.org/10.1007/978-3-319-73165-0_18}.
		
		\bibitem[{Bekoulis et~al.(2018{\natexlab{a}})Bekoulis, Deleu, Demeester and
			Develder}]{bekoulis:18b}
		Bekoulis, G., Deleu, J., Demeester, T. and Develder, C. (2018{\natexlab{a}})
		Adversarial training for multi-context joint entity and relation extraction.
		\newblock In \textit{Proceedings of the 2018 Conference on Empirical Methods in
			Natural Language Processing}, 2830--2836.
		
		\bibitem[{Bekoulis et~al.(2018{\natexlab{b}})Bekoulis, Deleu, Demeester and
			Develder}]{bekoulis:18a}
		Bekoulis, G., Deleu, J., Demeester, T. and Develder, C. (2018{\natexlab{b}}) Joint entity recognition and relation extraction as a
		multi-head selection problem.
		\newblock \textit{Expert Systems with Applications}, \textbf{114}, 34--45.
		
		\bibitem[{Bengtson and Roth(2008)}]{bengtson2008understanding}
		Bengtson, E. and Roth, D. (2008) Understanding the value of features for
		coreference resolution.
		\newblock In \textit{Proceedings of the Conference on Empirical Methods in
			Natural Language Processing, {EMNLP} 2008, 25-27 October 2008, Honolulu,
			Hawaii, USA, {A} meeting of SIGDAT, a Special Interest Group of the {ACL}},
		294--303.
		\newblock \urlprefix\url{http://www.aclweb.org/anthology/D08-1031}.
		
		\bibitem[{Bennani-Smires et~al.(2018)Bennani-Smires, Musat, Hossmann, Baeriswyl
			and Jaggi}]{DBLP:journals/corr/abs-1801-04470}
		Bennani-Smires, K., Musat, C., Hossmann, A., Baeriswyl, M. and Jaggi, M. (2018)
		Simple unsupervised keyphrase extraction using sentence embeddings.
		\newblock In \textit{Proceedings of the 22nd Conference on Computational
			Natural Language Learning, Brussels, Belgium, October, 2018}, 221--229.
		Brussels, Belgium: Association for Computational Linguistics.
		\newblock \urlprefix\url{https://www.aclweb.org/anthology/K18-1022}.
		
		
		
		
		\bibitem[{Blei et~al.(2003)Blei, Ng and Jordan}]{blei2003latent}
		Blei, D.~M., Ng, A.~Y. and Jordan, M.~I. (2003) Latent dirichlet allocation.
		\newblock \textit{Journal of Machine Learning Research}, \textbf{3}, 993--1022.
		\newblock \urlprefix\url{http://www.jmlr.org/papers/v3/blei03a.html}.
		
		
		\bibitem[{Bojanowski et~al.(2017)Bojanowski, Grave, Joulin and
			Mikolov}]{DBLP:journals/tacl/BojanowskiGJM17}
		Bojanowski, P., Grave, E., Joulin, A. and Mikolov, T. (2017) Enriching word
		vectors with subword information.
		\newblock \textit{{TACL}}, \textbf{5}, 135--146.
		\newblock
		\urlprefix\url{https://transacl.org/ojs/index.php/tacl/article/view/999}.
		
		\bibitem[{Boudin(2016)}]{DBLP:conf/coling/Boudin16}
		Boudin, F. (2016) Pke: an open source python-based keyphrase extraction
		toolkit.
		\newblock In \textit{{COLING} 2016, 26th International Conference on
			Computational Linguistics, Proceedings of the Conference System
			Demonstrations, Osaka, Japan, December 11-16, 2016}, 69--73.
		\newblock \urlprefix\url{http://aclweb.org/anthology/C/C16/C16-2015.pdf}.
		
		\bibitem[{Boudin et~al.(2016)Boudin, Mougard and
			Cram}]{DBLP:conf/aclnut/BoudinMC16}
		Boudin, F., Mougard, H. and Cram, D. (2016) How document pre-processing affects
		keyphrase extraction performance.
		\newblock In \textit{Proceedings of the 2nd Workshop on Noisy User-generated
			Text, NUT@COLING 2016, Osaka, Japan, December 11, 2016}, 121--128.
		\newblock \urlprefix\url{https://aclanthology.info/papers/W16-3917/w16-3917}.
		
		\bibitem[{Boudin(2018)}]{Boudin18Multipartite}
		Boudin, F. (2018) Unsupervised keyphrase extraction with multipartite graphs.
		\newblock In \textit{Proceedings of the 2018 Conference of the North American
			Chapter of the Association for Computational Linguistics: Human Language
			Technologies, NAACL-HLT, New Orleans, Louisiana, USA, June 1-6, 2018, Volume
			2 (Short Papers)}, 667--672.
		\newblock \urlprefix\url{https://aclanthology.info/papers/N18-2105/n18-2105}.
		
		\bibitem[{Bougouin et~al.(2013)Bougouin, Boudin and
			Daille}]{bougouin2013topicrank}
		Bougouin, A., Boudin, F. and Daille, B. (2013) Topic{R}ank: Graph-based topic
		ranking for keyphrase extraction.
		\newblock In \textit{Proceedings of the 6th International Joint Conference on
			Natural Language Processing, {IJCNLP} 2013, Nagoya, Japan, October 14-18,
			2013}, 543--551.
		\newblock \urlprefix\url{http://aclweb.org/anthology/I/I13/I13-1062.pdf}.
		
		\bibitem[{Bougouin et~al.(2016)Bougouin, Boudin and
			Daille}]{GraphCoR2016Bougouin}
		Bougouin, A., Boudin, F. and Daille, B. (2016) Keyphrase annotation with graph co-ranking.
		\newblock In \textit{{COLING} 2016, 26th International Conference on
			Computational Linguistics, Proceedings of the Conference: Technical Papers,
			December 11-16, 2016, Osaka, Japan}, 2945--2955.
		\newblock \urlprefix\url{http://aclweb.org/anthology/C/C16/C16-1277.pdf}.
		
		\bibitem[{Brin and Page(1998)}]{grin+page1998}
		Brin, S. and Page, L. (1998) The anatomy of a large-scale hypertextual web
		search engine.
		\newblock \textit{Computer Networks}, \textbf{30}, 107--117.
		\newblock \urlprefix\url{https://doi.org/10.1016/S0169-7552(98)00110-X}.
		
		\bibitem[{Buckley and Voorhees(2004)}]{buckley2004retrieval}
		Buckley, C. and Voorhees, E.~M. (2004) Retrieval evaluation with incomplete
		information.
		\newblock In \textit{Proceedings of the 27th Annual International {ACM} {SIGIR}
			Conference on Research and Development in Information Retrieval, {SIGIR}
			2004, Sheffield, UK, July 25-29, 2004}, 25--32.
		\newblock \urlprefix\url{http://doi.acm.org/10.1145/1008992.1009000}.
		
		\bibitem[{Campos et~al.(2018{\natexlab{a}})Campos, Mangaravite, Pasquali, Jorge, Nunes and
			Jatowt}]{YAKE2018Campos}
		Campos, R., Mangaravite, V., Pasquali, A., Jorge, A.~M., Nunes, C. and Jatowt,
		A. (2018{\natexlab{a}}) Yake! collection-independent automatic keyword extractor.
		\newblock In \textit{Advances in Information Retrieval - 40th European
			Conference on {IR} Research, {ECIR} 2018, Grenoble, France, March 26-29,
			2018, Proceedings}, 806--810.
		\newblock \urlprefix\url{https://doi.org/10.1007/978-3-319-76941-7\_80}.
		
		\bibitem[{Campos et~al.(2018{\natexlab{b}})Campos, Mangaravite, Pasquali, Jorge, Nunes and
			Jatowt}]{CamposSpringer}
		Campos, R., Mangaravite, V., Pasquali, A., Jorge, A.~M., Nunes, C. and Jatowt,
		A. (2018{\natexlab{b}}) A text feature based automatic keyword extraction method for single
		documents.
		\newblock In \textit{Advances in Information Retrieval - 40th European
			Conference on {IR} Research, {ECIR} 2018, Grenoble, France, March 26-29,
			2018, Proceedings}, 684--691.
		\newblock \urlprefix\url{https://doi.org/10.1007/978-3-319-76941-7\_63}.
		
		\bibitem[{Caragea et~al.(2014)Caragea, Bulgarov, Godea and
			Gollapalli}]{caragea2014citation}
		Caragea, C., Bulgarov, F.~A., Godea, A. and Gollapalli, S.~D. (2014)
		Citation-enhanced keyphrase extraction from research papers: {A} supervised
		approach.
		\newblock In \textit{Proceedings of the 2014 Conference on Empirical Methods in
			Natural Language Processing, {EMNLP} 2014, Doha, Qatar, October 25-29, 2014,
			{A} meeting of SIGDAT, a Special Interest Group of the {ACL}}, 1435--1446.
		\newblock \urlprefix\url{http://aclweb.org/anthology/D/D14/D14-1150.pdf}.
		
		\bibitem[{Caruana(1993)}]{DBLP:conf/icml/Caruana93}
		Caruana, R. (1993) Multitask learning: {A} knowledge-based source of inductive
		bias.
		\newblock In \textit{Machine Learning, Proceedings of the Tenth International
			Conference, University of Massachusetts, Amherst, MA, USA, June 27-29, 1993},
		41--48.
		
		
		\bibitem[{Cer et~al.(2018)Cer, Yang, Kong, Hua, Limtiaco, John, Constant,
			Guajardo{-}Cespedes, Yuan, Tar, Sung, Strope and
			Kurzweil}]{DBLP:journals/corr/abs-1803-11175}
		Cer, D., Yang, Y., Kong, S., Hua, N., Limtiaco, N., John, R.~S., Constant, N.,
		Guajardo{-}Cespedes, M., Yuan, S., Tar, C., Sung, Y., Strope, B. and
		Kurzweil, R. (2018) Universal sentence encoder.
		\newblock \textit{CoRR}, \textbf{abs/1803.11175}.
		\newblock \urlprefix\url{http://arxiv.org/abs/1803.11175}.
		
		\bibitem[{Chen and Goodman(1999)}]{DBLP:journals/csl/ChenG99}
		Chen, S.~F. and Goodman, J. (1999) An empirical study of smoothing techniques
		for language modeling.
		\newblock \textit{Computer Speech {\&} Language}, \textbf{13}, 359--393.
		\newblock \urlprefix\url{https://doi.org/10.1006/csla.1999.0128}.
		
		\bibitem[{Chen et~al.(2018)Chen, Zhang, Wu, Yan and Li}]{CorrRNN2018Chen}
		Chen, J., Zhang, X., Wu, Y., Yan, Z. and Li, Z. (2018) Keyphrase generation
		with correlation constraints.
		\newblock In \textit{Proceedings of the 2018 Conference on Empirical Methods in
			Natural Language Processing, Brussels, Belgium, October 31 - November 4,
			2018}, 4057--4066.
		\newblock \urlprefix\url{https://aclanthology.info/papers/D18-1439/d18-1439}.
		
		\bibitem[{Chuang et~al.(2012{\natexlab{a}})Chuang, Manning and
			Heer}]{DBLP:conf/avi/ChuangMH12}
		Chuang, J., Manning, C.~D. and Heer, J. (2012{\natexlab{a}}) Termite:
		visualization techniques for assessing textual topic models.
		\newblock In \textit{Proceedings of the International Working Conference on
			Advanced Visual Interfaces, {AVI} 2012, Capri Island, Naples, Italy, May
			22-25, 2012}, 74--77.
		\newblock \urlprefix\url{http://doi.acm.org/10.1145/2254556.2254572}.
		
		\bibitem[{Chuang et~al.(2012{\natexlab{b}})Chuang, Manning and
			Heer}]{DBLP:journals/tochi/ChuangMH12}
		Chuang, J., Manning, C.~D. and Heer, J. (2012{\natexlab{b}}) "without the clutter of unimportant words":
		Descriptive keyphrases for text visualization.
		\newblock \textit{{ACM} Transactions on Computer-Human Interaction},
		\textbf{19}, 19:1--19:29.
		\newblock \urlprefix\url{http://doi.acm.org/10.1145/2362364.2362367}.
		
		\bibitem[{Danesh et~al.(2015)Danesh, Sumner and
			Martin}]{DBLP:conf/starsem/DaneshSM15}
		Danesh, S., Sumner, T. and Martin, J.~H. (2015) Sgrank: Combining statistical
		and graphical methods to improve the state of the art in unsupervised
		keyphrase extraction.
		\newblock In \textit{Proceedings of the Fourth Joint Conference on Lexical and
			Computational Semantics, Denver, Colorado, {USA.}, June 4-5, 2015}, 117--126.
		\newblock \urlprefix\url{http://aclweb.org/anthology/S/S15/S15-1013.pdf}.
		
		\bibitem[{Deerwester et~al.(1990)Deerwester, Dumais, Landauer, Furnas and
			Harshman}]{deerwester1990indexing}
		Deerwester, S.~C., Dumais, S.~T., Landauer, T.~K., Furnas, G.~W. and Harshman,
		R.~A. (1990) Indexing by latent semantic analysis.
		\newblock \textit{{J}ournal of the {A}merican {S}ociety for {I}nformation
			{S}cience}, \textbf{41}, 391--407.
		\newblock
		\urlprefix\url{https://doi.org/10.1002/(SICI)1097-4571(199009)41:6<391::AID-ASI1>3.0.CO;2-9}.
		
		\bibitem[{Devlin et~al.(2019)Devlin, Chang, Lee and
			Toutanova}]{DBLP:conf/naacl/DevlinCLT19}
		Devlin, J., Chang, M., Lee, K. and Toutanova, K. (2019) {BERT:} pre-training of
		deep bidirectional transformers for language understanding.
		\newblock In \textit{Proceedings of the 2019 Conference of the North American
			Chapter of the Association for Computational Linguistics: Human Language
			Technologies, {NAACL-HLT} 2019, Minneapolis, MN, USA, June 2-7, 2019, Volume
			1 (Long and Short Papers)}, 4171--4186.
		\newblock \urlprefix\url{https://aclweb.org/anthology/papers/N/N19/N19-1423/}.
		
		\bibitem[{Dice(1945)}]{dice1945measures}
		Dice, L.~R. (1945) Measures of the amount of ecologic association between
		species.
		\newblock \textit{Ecology}, \textbf{26}, 297--302.
		\newblock \urlprefix\url{http://dx.doi.org/10.2307/1932409}.
		
		\bibitem[{El{-}Beltagy and Rafea(2009)}]{DBLP:journals/is/El-BeltagyR09}
		El{-}Beltagy, S.~R. and Rafea, A.~A. (2009) Kp-miner: {A} keyphrase extraction
		system for english and arabic documents.
		\newblock \textit{Information Systems}, \textbf{34}, 132--144.
		\newblock \urlprefix\url{https://doi.org/10.1016/j.is.2008.05.002}.
		
		\bibitem[{Elkan and Noto(2008)}]{elkan2008learning}
		Elkan, C. and Noto, K. (2008) Learning classifiers from only positive and
		unlabeled data.
		\newblock In \textit{Proceedings of the 14th {ACM} {SIGKDD} International
			Conference on Knowledge Discovery and Data Mining, Las Vegas, Nevada, USA,
			August 24-27, 2008}, 213--220.
		\newblock \urlprefix\url{http://doi.acm.org/10.1145/1401890.1401920}.
		
		\bibitem[{Ferragina and Scaiella(2010)}]{WikiRankFerragina2010}
		Ferragina, P. and Scaiella, U. (2010) {TAGME:} on-the-fly annotation of short
		text fragments (by wikipedia entities).
		\newblock In \textit{Proceedings of the 19th {ACM} Conference on Information
			and Knowledge Management, {CIKM}, October 26-30, Toronto, Ontario, Canada,
			2010}, 1625--1628.
		\newblock \urlprefix\url{https://doi.org/10.1145/1871437.1871689}.
		
		
		\bibitem[{Figueroa et~al.(2018)Figueroa, Chen and
			Chen}]{DBLP:journals/csl/FigueroaCC18}
		Figueroa, G., Chen, P. and Chen, Y. (2018) Rankup: Enhancing graph-based
		keyphrase extraction methods with error-feedback propagation.
		\newblock \textit{Computer Speech {\&} Language}, \textbf{47}, 112--131.
		\newblock \urlprefix\url{https://doi.org/10.1016/j.csl.2017.07.004}.
		
		\bibitem[{Florescu and
			Caragea(2017{\natexlab{a}})}]{DBLP:conf/ecir/FlorescuC17}
		Florescu, C. and Caragea, C. (2017{\natexlab{a}}) A new scheme for scoring
		phrases in unsupervised keyphrase extraction.
		\newblock In \textit{Proceedings of the Advances in Information Retrieval -
			39th European Conference on {IR} Research, {ECIR} 2017, Aberdeen, UK, April
			8-13, 2017}, 477--483.
		\newblock \urlprefix\url{https://doi.org/10.1007/978-3-319-56608-5_37}.
		
		\bibitem[{Florescu and Caragea(2017{\natexlab{b}})}]{DBLP:conf/acl/FlorescuC17}
		Florescu, C. and Caragea, C. (2017{\natexlab{b}}) Position{R}ank: An unsupervised approach to keyphrase
		extraction from scholarly documents.
		\newblock In \textit{Proceedings of the 55th Annual Meeting of the Association
			for Computational Linguistics, {ACL} 2017, Vancouver, Canada, July 30 -
			August 4, 2017, Volume 1: Long Papers}, 1105--1115.
		\newblock \urlprefix\url{https://doi.org/10.18653/v1/P17-1102}.
		
		\bibitem[{Gollapalli and Caragea(2014)}]{gollapalli2014extracting}
		Gollapalli, S.~D. and Caragea, C. (2014) Extracting keyphrases from research
		papers using citation networks.
		\newblock In \textit{Proceedings of the 28th {AAAI} Conference on Artificial
			Intelligence, Qu{\'{e}}bec City, Qu{\'{e}}bec, Canada, July 27 -31, 2014},
		1629--1635.
		\newblock
		\urlprefix\url{http://www.aaai.org/ocs/index.php/AAAI/AAAI14/paper/view/8662}.
		
		\bibitem[{Gollapalli et~al.(2017)Gollapalli, Li and
			Yang}]{DBLP:conf/aaai/GollapalliLY17}
		Gollapalli, S.~D., Li, X. and Yang, P. (2017) Incorporating expert knowledge
		into keyphrase extraction.
		\newblock In \textit{Proceedings of the 31st {AAAI} Conference on Artificial
			Intelligence, San Francisco, California, {USA}, February 4-9, 2017},
		3180--3187.
		\newblock
		\urlprefix\url{http://aaai.org/ocs/index.php/AAAI/AAAI17/paper/view/14628}.
		
		\bibitem[{Gutwin et~al.(1999)Gutwin, Paynter, Witten, Nevill{-}Manning and
			Frank}]{gutwin1999improving}
		Gutwin, C., Paynter, G.~W., Witten, I.~H., Nevill{-}Manning, C.~G. and Frank,
		E. (1999) Improving browsing in digital libraries with keyphrase indexes.
		\newblock \textit{Decision Support Systems}, \textbf{27}, 81--104.
		\newblock \urlprefix\url{https://doi.org/10.1016/S0167-9236(99)00038-X}.
		
		\bibitem[{Hasan and Ng(2010)}]{hasan+ng2010}
		Hasan, K.~S. and Ng, V. (2010) Conundrums in unsupervised keyphrase extraction:
		Making sense of the state-of-the-art.
		\newblock In \textit{Proceedings of the 23rd International Conference on
			Computational Linguistics, {COLING} 2010, Beijing, China, August 23-27, 2010,
			Posters Volume}, 365--373.
		\newblock \urlprefix\url{http://aclweb.org/anthology/C/C10/C10-2042.pdf}.
		
		\bibitem[{Hasan and Ng(2014)}]{hasan+ng2014}
		Hasan, K.~S. and Ng, V. (2014) Automatic keyphrase extraction: {A} survey of the state of the art.
		\newblock In \textit{Proceedings of the 52nd Annual Meeting of the Association
			for Computational Linguistics, {ACL} 2014, Baltimore, MD, USA, June 22-27,
			2014, Volume 1: Long Papers}, 1262--1273.
		\newblock \urlprefix\url{http://aclweb.org/anthology/P/P14/P14-1119.pdf}.
		
		\bibitem[{Hashimoto et~al.(2017)Hashimoto, Xiong, Tsuruoka and
			Socher}]{DBLP:conf/emnlp/HashimotoXTS17}
		Hashimoto, K., Xiong, C., Tsuruoka, Y. and Socher, R. (2017) A joint many-task
		model: Growing a neural network for multiple {NLP} tasks.
		\newblock In \textit{Proceedings of the 2017 Conference on Empirical Methods in
			Natural Language Processing, {EMNLP} 2017, Copenhagen, Denmark, September
			9-11, 2017}, 1923--1933.
		\newblock \urlprefix\url{https://aclanthology.info/papers/D17-1206/d17-1206}.
		
		
		\bibitem[{Haveliwala(2002)}]{DBLP:conf/www/Haveliwala02}
		Haveliwala, T.~H. (2002) Topic-sensitive pagerank.
		\newblock In \textit{Proceedings of the Eleventh International World Wide Web
			Conference, {WWW} 2002, Honolulu, Hawaii, May 7-11, 2002}, 517--526.
		\newblock \urlprefix\url{http://doi.acm.org/10.1145/511446.511513}.
		
		\bibitem[{Herings et~al.(2005)Herings, Laan and Talman}]{herings+van+talman}
		Herings, P. J.-J., Laan, G. v.~d. and Talman, D. (2005) Measuring the power of
		nodes in digraphs.
		\newblock \textit{Social Choice and Welfare}, \textbf{24}, 439--454.
		\newblock \urlprefix\url{https://doi.org/10.1007/s00355-003-0308-9}.
		
		\bibitem[{Hulth(2003)}]{hulth2003improved}
		Hulth, A. (2003) Improved automatic keyword extraction given more linguistic
		knowledge.
		\newblock In \textit{Proceedings of the 2003 Conference on Empirical Methods in
			Natural Language Processing, EMNLP 2003, Stroudsburg, PA, USA, 2003},
		216--223. Association for Computational Linguistics.
		\newblock \urlprefix\url{https://doi.org/10.3115/1119355.1119383}.
		
		\bibitem[{Hulth and Megyesi(2006)}]{DBLP:conf/acl/HulthM06}
		Hulth, A. and Megyesi, B. (2006) A study on automatically extracted keywords in
		text categorization.
		\newblock In \textit{{ACL} 2006, 21st International Conference on Computational
			Linguistics and 44th Annual Meeting of the Association for Computational
			Linguistics, Proceedings of the Conference, Sydney, Australia, 17-21 July
			2006}.
		\newblock \urlprefix\url{http://aclweb.org/anthology/P06-1068}.
		
		
		\bibitem[{Jiang et~al.(2009)Jiang, Hu and Li}]{jiang2009ranking}
		Jiang, X., Hu, Y. and Li, H. (2009) A ranking approach to keyphrase extraction.
		\newblock In \textit{Proceedings of the 32nd Annual International {ACM} {SIGIR}
			Conference on Research and Development in Information Retrieval, {SIGIR}
			2009, Boston, MA, USA, July 19-23, 2009}, 756--757.
		\newblock \urlprefix\url{http://doi.acm.org/10.1145/1571941.1572113}.
		
		\bibitem[{Kim et~al.(2010{\natexlab{a}})Kim, Baldwin and
			Kan}]{kim2010evaluating}
		Kim, S.~N., Baldwin, T. and Kan, M. (2010{\natexlab{a}}) Evaluating n-gram
		based evaluation metrics for automatic keyphrase extraction.
		\newblock In \textit{Proceedings of the 23rd International Conference on
			Computational Linguistics, {COLING} 2010, Beijing, China, August 23-27,
			2010}, 572--580.
		\newblock \urlprefix\url{http://aclweb.org/anthology/C10-1065}.
		
		\bibitem[{Kim and Kan(2009)}]{DBLP:conf/mwe/KimK09}
		Kim, S.~N. and Kan, M. (2009) Re-examining automatic keyphrase extraction
		approaches in scientific articles.
		\newblock In \textit{Proceedings of the Workshop on Multiword Expressions:
			Identification, Interpretation, Disambiguation and Applications, MWE@IJCNLP
			2009, Singapore, August 6, 2009}, 9--16.
		\newblock \urlprefix\url{https://aclanthology.info/papers/W09-2902/w09-2902}.
		
		\bibitem[{Kim et~al.(2010{\natexlab{b}})Kim, Medelyan, Kan and
			Baldwin}]{Kim:semeval2010}
		Kim, S.~N., Medelyan, O., Kan, M. and Baldwin, T. (2010{\natexlab{b}})
		Semeval-2010 task 5 : Automatic keyphrase extraction from scientific
		articles.
		\newblock In \textit{Proceedings of the 5th International Workshop on Semantic
			Evaluation, SemEval@ACL 2010, Uppsala, Sweden, July 15-16, 2010}, 21--26.
		\newblock \urlprefix\url{http://aclweb.org/anthology/S/S10/S10-1004.pdf}.
		
		\bibitem[{Kim et~al.(2013)Kim, Medelyan, Kan and
			Baldwin}]{DBLP:journals/lre/KimMKB13}
		Kim, S.~N., Medelyan, O., Kan, M. and Baldwin, T. (2013) Automatic keyphrase extraction from scientific articles.
		\newblock \textit{Language Resources and Evaluation}, \textbf{47}, 723--742.
		\newblock \urlprefix\url{https://doi.org/10.1007/s10579-012-9210-3}.
		
		\bibitem[{El{-}Kishky et~al.(2014)El{-}Kishky, Song, Wang, Voss and
			Han}]{DBLP:journals/pvldb/El-KishkySWVH14}
		El{-}Kishky, A., Song, Y., Wang, C., Voss, C.~R. and Han, J. (2014) Scalable
		topical phrase mining from text corpora.
		\newblock \textit{{PVLDB}}, \textbf{8}, 305--316.
		\newblock \urlprefix\url{http://www.vldb.org/pvldb/vol8/p305-ElKishky.pdf}.
		
		\bibitem[{Kleinberg(1999)}]{kleinberg1999}
		Kleinberg, J.~M. (1999) Authoritative sources in a hyperlinked environment.
		\newblock \textit{Journal of the {ACM}}, \textbf{46}, 604--632.
		\newblock \urlprefix\url{http://doi.acm.org/10.1145/324133.324140}.
		
		\bibitem[{Krapivin et~al.(2008)Krapivin, Autayeu and Marchese}]{krapivin2009}
		Krapivin, M., Autayeu, A. and Marchese, M. (2008) Large dataset for keyphrases
		extraction.
		\newblock In \textit{Technical Report DISI-09-055}. Trento, Italy.
		
		\bibitem[{Langville and Meyer(2003)}]{pagerankLangville2003}
		Langville, A.~N. and Meyer, C.~D. (2003) Survey: Deeper inside pagerank.
		\newblock \textit{Internet Mathematics}, \textbf{1}, 335--380.
		\newblock \urlprefix\url{https://doi.org/10.1080/15427951.2004.10129091}.
		
		\bibitem[{Lau and Baldwin(2016)}]{DBLP:conf/rep4nlp/LauB16}
		Lau, J.~H. and Baldwin, T. (2016) An empirical evaluation of doc2vec with
		practical insights into document embedding generation.
		\newblock In \textit{Proceedings of the 1st Workshop on Representation Learning
			for NLP, Rep4NLP@ACL 2016, Berlin, Germany, August 11, 2016}, 78--86.
		\newblock \urlprefix\url{https://doi.org/10.18653/v1/W16-1609}.
		
		\bibitem[{Lin and Hovy(2003)}]{lin2003automatic}
		Lin, C. and Hovy, E.~H. (2003) Automatic evaluation of summaries using n-gram
		co-occurrence statistics.
		\newblock In \textit{Proceedings of the Human Language Technology Conference of
			the North American Chapter of the Association for Computational Linguistics,
			{HLT-NAACL} 2003, Edmonton, Canada, May 27 - June 1, 2003}.
		\newblock \urlprefix\url{http://aclweb.org/anthology/N/N03/N03-1020.pdf}.
		
		\bibitem[{Liu et~al.(2010)Liu, Huang, Zheng and Sun}]{liu2010automatic}
		Liu, Z., Huang, W., Zheng, Y. and Sun, M. (2010) Automatic keyphrase extraction
		via topic decomposition.
		\newblock In \textit{Proceedings of the 2010 Conference on Empirical Methods in
			Natural Language Processing, {EMNLP} 2010, {MIT} Stata Center, Massachusetts,
			USA, October 9-11, 2010, {A} meeting of SIGDAT, a Special Interest Group of
			the {ACL}}, 366--376.
		\newblock \urlprefix\url{http://www.aclweb.org/anthology/D10-1036}.
		
		\bibitem[{Liu et~al.(2009)Liu, Li, Zheng and Sun}]{liu2009clustering}
		Liu, Z., Li, P., Zheng, Y. and Sun, M. (2009) Clustering to find exemplar terms
		for keyphrase extraction.
		\newblock In \textit{Proceedings of the 2009 Conference on Empirical Methods in
			Natural Language Processing, {EMNLP} 2009, Singapore, August 6-7, 2009, {A}
			meeting of SIGDAT, a Special Interest Group of the {ACL}}, 257--266.
		\newblock \urlprefix\url{http://www.aclweb.org/anthology/D09-1027}.
		
		\bibitem[{Liu and {\"{O}}zsu(2009)}]{DBLP:reference/db/2009}
		Liu, L. and {\"{O}}zsu, M.~T. (eds.) (2009) \textit{Encyclopedia of Database
			Systems}.
		\newblock Springer {US}.
		\newblock \urlprefix\url{https://doi.org/10.1007/978-0-387-39940-9}.
		
		\bibitem[{Mahata et~al.(2018)Mahata, Kuriakose, Shah and
			Zimmermann}]{key2vec2018Mahata}
		Mahata, D., Kuriakose, J., Shah, R.~R. and Zimmermann, R. (2018) Key2vec:
		Automatic ranked keyphrase extraction from scientific articles using phrase
		embeddings.
		\newblock In \textit{Proceedings of the 2018 Conference of the North American
			Chapter of the Association for Computational Linguistics: Human Language
			Technologies, NAACL-HLT, New Orleans, Louisiana, USA, June 1-6, 2018, Volume
			2 (Short Papers)}, 634--639.
		\newblock \urlprefix\url{https://aclanthology.info/papers/N18-2100/n18-2100}.
		
		\bibitem[{Marujo et~al.(2012)Marujo, Gershman, Carbonell, Frederking and
			Neto}]{newsMarujo2012}
		Marujo, L., Gershman, A., Carbonell, J.~G., Frederking, R.~E. and Neto, J.~P.
		(2012) Supervised topical key phrase extraction of news stories using
		crowdsourcing, light filtering and co-reference normalization.
		\newblock In \textit{Proceedings of the Eighth International Conference on
			Language Resources and Evaluation, {LREC} 2012, Istanbul, Turkey, May 23-25,
			2012}, 399--403.
		\newblock
		\urlprefix\url{http://www.lrec-conf.org/proceedings/lrec2012/summaries/672.html}.
		
		\bibitem[{Marujo et~al.(2011)Marujo, Viveiros and
			Neto}]{DBLP:conf/interspeech/MarujoVN11}
		Marujo, L., Viveiros, M. and Neto, J.~P. (2011) Keyphrase cloud generation of
		broadcast news.
		\newblock In \textit{Proceedings of the {INTERSPEECH} 2011, 12th Annual
			Conference of the International Speech Communication Association, Florence,
			Italy, August 27-31, 2011}, 2393--2396.
		\newblock
		\urlprefix\url{http://www.isca-speech.org/archive/interspeech\_2011/i11\_2393.html}.
		
		
		\bibitem[{McIlraith and Weinberger(2018)}]{DBLP:conf/aaai/2018}
		McIlraith, S.~A. and Weinberger, K.~Q. (2018) Learning feature representations
		for keyphrase extraction.
		\newblock In \textit{Proceedings of the Thirty-Second {AAAI} Conference on
			Artificial Intelligence, New Orleans, Louisiana, USA, February 2-7, 2018}.
		\newblock
		\urlprefix\url{https://www.aaai.org/ocs/index.php/AAAI/AAAI18/schedConf/presentations}.
		
		\bibitem[{Medelyan et~al.(2009)Medelyan, Frank and Witten}]{medelyan2009human}
		Medelyan, O., Frank, E. and Witten, I.~H. (2009) Human-competitive tagging
		using automatic keyphrase extraction.
		\newblock In \textit{Proceedings of the 2009 Conference on Empirical Methods in
			Natural Language Processing, {EMNLP} 2009, Singapore, August 6-7, 2009, {A}
			meeting of SIGDAT, a Special Interest Group of the {ACL}}, 1318--1327.
		\newblock \urlprefix\url{http://www.aclweb.org/anthology/D09-1137}.
		
		\bibitem[{Meng et~al.(2017)Meng, Zhao, Han, He, Brusilovsky and
			Chi}]{meng2017deep}
		Meng, R., Zhao, S., Han, S., He, D., Brusilovsky, P. and Chi, Y. (2017) Deep
		keyphrase generation.
		\newblock In \textit{Proceedings of the 55th Annual Meeting of the Association
			for Computational Linguistics, {ACL} 2017, Vancouver, Canada, July 30 -
			August 4, 2017, Volume 1: Long Papers}, 582--592.
		\newblock \urlprefix\url{https://doi.org/10.18653/v1/P17-1054}.
		
		\bibitem[{Merity et~al.(2018)Merity, Keskar and
			Socher}]{DBLP:journals/corr/abs-1803-08240}
		Merity, S., Keskar, N.~S. and Socher, R. (2018) An analysis of neural language
		modeling at multiple scales.
		\newblock \textit{CoRR}, \textbf{abs/1803.08240}.
		\newblock \urlprefix\url{http://arxiv.org/abs/1803.08240}.
		
		
		\bibitem[{Mihalcea and Tarau(2004)}]{mihalcea+tatau2004}
		Mihalcea, R. and Tarau, P. (2004) Text{R}ank: Bringing order into text.
		\newblock In \textit{Proceedings of the 2004 Conference on Empirical Methods in
			Natural Language Processing, {EMNLP} 2004, Barcelona, Spain, July 25-26,
			2004, {A} meeting of SIGDAT, a Special Interest Group of the ACL, held in
			conjunction with {ACL} 2004}, 404--411.
		\newblock \urlprefix\url{http://www.aclweb.org/anthology/W04-3252}.
		
		\bibitem[{Mikolov et~al.(2013)Mikolov, Chen, Corrado and
			Dean}]{DBLP:journals/corr/abs-1301-3781}
		Mikolov, T., Chen, K., Corrado, G. and Dean, J. (2013) Efficient estimation of
		word representations in vector space.
		\newblock In \textit{Proceedings of the 2013 International Conference on
			Learning Representations, ICLR 2013, Workshop Track, Scottsdale, Arizona,
			USA, May 2-4, 2013}.
		\newblock \urlprefix\url{http://arxiv.org/abs/1301.3781}.
		
		\bibitem[{Miwa and Bansal(2016)}]{DBLP:conf/acl/MiwaB16}
		Miwa, M. and Bansal, M. (2016) End-to-end relation extraction using lstms on
		sequences and tree structures.
		\newblock In \textit{Proceedings of the 54th Annual Meeting of the Association
			for Computational Linguistics, {ACL} 2016, August 7-12, 2016, Berlin,
			Germany, Volume 1: Long Papers}.
		\newblock \urlprefix\url{http://aclweb.org/anthology/P/P16/P16-1105.pdf}.
		
		\bibitem[{Moldovan et~al.(2000)Moldovan, Harabagiu, Pasca, Mihalcea, Girju,
			Goodrum and Rus}]{Moldovan}
		Moldovan, D.~I., Harabagiu, S.~M., Pasca, M., Mihalcea, R., Girju, R., Goodrum,
		R. and Rus, V. (2000) The structure and performance of an open-domain
		question answering system.
		\newblock In \textit{38th Annual Meeting of the Association for Computational
			Linguistics, Hong Kong, China, October 1-8, 2000.}
		\newblock \urlprefix\url{http://www.aclweb.org/anthology/P00-1071}.
		
		\bibitem[{Nguyen and Kan(2007)}]{DBLP:conf/icadl/NguyenK07}
		Nguyen, T.~D. and Kan, M. (2007) Keyphrase extraction in scientific
		publications.
		\newblock In \textit{Proceedings of the Asian Digital Libraries. Looking Back
			10 Years and Forging New Frontiers, 10th International Conference on Asian
			Digital Libraries, {ICADL} 2007, Hanoi, Vietnam, December 10-13, 2007},
		317--326.
		\newblock \urlprefix\url{https://doi.org/10.1007/978-3-540-77094-7_41}.
		
		\bibitem[{Nguyen and Luong(2010)}]{DBLP:conf/semeval/NguyenL10}
		Nguyen, T.~D. and Luong, M. (2010) {WINGNUS:} keyphrase extraction utilizing
		document logical structure.
		\newblock In \textit{Proceedings of the 5th International Workshop on Semantic
			Evaluation, SemEval@ACL 2010, Uppsala, Sweden, July 15-16, 2010}, 166--169.
		\newblock \urlprefix\url{http://aclweb.org/anthology/S/S10/S10-1035.pdf}.
		
		\bibitem[{Pagliardini et~al.(2018)Pagliardini, Gupta and
			Jaggi}]{DBLP:conf/naacl/PagliardiniGJ18}
		Pagliardini, M., Gupta, P. and Jaggi, M. (2018) Unsupervised learning of
		sentence embeddings using compositional n-gram features.
		\newblock In \textit{Proceedings of the 2018 Conference of the North American
			Chapter of the Association for Computational Linguistics: Human Language
			Technologies, {NAACL-HLT} 2018, New Orleans, Louisiana, USA, June 1-6, 2018,
			Volume 1 (Long Papers)}, 528--540.
		\newblock \urlprefix\url{https://aclanthology.info/papers/N18-1049/n18-1049}.
		
		\bibitem[{Papagiannopoulou and Tsoumakas(2018)}]{papagiannopoulou2018local}
		Papagiannopoulou, E. and Tsoumakas, G. (2018) Local word vectors guiding
		keyphrase extraction.
		\newblock \textit{Information Processing \& Management}, \textbf{54}, 888--902.
		\newblock \urlprefix\url{https://doi.org/10.1016/j.ipm.2018.06.004}.
		
		\bibitem[{Papineni et~al.(2002)Papineni, Roukos, Ward and
			Zhu}]{papineni2002bleu}
		Papineni, K., Roukos, S., Ward, T. and Zhu, W. (2002) Bleu: a method for
		automatic evaluation of machine translation.
		\newblock In \textit{Proceedings of the 40th Annual Meeting of the Association
			for Computational Linguistics, Philadelphia, PA, {USA.}, July 6-12, 2002},
		311--318.
		\newblock \urlprefix\url{http://www.aclweb.org/anthology/P02-1040.pdf}.
		
		\bibitem[{Pennington et~al.(2014)Pennington, Socher and
			Manning}]{Pennington14glove:global}
		Pennington, J., Socher, R. and Manning, C.~D. (2014) Glove: Global vectors for
		word representation.
		\newblock In \textit{Proceedings of the 2014 Conference on Empirical Methods in
			Natural Language Processing, {EMNLP} 2014, Doha, Qatar, October 25-29, 2014,
			{A} meeting of SIGDAT, a Special Interest Group of the {ACL}}, 1532--1543.
		\newblock \urlprefix\url{http://aclweb.org/anthology/D/D14/D14-1162.pdf}.
		
		\bibitem[{Przybocki and Martin(1999)}]{przybocki19991999}
		Przybocki, M.~A. and Martin, A.~F. (1999) The 1999 {NIST} speaker recognition
		evaluation, using summed two-channel telephone data for speaker detection and
		speaker tracking.
		\newblock In \textit{Proceedings of the Sixth European Conference on Speech
			Communication and Technology, {EUROSPEECH} 1999, Budapest, Hungary, September
			5-9, 1999}.
		\newblock
		\urlprefix\url{http://www.isca-speech.org/archive/eurospeech\_1999/e99\_2215.html}.
		
		\bibitem[{Rose et~al.(2010)Rose, Engel, Cramer and Cowley}]{rose2010automatic}
		Rose, S., Engel, D., Cramer, N. and Cowley, W. (2010) Automatic keyword
		extraction from individual documents.
		\newblock \textit{Text Mining: Applications and Theory}, 1--20.
		\newblock \urlprefix\url{http://dx.doi.org/10.1002/9780470689646.ch1}.
		
		\bibitem[{Rousseau and Vazirgiannis(2015)}]{DBLP:conf/ecir/RousseauV15}
		Rousseau, F. and Vazirgiannis, M. (2015) Main core retention on graph-of-words
		for single-document keyword extraction.
		\newblock In \textit{Proceedings of the Advances in Information Retrieval -
			37th European Conference on {IR} Research, {ECIR} 2015, Vienna, Austria,
			March 29 - April 2, 2015}, 382--393.
		\newblock \urlprefix\url{https://doi.org/10.1007/978-3-319-16354-3_42}.
		
		\bibitem[{Rush et~al.(2015)Rush, Chopra and Weston}]{summarizationRush2015}
		Rush, A.~M., Chopra, S. and Weston, J. (2015) A neural attention model for
		abstractive sentence summarization.
		\newblock In \textit{Proceedings of the 2015 Conference on Empirical Methods in
			Natural Language Processing, {EMNLP} 2015, Lisbon, Portugal, September 17-21,
			2015}, 379--389.
		\newblock \urlprefix\url{http://aclweb.org/anthology/D/D15/D15-1044.pdf}.
		
		\bibitem[{Shi et~al.(2008)Shi, Jiao, Hou and Li}]{Wiki2008shi}
		Shi, T., Jiao, S., Hou, J. and Li, M. (2008) Improving keyphrase extraction
		using wikipedia semantics.
		\newblock In \textit{Proceedings of Second International Symposium on
			Intelligent Information Technology Application, 2008. IITA'08.}, vol.~2,
		42--46. IEEE.
		
		\bibitem[{Shi et~al.(2017)Shi, Zheng, Yu, Cheng and
			Zou}]{DBLP:journals/dase/ShiZYCZ17}
		Shi, W., Zheng, W., Yu, J.~X., Cheng, H. and Zou, L. (2017) Keyphrase
		extraction using knowledge graphs.
		\newblock \textit{Data Science and Engineering}, \textbf{2}, 275--288.
		\newblock \urlprefix\url{https://doi.org/10.1007/s41019-017-0055-z}.
		
		\bibitem[{Song et~al.(2006)Song, Song, Allen and
			Obradovic}]{DBLP:conf/jcdl/SongSAO06}
		Song, M., Song, I., Allen, R.~B. and Obradovic, Z. (2006) Keyphrase
		extraction-based query expansion in digital libraries.
		\newblock In \textit{{ACM/IEEE} Joint Conference on Digital Libraries, {JCDL}
			2006, Chapel Hill, NC, USA, June 11-15, 2006, Proceedings}, 202--209.
		\newblock \urlprefix\url{https://doi.org/10.1145/1141753.1141800}.
		
		\bibitem[{S{\o}gaard and Bingel(2017)}]{SogaardB17}
		S{\o}gaard, A. and Bingel, J. (2017) Identifying beneficial task relations for
		multi-task learning in deep neural networks.
		\newblock In \textit{Proceedings of the 15th Conference of the European Chapter
			of the Association for Computational Linguistics, {EACL} 2017, Valencia,
			Spain, April 3-7, 2017, Volume 2: Short Papers}, 164--169.
		\newblock \urlprefix\url{https://aclanthology.info/papers/E17-2026/e17-2026}.
		
		\bibitem[{Sterckx et~al.(2016)Sterckx, Caragea, Demeester and
			Develder}]{sterckx2016supervised}
		Sterckx, L., Caragea, C., Demeester, T. and Develder, C. (2016) Supervised
		keyphrase extraction as positive unlabeled learning.
		\newblock In \textit{Proceedings of the 2016 Conference on Empirical Methods in
			Natural Language Processing, {EMNLP} 2016, Austin, Texas, USA, November 1-4,
			2016}, 1924--1929.
		\newblock \urlprefix\url{http://aclweb.org/anthology/D/D16/D16-1198.pdf}.
		
		\bibitem[{Sterckx et~al.(2015{\natexlab{a}})Sterckx, Demeester, Deleu and
			Develder}]{DBLP:conf/www/SterckxDDD15}
		Sterckx, L., Demeester, T., Deleu, J. and Develder, C. (2015{\natexlab{a}})
		Topical word importance for fast keyphrase extraction.
		\newblock In \textit{Proceedings of the 24th International Conference on World
			Wide Web Companion, {WWW} 2015, Florence, Italy, May 18-22, 2015 - Companion
			Volume}, 121--122.
		\newblock \urlprefix\url{http://doi.acm.org/10.1145/2740908.2742730}.
		
		\bibitem[{Sterckx et~al.(2015{\natexlab{b}})Sterckx, Demeester, Deleu and
			Develder}]{DBLP:conf/www/SterckxDDD15a}
		Sterckx, L., Caragea, C., Demeester, T. and Develder, C. (2015{\natexlab{b}}) When topic models disagree: Keyphrase extraction with
		multiple topic models.
		\newblock In \textit{Proceedings of the 24th International Conference on World
			Wide Web Companion, {WWW} 2015, Florence, Italy, May 18-22, 2015 - Companion
			Volume}, 123--124.
		\newblock \urlprefix\url{http://doi.acm.org/10.1145/2740908.2742731}.
		
		\bibitem[{Sterckx et~al.(2018)Sterckx, Demeester, Deleu and
			Develder}]{DBLP:journals/lre/SterckxDDD18}
		Sterckx, L., Caragea, C., Demeester, T. and Develder, C. (2018) Creation and evaluation of large keyphrase extraction collections
		with multiple opinions.
		\newblock \textit{Language Resources and Evaluation}, \textbf{52}, 503--532.
		\newblock \urlprefix\url{https://doi.org/10.1007/s10579-017-9395-6}.
		
		\bibitem[{Stubbs(2003)}]{Stubbs2003}
		Stubbs, M. (2003) Two quantitative methods of studying phraseology in english.
		\newblock \textit{International Journal of Corpus Linguistics}, \textbf{7},
		215--244.
		
		\bibitem[{Teneva and Cheng(2017)}]{DBLP:conf/acl/TenevaC17}
		Teneva, N. and Cheng, W. (2017) Salience rank: Efficient keyphrase extraction
		with topic modeling.
		\newblock In \textit{Proceedings of the 55th Annual Meeting of the Association
			for Computational Linguistics, {ACL} 2017, Vancouver, Canada, July 30 -
			August 4, Volume 2: Short Papers}, 530--535.
		\newblock \urlprefix\url{https://doi.org/10.18653/v1/P17-2084}.
		
		\bibitem[{Tomokiyo and Hurst(2003)}]{tomokiyo2003language}
		Tomokiyo, T. and Hurst, M. (2003) A language model approach to keyphrase
		extraction.
		\newblock In \textit{Proceedings of the 2003 ACL Workshop on Multiword
			Expressions: Analysis, Acquisition and Treatment, MWE 2003, Sapporo, Japan,
			2003, Volume - 18}, 33--40.
		\newblock \urlprefix\url{https://doi.org/10.3115/1119282.1119287}.
		
		\bibitem[{Tu et~al.(2016)Tu, Lu, Liu, Liu and Li}]{CoverageMechTu2016}
		Tu, Z., Lu, Z., Liu, Y., Liu, X. and Li, H. (2016) Modeling coverage for neural
		machine translation.
		\newblock In \textit{Proceedings of the 54th Annual Meeting of the Association
			for Computational Linguistics, {ACL} 2016, August 7-12, 2016, Berlin,
			Germany, Volume 1: Long Papers}.
		\newblock \urlprefix\url{http://aclweb.org/anthology/P/P16/P16-1008.pdf}.
		
		\bibitem[{Turney(2001)}]{DBLP:conf/ecml/Turney01}
		Turney, P.~D. (2001) Mining the web for synonyms: {PMI-IR} versus {LSA} on
		{TOEFL}.
		\newblock In \textit{Proceedings of the 12th European Conference on Machine
			Learning: {ECML} 2001, Freiburg, Germany, September 5-7, 2001}, 491--502.
		\newblock \urlprefix\url{https://doi.org/10.1007/3-540-44795-4\_42}.
		
		\bibitem[{Voorhees(1999)}]{voorhees1999trec}
		Voorhees, E.~M. (1999) The {TREC-8} question answering track report.
		\newblock In \textit{Proceedings of The Eighth Text REtrieval Conference,
			{TREC} 1999, Gaithersburg, Maryland, USA, November 17-19, 1999}.
		\newblock
		\urlprefix\url{http://trec.nist.gov/pubs/trec8/papers/qa\_report.pdf}.
		
		\bibitem[{Wan and Xiao(2008)}]{wan+xiao2008}
		Wan, X. and Xiao, J. (2008) Single document keyphrase extraction using
		neighborhood knowledge.
		\newblock In \textit{Proceedings of the 23rd {AAAI} Conference on Artificial
			Intelligence, {AAAI} 2008, Chicago, Illinois, USA, July 13-17, 2008},
		855--860.
		\newblock \urlprefix\url{http://www.aaai.org/Library/AAAI/2008/aaai08-136.php}.
		
		\bibitem[{Wang and Li(2017)}]{DBLP:conf/semeval/WangL17}
		Wang, L. and Li, S. (2017) Pku{\_}icl at semeval-2017 task 10: Keyphrase
		extraction with model ensemble and external knowledge.
		\newblock In \textit{Proceedings of the 11th International Workshop on Semantic
			Evaluation, SemEval@ACL 2017, Vancouver, Canada, August 3-4, 2017}, 934--937.
		\newblock \urlprefix\url{https://doi.org/10.18653/v1/S17-2161}.
		
		\bibitem[{Wang et~al.(2014)Wang, Liu and McDonald}]{Wang2014}
		Wang, R., Liu, W. and McDonald, C. (2014) Corpus-independent generic keyphrase
		extraction using word embedding vectors.
		\newblock \textit{Software Engineering Research Conference}, 39.
		
		\bibitem[{Wang et~al.(2015)Wang, Liu and McDonald}]{DBLP:conf/adc/WangLM15}
		Wang, R., Liu, W. and McDonald, C. (2015) Using word embeddings to enhance keyword identification for
		scientific publications.
		\newblock In \textit{Proceedings of the Databases Theory and Applications -
			26th Australasian Database Conference, {ADC} 2015, Melbourne, VIC, Australia,
			June 4-7, 2015}, 257--268.
		\newblock \urlprefix\url{https://doi.org/10.1007/978-3-319-19548-3_21}.
		
		\bibitem[{Wang et~al.(2018)Wang, Liu, Qin, Xu, Wang, Chen and
			Xiong}]{DBLP:conf/icdm/WangLQXWCX18}
		Wang, Y., Liu, Q., Qin, C., Xu, T., Wang, Y., Chen, E. and Xiong, H. (2018)
		Exploiting topic-based adversarial neural network for cross-domain keyphrase
		extraction.
		\newblock In \textit{Proceedings of the {IEEE} International Conference on Data
			Mining, {ICDM} 2018, Singapore, November 17-20, 2018}, 597--606.
		\newblock \urlprefix\url{https://doi.org/10.1109/ICDM.2018.00075}.
		
		\bibitem[{Witten(2003)}]{witten1999browsing}
		Witten, I.~H. (2003) Browsing around a digital library.
		\newblock In \textit{Proceedings of the 14th Annual {ACM-SIAM} Symposium on
			Discrete Algorithms, Baltimore, Maryland, {USA}, January 12-14, 2003},
		99--99.
		\newblock \urlprefix\url{http://dl.acm.org/citation.cfm?id=644108.644125}.
		
		\bibitem[{Witten et~al.(1999)Witten, Paynter, Frank, Gutwin and
			Nevill{-}Manning}]{witten1999kea}
		Witten, I.~H., Paynter, G.~W., Frank, E., Gutwin, C. and Nevill{-}Manning,
		C.~G. (1999) {KEA:} practical automatic keyphrase extraction.
		\newblock In \textit{Proceedings of the 4th {ACM} conference on Digital
			Libraries, Berkeley, CA, {USA}, August 11-14, 1999}, 254--255.
		\newblock \urlprefix\url{http://doi.acm.org/10.1145/313238.313437}.
		
		
		\bibitem[{Won et~al.(2019)Won, Martins and Raimundo}]{wonautomatic}
		Won, M., Martins, B. and Raimundo, F. (2019) Automatic extraction of relevant
		keyphrases for the study of issue competition.
		\newblock In \textit{Proceedings of the 20th International Conference on
			Computational Linguistics and Intelligent Text Processing, Berkeley, La
			Rochelle, {France}, April 7-13, 2019}.
		
		\bibitem[{Yang et~al.(2019)Yang, Dai, Yang, Carbonell, Salakhutdinov and
			Le}]{DBLP:journals/corr/abs-1906-08237}
		Yang, Z., Dai, Z., Yang, Y., Carbonell, J.~G., Salakhutdinov, R. and Le, Q.~V.
		(2019) Xlnet: Generalized autoregressive pretraining for language
		understanding.
		\newblock \textit{CoRR}, \textbf{abs/1906.08237}.
		\newblock \urlprefix\url{http://arxiv.org/abs/1906.08237}.
		
		\bibitem[{Yang et~al.(2018)Yang, Liang, Zhao, Xu, Zhu and
			Qu}]{DBLP:journals/mta/YangLZXZQ18}
		Yang, M., Liang, Y., Zhao, W., Xu, W., Zhu, J. and Qu, Q. (2018) Task-oriented
		keyphrase extraction from social media.
		\newblock \textit{Multimedia Tools and Applications}, \textbf{77}, 3171--3187.
		\newblock \urlprefix\url{https://doi.org/10.1007/s11042-017-5041-y}.
		
		\bibitem[{Ye and Wang(2018)}]{semi2018YE}
		Ye, H. and Wang, L. (2018) Semi-supervised learning for neural keyphrase
		generation.
		\newblock In \textit{Proceedings of the 2018 Conference on Empirical Methods in
			Natural Language Processing, Brussels, Belgium, October 31 - November 4,
			2018}, 4142--4153.
		\newblock \urlprefix\url{https://aclanthology.info/papers/D18-1447/d18-1447}.
		
		\bibitem[{Yu and Ng(2018)}]{wikirank2018yu}
		Yu, Y. and Ng, V. (2018) Wikirank: Improving keyphrase extraction based on
		background knowledge.
		\newblock In \textit{Proceedings of the 11th edition of the Language Resources
			and Evaluation Conference, {LREC} 2018, 7-12 May 2018, Miyazaki (Japan)},
		3723--3727.
		\newblock
		\urlprefix\url{http://www.lrec-conf.org/proceedings/lrec2018/pdf/871.pdf}.
		
		\bibitem[{Zesch and Gurevych(2009)}]{zesch2009approximate}
		Zesch, T. and Gurevych, I. (2009) Approximate matching for evaluating keyphrase
		extraction.
		\newblock In \textit{Proceedings of the Recent Advances in Natural Language
			Processing, {RANLP}, Borovets, Bulgaria, 14-16 September, 2009}, 484--489.
		\newblock \urlprefix\url{http://aclweb.org/anthology/R/R09/R09-1086.pdf}.
		
		\bibitem[{Zhang et~al.(2016)Zhang, Wang, Gong and
			Huang}]{DBLP:conf/emnlp/ZhangWGH16}
		Zhang, Q., Wang, Y., Gong, Y. and Huang, X. (2016) Keyphrase extraction using
		deep recurrent neural networks on twitter.
		\newblock In \textit{Proceedings of the 2016 Conference on Empirical Methods in
			Natural Language Processing, {EMNLP} 2016, Austin, Texas, USA, November 1-4,
			2016}, 836--845.
		\newblock \urlprefix\url{http://aclweb.org/anthology/D/D16/D16-1080.pdf}.
		
		\bibitem[{Zhang et~al.(2017)Zhang, Chang, Liu, Gollapalli, Li and
			Xiao}]{DBLP:conf/cikm/ZhangCLG0X17}
		Zhang, Y., Chang, Y., Liu, X., Gollapalli, S.~D., Li, X. and Xiao, C. (2017)
		{MIKE:} keyphrase extraction by integrating multidimensional information.
		\newblock In \textit{Proceedings of the 2017 {ACM} on Conference on Information
			and Knowledge Management, {CIKM} 2017, Singapore, November 06 - 10, 2017},
		1349--1358.
		\newblock \urlprefix\url{http://doi.acm.org/10.1145/3132847.3132956}.
		
		\bibitem[{Zhang et~al.(2004)Zhang, Zincir{-}Heywood and
			Milios}]{ZhangZM04}
		Zhang, Y., Zincir{-}Heywood, A.~N. and Milios, E.~E. (2004) World wide web site
		summarization.
		\newblock \textit{Web Intelligence and Agent Systems}, \textbf{2}, 39--53.
		\newblock \urlprefix\url{http://content.iospress.com/articles/web-intelligence-and-agent-systems-an-international-journal/wia00026}.
		
		
	\end{thebibliography}
\end{document}